\documentclass{elsarticle}

\usepackage{graphicx} 
\usepackage{amsmath}
\usepackage{longtable}
\usepackage{xspace} 
\usepackage{algorithm}
\usepackage{algorithmic}
\usepackage{caption}
\usepackage{subcaption}
\usepackage{cleveref}
\makeatletter
\renewcommand{\p@subfigure}{\thefigure}

\makeatother
\usepackage{float}
\usepackage{placeins} 
\usepackage[top=1.5in]{geometry}  
\usepackage[export]{adjustbox}
\usepackage{tabularx}

\date{August 2024}
\usepackage{titlesec}  

\newcounter{subsubsubsection}[subsubsection]
\renewcommand{\thesubsubsubsection}{\thesubsubsection.\arabic{subsubsubsection}}

\titleclass{\subsubsubsection}{straight}[\subsubsection]

\titleformat{\subsubsubsection}
  {\normalfont\normalsize\bfseries}{\thesubsubsubsection}{1em}{}

\titlespacing*{\subsubsubsection}
  {0pt}{1.0em plus 0.5em minus 0.2em}{0.5em}

\makeatletter
\@addtoreset{subsubsubsection}{subsubsection}
\makeatother
\begin{document}
\sloppy 
\begin{frontmatter}



\title{ExpSpeech-Net: Multimodal Fusion of Expression and Speech for Deepfake Detection}


\author{Ruchika Sharma, Rudresh Dwivedi}

\affiliation{organization={Department of Computer Science and Engineering, Netaji Subhas University of Technology},
            addressline={Dwarka}, 
            city={Delhi},
            postcode={110078}, 
            state={NCT of Delhi},
            country={India}}

\begin{abstract}
Deepfake videos are increasingly challenging the credibility of online content. Many existing detection methodology relies on complex, resource-intensive models, which limit their practical use. The study introduces the ExpSpeech-Net deepfake detection (SqN-R-DFD) model, which utilizes SqueezeNet and RNN (Recurrent Neural Network) as its backbone, providing a lightweight and efficient deepfake detection framework that simultaneously analyzes facial expressions and speech patterns. The approach incorporates advanced feature extraction, such as ISLBT-based features for image and MPNCC for signals, along with a smart feature-selection strategy using SASMA (Sandpiper-Assisted Slime Mould Algorithm), ensuring optimal and balanced input to the detection models. By combining SqueezeNet and an RNN, subtle inconsistencies in deepfake videos are captured effectively. The framework achieves $94.5\%$ accuracy, precision of $99.3\%$, and F-measure of $96.8\%$, outperforming conventional methods. This demonstrates that integrating multiple modalities with intelligent preprocessing and feature selection enables practical, real-time deepfake detection suitable for everyday applications.
\end{abstract}

\begin{graphicalabstract}
     \includegraphics[scale=0.7]{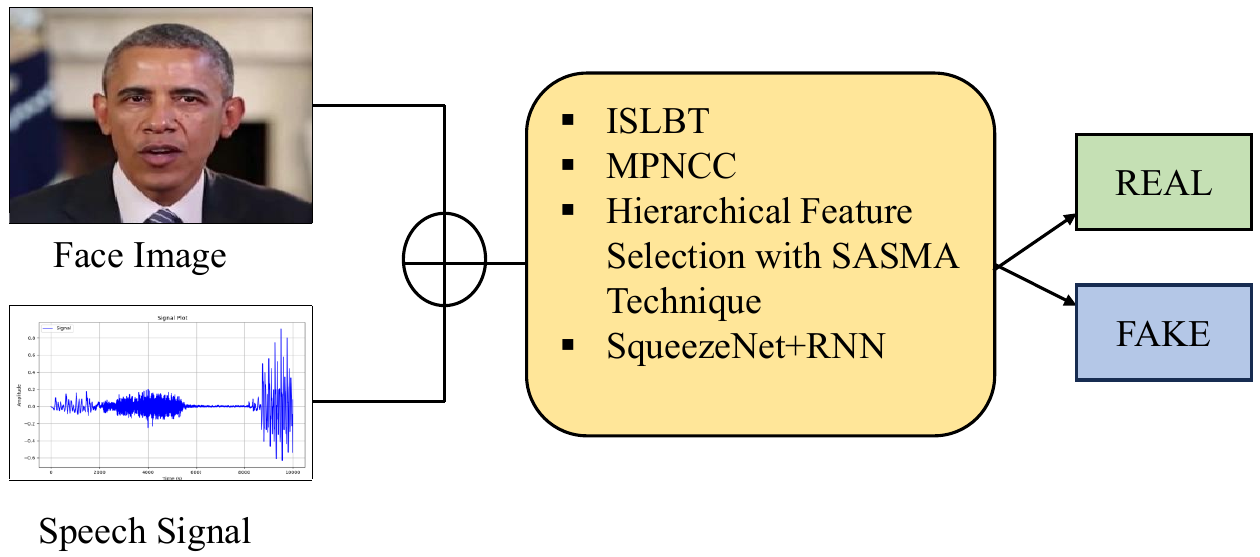}
\end{graphicalabstract}

\newpage

\begin{highlights}
	\item Proposed a robust deepfake detector fusing facial expressions with speech cues.
	\item Proposed ISLBT-based features using enhanced LBP over shape-free patches.
	\item Proposed MPNCC, a robust enhancement of PNCC tailored to noisy environments.
	\item Introduction of HFS-driven optimization incorporating a novel SASMA algorithm.	

\end{highlights}
\newpage

\end{frontmatter}

\section{Introduction}
Over the last decade, there has been a significant shift in internet traffic from text-based content to multimedia files. The rise of expansive multimedia social platforms such as Instagram, TikTok, and WeChat has profoundly impacted our daily lives \cite{wang2022deepfake}. 
However, these platforms pose potential risks to public safety, political security, and personal privacy, which can harm their image.
The amalgamation of forgery techniques and AI (Artificial Intelligence) advancements has significantly improved the difficulty of distinguishing genuine from manipulated digital media \cite{tolosana2022deepfakes}. At its worst, it can incite organized and destabilizing public actions driven by distorted impressions or ideas. Deepfakes \cite{dong2023contrastive} refer to artificially created audio-visual representations of individuals, often produced without consent, that can defame public figures. Simultaneously, many detection methods have improved to detect subtle manipulations left inadvertently by recent generation techniques; however, many digital forensic methodologies fail to identify unseen forgeries. 
The emergence of Deepfake technology presents critical safety concerns, demanding the development of novel techniques for detection and mitigation. The methods proposed in \cite{tran2022generalization} leverage DNNs (Deep Neural Networks) as a feature extractor to efficiently explore forensic clues in a data-centric approach, demonstrating significant potential in detection accuracy. However, the use of complex and deeper network architectures results in an extensive number of network parameters and FLOPs (Floating-Point Operations). Detecting deepfakes remains a challenging task for humans; however, recent studies indicate that a CNN (Convolutional Neural Network) trained specifically for this purpose can surprisingly achieve effective detection.\\  A joint distillation-based compression framework proposed by Xu et al. \cite{xu2023novel} achieved competitive detection performance by reducing the computational cost. However, enhancing the suggested scheme for feature distillation with a heterogeneous framework in transitional layers is difficult. To address these challenges, a novel approach for effective deep-fake detection utilizing a multimodal data fusion approach is proposed and described in section \ref{proposed}. \\


The major contribution of this work is as follows:
\begin{enumerate}
   
    \item [1.] Proposed ISLBT-based features, in which the standard form of  LBP (Local Binary Pattern) is performed on the shape-free patches to extract features from the preprocessed face image, are modified along with local and global features, DSBME, and deep features are extracted. 
    \item[2.]  MPNCC-based features are extracted where approaches such as pre-emphasis, hamming window, IMS (Improved Magnitude Squared), large time power calculation, channel bias minimizing, DCT (Discrete Cosine Transform), and CMN (Cepstral Mean Normalization) are utilized in  PNCC (Power Normalized Cepstral Coefficient) structure, together with the features such as MFCC (Mel-Frequency Cepstral Coefficients) and chroma-based features.
    \item[3.] Proposed HFS technique that uses the SASMA strategy to select the features optimally.
    \item[4.] Contributes the face cascade method and the DSN (Double Sigmoid Normalization) approach to pre-process the face image and the speech signal, respectively.
    \item[5.] Contributing a hybrid detection phase model that adopts models such as SqueezeNet and RNN to train the features and provide an efficient detection outcome.\\\\
    The remainder of the manuscript is organized as follows: The literature review on video deepfake detection approaches is given in Section \ref{literature review}. The methodology of the approach, based on a multimodal approach based on multimodal data fusion, is given in Section \ref{proposed}. The experimental analysis on the proposed model over existing approaches is discussed in Section \ref{result}. Section \ref{conclusion} concludes our manuscript.
\end{enumerate}
\section{Literature Review} \label{literature review}
The literature review is organized into three sections. The first discusses general Deepfake detection methods that analyze inconsistencies over spatial, temporal, and frequency, and GAN-generated images. The second focuses on fusion-based approaches that integrate multi-modal or multi-domain cues for improved robustness. The final section reviews studies related to the proposed model, enabling a focused comparison of methodology, feature representation, and performance.\\\\
Hu et al. \cite{hu2021detecting} analyzed both the temporal aspects and individual frames of compressed Deepfake videos. Frame-level analysis reduces noise from video compression, preventing overfitting, while temporality-level analysis captures temporal relationships among features. By combining both approaches, the method outperforms existing techniques in accurately detecting compressed Deepfakes.
 Chintha et al. \cite{chintha2020recurrent} introduced advanced techniques to detect audio spoofing and visual deepfakes using convolutional latent representations, bidirectional recurrent structures, and entropy-based cost functions, thereby detecting spatial and temporal patterns. 
 Furthermore, the models demonstrated strong generalization across new domains, demonstrating the efficacy of these innovative approaches. Guarnera et al. \cite{guarnera2020fighting} used the expectation maximization algorithm to extract a distinctive fingerprint that captures CT (convolutional Traces) left by GAN (Generative Adversarial Network) during image generation. The CTs demonstrated strong discriminative power, achieving over $98\%$ classification accuracy across deepfakes from GAN architectures, including non-facial images. The method proved reliable and independent of the image semantics, outperforming existing detection techniques and showing resilience to various attacks. Xu et al. \cite{xu2023novel} presented an advanced framework for compressing models in deepfake detection. It involved a two-step process: pre-training a teacher network with labeled samples, followed by transferring knowledge to a lightweight student network. Moreover, a joint distillation loss comprising knowledge distillation, gradient-guided feature distillation losses, and cross-entropy ensured effective knowledge transfer. Additionally, a decayed teaching strategy fine-tuned characteristic distillation to minimize the risk of negative transfer. Rafique et al. \cite{rafique2023deep} proposed an automated method to classify deep-fake images using DL (Deep Learning) and ML (Machine Learning). It performed ELA (Error Level Analysis) to detect modifications, then extracted features with CNNs. The features were classified using SVM (Support Vector Machines) and K-NN (K-Nearest Neighbors) with hyperparameter optimization.\\
 
 Dey et al. \cite{dey2026advancements} demonstrated that GANs and diffusion models are utilized in deepfake medical imaging to create realistic synthetic images across modalities such as MRI, CT, and X-ray. Deep learning, transformer-based architectures, and frequency-domain analysis are come of the deep learning methodologies adopted to identify spatial and spectral inconsistencies. Though the approach faces challenges such as generalization and intrepretability issues. Das et al. \cite{das2026htmdf} proposed HTMDF-DD (Hybrid Triple Modality–based Spatial–Temporal Early Fusion) that integrates spatial feature extraction, temporal modelling and hybrid fusion strategies to capture inconsistencies between frames. spatial artifacts, motion irregularities, and cross-modality correlations are key features of the proposed model. Though the model faces challenges due to limited multimodal datasets, generalization and synchronization issues. Eutamene et al. \cite{eutamene2025integrating} proposed a model that utilizes multimodal fusion framework to integrate perceptual quality analysis with caption-based semantic features to detect deepfakes. Both low-level visual distortions and high-level semantic inconsistencies are captured between video content and captions generated. Though the model faced many challenges such as multimodal alignment and generalization to unseen deepfakes.\\
 
Suganthi et al. \cite{suganthi2022deep} proposed a DL-based deepfake face detection and recognition model. This work detected deepfake face images using the FF-LBPH (Fisher Face algorithm with Local Binary Pattern Histogram) for face recognition. It reduced dimensions with LBPH (Local Binary Pattern Histogram) and applied a DBN (Deep Belief Network) with RBM (Restricted Boltzmann Machines) for classification. Tu et al. \cite{tu2024face} proposed a model that utilizes an optical flow algorithm to extract expression key sequences. Further, a dual-branch network captures both spatial and temporal inconsistencies. Spatial branch and temporal branch incorporate a visual transformer and motion dynamics, respectively, thereby reducing redundancy and improving detection accuracy. However, it relies on precise key-sequence extraction, may struggle with low-quality or occluded faces, and can face limited generalization to unseen manipulation types. \\\\
  
Table \ref{table1} shows the features and challenges of extant works on video-based Deepfake detection. The recurrent convolutional framework \cite{chintha2020recurrent} conducts audio compression to validate the robustness of the system. Though using a larger dataset to acquire the benefits from the parameters of training samples is difficult. Using the CNN \cite{guarnera2020fighting} approach, the proposed scheme attained $93\%$ precision in deepfake detection. 
\begin{center}
\begin{longtable}[!htbp] {|p{1.85cm}|p{2.8cm}|p{3cm}|p{5cm}|}
 \hline
\textbf{Author(s) Citation} & \textbf{Methodology} & \textbf{Features} & \textbf{Challenges} \\
       \hline
       Hu et al. \cite{hu2021detecting}  & CNN & Vigorous to the compressor factor & Need to enhance the system robustness to recognize numerous kinds of facial videos.\\ \hline
       Chintha et al. \cite{chintha2020recurrent}  & Recurrent Convolutional framework & It conducts audio compression to validate robustness of the system & Has to use a larger dataset to acquire the benefits from the parameters of the training sample.\\ \hline
       Guarnera et al. \cite{guarnera2020fighting} & CNN & It attained $93\%$ of precision in detecting fake tasks & It is necessary to examine on a popular real image dataset.\\ \hline
       
       Xu et al. \cite{xu2023novel} & Compression framework based on joint distillation & It attained competitive detection performance through lowering the computation cost. & Need to improve the suggested scheme for characteristic distillation with a heterogeneous framework in transitional layers.\\ \hline 
      
       Dey et al. \cite{dey2026advancements} & GAN- based models, Diffusion models & Spatial artifacts, inconsistencies in the texture, and anomalies in frequency & Cross-domain generalization issues and adversarial robustness.\\ \hline
       
       Das et al. \cite{das2026htmdf} & Hybrid triple-modality framework with early fusion & Spatial and temporal inconsistencies & computational complexity is very high, generalization to unseen deepfakes.\\ \hline
          
        Eutamene et al. \cite{eutamene2025integrating} & Multimodal fusion framework integrating visual and textual representations & Texture and compression artifacts & Limited annotated datasets and generalization to unseen deepfakes. \\ \hline
       Suganthi et al. \cite{suganthi2022deep} & DBN & It offered less examination loss with better validation accuracy & Different classifiers and distance metric measures were not explored and integrated, leading to less precision in the detection of deepfake face images.\\ \hline
       Rafique et al. \cite{rafique2023deep} & DL & It achieved better accuracy with maximum performance & It is essential to examine additional CNN architectures on video-based deepfake detection. \\ \hline  
       Tu et al. \cite{tu2024face} & SVM & Transformer and focal self-attention along with optical flow differences & Generalization to unseen data. Certain conditions, such as occlusion, low resolution, and subtle expression, may deteriorate the performance. \\ \hline
       \caption{Features and Challenges of Existing Deepfake Detection Techniques.} \label{table1}
\vspace{-0.25cm}
   \end{longtable}
    \end{center}

\section{\textbf{Proposed Methodology}} \label{proposed}
Section \ref{proposed} demonstrates the multimodal data fusion approach adopted to detect deepfakes in the architecture, preprocessing phase, feature extraction phase, followed by the feature selection phase in subsections \ref{architecture}, \ref{preprocess2}, \ref{extraction1}, and \ref{featureselection}. Finally, the detection phase has been enclosed, with the detection stage proclaimed in subsection \ref{detection}.  

\subsection{\textbf{Proposed Architecture}} \label{architecture}
\begin{figure}[!htbp]
    \centering
    \includegraphics[width=0.7\linewidth]{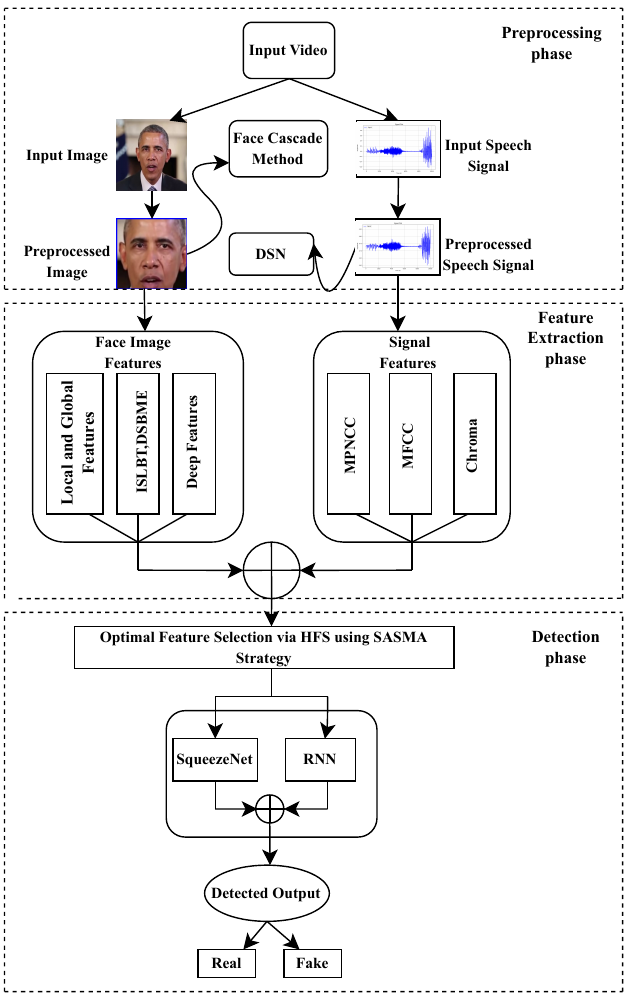}
    \caption{Structural Framework of Proposed Method.}
    \label{figure1}
\end{figure}

In our work, we propose a novel approach for effective deep-fake detection, as illustrated in Figure \ref{figure1}, employing a multimodal data fusion technique that integrates information from two modalities: face images and speech signals extracted from input videos. The initial step involves pre-processing of face images and speech signals to enhance their quality, using the face cascade method for face images and the DSN approach for speech signals. Subsequently, feature extraction is carried out on the pre-processed face images, encompassing local and global features, as well as ISLBT, DSBME, and deep features. Similarly, features such as MPNCC, MFCC, and chroma features are extracted from the preprocessed speech signals. The Hierarchical Feature Selection (HFS) technique is then introduced to select features using the SASMA strategy optimally, and these selected features are balanced. Finally, in the detection phase, models such as SqueezeNet and RNN are used to train the selected features individually, resulting in an efficient and accurate detection output. However, the use of complex and deeper network architectures results in an increased computational cost.

\subsection{\textbf{Preprocessing for both face image and speech signal}} \label{preprocess2}
Consider $V^{inp}$ as the input video, $F^{img}$ and $S^{sgl}$ as the input face image and speech signal acquired from $V^{inp}$ that can be preprocessed at first to improve the quality of the image and signal, respectively. The pre-processing techniques used in this research for both $F^{img}$ and $S^{sgl}$ are given below.

\textbf{Face Image:} Preprocessing of face image ($F^{img}$) is performed using the face cascade method based on the Haar cascade classifier, a type of classifier introduced by Viola and Jones \cite{deshpande2017face}, which efficiently detects and isolates facial regions. This method applies a series of weak classifiers and filters trained using the AdaBoost algorithm, progressively eliminating non-face regions and reducing false positives. The process enhances detection accuracy and computational efficiency, producing a refined facial image ($F^{pre}$) suitable for further processing.

\textbf{Speech signal:} Preprocessing speech signals ($S^{sgl}$) involves the DSN approach \cite{khalifa2013adaptive} to standardize amplitude variations and enhance signal quality. DSN applies two sigmoid functions to scale both positive and negative amplitude values, compressing extremes and ensuring a more uniform amplitude distribution. This reduces clipping and distortion, resulting in a normalized signal ($S^{pre}$) that improves learning stability and model performance in subsequent analysis. The formulation of DSN is expressed as in Eq. \ref{equation1}

\begin{equation}\label{equation1}
    DSN_{k}^{N} =
\begin{cases} 
\frac{1}{1 + \exp \left(-2 \left( \frac{S_{k}^{sgl} - t}{r_1} \right) \right)}, & \text{if } S_{k}^{sgl} < t \\
\frac{1}{1 + \exp \left(-2 \left( \frac{S_{k}^{sgl} - t}{r_2} \right) \right)}, & \text{if } S_{k}^{sgl} \geq t
\end{cases}
\end{equation}
where $r_1$ and $r_2$ indicate the left and right edges of the region, and $t$ indicates the reference operating point. Then, the normalized or preprocessed speech signal is denoted by $S^{pre}$.

\subsection{\textbf{Feature Extraction from face image and speech signal}}\label{extraction1}
Feature extraction is a crucial step in both face image and speech signal processing, to represent relevant information in a compact form for subsequent analysis. In the context of face image processing, feature extraction techniques include local and global features, as well as SLBT (Shape Local Binary Texture), DSBME, and deep features illustrated in subsubsection \ref{extraction2}. For speech signal processing, the features such as MPNCC, MFCC, and chroma are extracted as demonstrated in subsubsection \ref{extractionspeech}. 
\subsubsection{\textbf{Extraction of features from face image}} \label{extraction2}
Face image feature extraction involves capturing distinctive characteristics that can be used for face recognition, classification, or other computer vision tasks. Feature extraction methods can be broadly categorized into local and global features \cite{bosch2011combining}.
\vspace{1mm}

\textbf{Local Features:} Local features focus on capturing detailed information from specific regions or key points within a face. Here, HOG (Histogram of Oriented Gradients) features are extracted. HOG represents the distribution of gradient orientations in an image, often being utilized for face detection. The extracted local feature is denoted by $L^{f}$.

\textbf{Global Features:} Global features aim to represent the overall characteristics of the entire face. These features are often derived from the entire face region. Here, color histogram-based features are extracted. A color histogram is a representation of the distribution of colors in an image. It provides a quantitative summary of the color content, showing the frequency of each color or color bin present in an image. The histogram is constructed by dividing the color space into discrete bins and counting the number of pixels that fall into each bin. The extracted global features is denoted by $G^{f}$.

Thereby, the extracted local and global features are denoted by $LG^{f}$ = [$L^{f}$, $G^{f}$].

\subsubsubsection{\textbf{ISLBT Features (Improved shape local binary texture)}}
ISBLT is an extension of SLBT \cite{lakshmiprabha2012face} that is equivalent to AAM (Active Appearance Model), but the only difference is that SLBT considers LBP texture features rather than intensity values for texture modeling. This feature merges both texture and shape details. The ISLBT-based texture and shape features are extracted from $F^{pre}$. Consider the image $F^{pre}$ and $Z$=[$Z_{1}$, $Z_{2}$,...,$Z_{n}$] as its shape landmark points. Through aligning these landmark points, the shape variations are acquired. Following this, PCA (Principal Component Analysis) is applied to those points. The representation of any shape vector within the training set can be expressed as in Eq. \ref{eq2} and Eq. \ref{eq3}. Here, $W_{v}$ indicates the weight parameter, $E_{v}$ indicates eigen vectors, and $\Bar{Z}$ indicates mean shape.
\begin{equation} \label{eq2}
    Z \approx \Bar{Z}+E_{v}W{v}
\end{equation}
\begin{equation} \label{eq3}
    W_{v}=E^{T}_{v}\left( Z-\bar{Z} \right)
\end{equation}

Texture modeling involves a process where training set images are transformed into a mean shape, creating shape-free patches. The size of these patches significantly influences performance, computational complexity, and processing time. In the context of AAM, the approach involves using the values of direct intensity from the shape-free patches to model texture. On the other hand, SLBT employs LBP on these shape-free patches to retrieve characteristics that are invariant to illumination variations. The utilization of LBP for feature extraction is preferred due to its faster computation and simplicity compared to existing methods, such as Gabor Wavelets.

Assume $\left( y_{c}, z_{c} \right)$ is the center pixel of $3\times3$ window and its intensity value is denoted as $u_{c}$. Their local texture is denoted as $P=p\left( u_{i} \right)$, where $u_{i} \left( i=0,1,2,...7 \right)$ is relevant to the grey values of $8$ neighour pixels. Moreover, the neighbour pixels are thresholded with $u_{c}$ can be represented as $p\left( c\left( u_{0}-u_{c} \right),...,c\left(u_{7}-u_{c}  \right) \right)$ along with their function is expressed as in Eq. \ref{eq4} and the LBP pattern is acquired using Eq. \ref{eq5}.

\begin{equation} \label{eq4}
    c(d) =
\begin{cases} 
1, & d > 0 \\
0, & d \leq 0
\end{cases}
\end{equation}
\begin{equation} \label{eq5}
    Lbp\left( y_{c}, z_{c} \right)=\sum_{i=0}^{7} c\left( u_{i}-u_{c} \right)2^{i}
\end{equation}

The above Eq. \ref{eq5} is improved because LBP is typically defined on a fixed neighborhood size, which can make it sensitive to scale changes. Variations in scale might affect the performance of LBP-based methods in scenarios where objects can appear at different scales. To address this, the improved form of the LBP pattern is acquired using Eq. \ref{eq6}.
\begin{equation} \label{eq6}
    Lbp\left( y_{c}, z_{c} \right)=\left[ \sum_{i=0}^{\frac{n}{2}} c\left( u_{i}-u_{c} \right)*\sigma\left( u_{i} \right) \right]2^{i}+\left[ \sum_{i=\frac{n}{2}+1}^{n} c\left( u_{i}-u_{c} \right) \right]2^{i}
\end{equation}
 where, $u_{i}=\frac{1}{1+e^{-(u_{i})}}*\left| u_{i} \right|$, $u_{i}$ indicates neighbour pixel  value, $u_{c}$ indicates centre pixel value and $\sigma$ \\\\ indicates standard deviation.\\

 Similarly, texture modeling is conducted using PCA according to Eq. \ref{eq7}. Here, $W_{t}$ indicates the weight parameter of texture modeling, $\bar{Z}$ indicates the mean vector, and $E_{v}$ indicates the eigen vectors.

\begin{equation} \label{eq7}
    W_{t}=E^{T}_{v}=\left( Z-\bar{Z} \right)
\end{equation}

Thereby, the combined form of shape and texture constraint vector is acquired using Eq. \ref{eq8}. Also, the shape texture constraint controlling texture, local and global shape is acquired through conducting PCA over the merged constraint, is given in Eq. \ref{eq9}. Here, $D_{v}$ indicates a diagonal matrix of weights for all shape constraints. $VT$ indicates shape texture parameter, $\bar{Z_{vt}}$ indicates mean vector, and $E^{T}_{vt}$ indicates eigen vectors.
\begin{equation} \label{eq8}
    W_{vt} =
\begin{bmatrix}
D_v W_v \\
W_t
\end{bmatrix}
\end{equation}
\begin{equation} \label{eq9}
    VT=E^{T}_{vt}=\left( Z_{vt}-\bar{Z_{vt}} \right)
\end{equation}
 
Thereby, ISLBT-based features extracted from $F^{pre}$ are denoted by $Slbt^{f}$.

\subsubsubsection{\textbf{Diamond Search for block motion estimation}}
Block matching motion estimation \cite{tham1998novel} is used to find a block in the previous video frame that most closely matches a block in the current frame. To make this process faster and more efficient, the Diamond Search (DS) algorithm is used. Unlike traditional methods that rely on rectangular search areas, DS uses a diamond-shaped search pattern that reduces computational cost while maintaining high accuracy in motion detection.
\begin{algorithm}
\caption{\textbf{Diamond Search Block Motion Estimation (DSBME)}}
\textbf{Input:} Current frame and previous frame \\
\textbf{Output:} Motion vector (MV) for each block
\begin{algorithmic}[1]
    \STATE \textbf{Start}
    \STATE Initialize a Large Diamond Search Pattern (LDSP) centered at $(0,0)$.
    \STATE Compute the Mean Absolute Difference (MAD) for all nine points in the LDSP.
    \IF{the point with the smallest MAD value is at the center}
        \STATE Proceed to Step 3.
    \ELSE
        \STATE Shift the diamond to the point with the smallest MAD value and repeat the search.
    \ENDIF
    \STATE Once the minimum MAD is found at the center, switch to a Small Diamond Search Pattern (SDSP) for refinement.
    \STATE Evaluate the four points around the center.
    \STATE Select the position with the lowest MAD as the final Motion Vector (MV).
    \STATE \textbf{End}
\end{algorithmic}
\end{algorithm}

Thereby, DSBME-based features extracted from $F^{pre}$ are denoted by $Dsbme^f$.
\subsubsubsection{\textbf{Deep Features:}}
Deep features \cite{bhatti2021facial} are extracted using VGG16 and ResNet50 from $F^{pre}$ for efficient image detection.\\
 VGG16 is a deep convolutional neural network with $16$ layers, including $13$ convolutional and $3$ fully connected layers. The preprocessed image is passed through several convolutional and pooling layers that use small ($3\times3$) filters to capture fine details. These layers gradually learn more complex patterns, and the final, fully connected layers produce high-level features that are useful for identifying important image characteristics. The features obtained from VGG16 are represented as $Vgg^{f}$.\\ 
 ResNet50 is another powerful deep learning model that uses special residual connections (or skip connections) to help the network learn efficiently and avoid problems such as vanishing gradients. The input first goes through convolution and pooling layers, followed by several residual blocks that extract deeper visual patterns. Instead of using large fully connected layers, ResNet50 applies global average pooling (GAP) to summarize the learned features into a compact form. The resulting features are represented as $Res^{f}$.  
Thus, the entire deep features extracted from $F^{pre}$ can be represented by $D^{f}$=[$Vgg^{f}$, $Res^{f}$]. All the features, such as local and global features, SLBT, DSBME, and deep features, are represented as $Face^{f}$=[$LG^{f}$, $Slbt^{f}$, $Dsbme^{f}$, $D^{f}$].

\subsubsection{\textbf{Extraction of features from speech signal}}\label{extractionspeech}
The concerned subsection illustrates the various features and their purpose of extraction from the speech signal described in subsubsubsection \ref{MPNCC}, \ref{MFCC}, and \ref{chroma}, respectively.
\subsubsubsection{\textbf{MPNCC}} \label{MPNCC}

\begin{figure}[!htbp]
    \centering
    \includegraphics[width=0.9\linewidth]{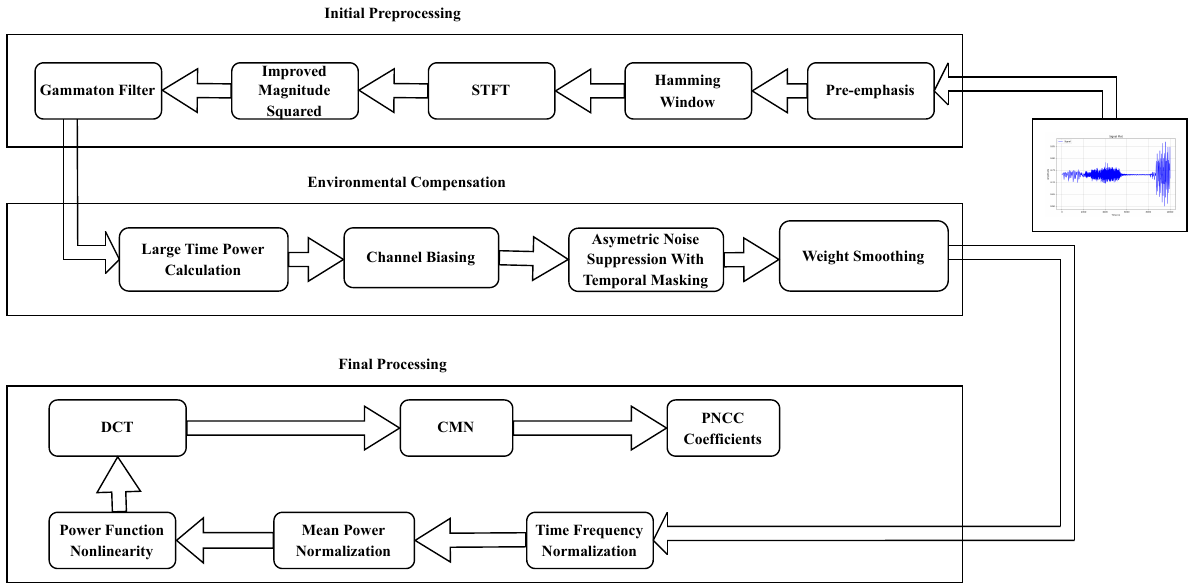}
    \caption{Proposed Structure of MPNCC}
    \label{figure2}
\end{figure}

The MPNCC is an enhanced version of PNCC \cite{thiruvengatanadhan2016speech}, designed to extract robust speech features even in noisy environments. The process involves three main stages: initial processing, environmental compensation, and final feature extraction, illustrated in Figure \ref{figure2}. \\
 
 In initial processing, the speech signal undergoes pre-emphasis to balance the frequency spectrum and is segmented into overlapping frames. Each frame is multiplied by a Hamming window to reduce spectral artifacts. The Short-Time Fourier Transform (STFT) computes the frequency content, followed by an improved magnitude squared (IMS) operation to obtain a more accurate power spectrum. A gammatone filter bank models the human auditory system, and large-time power calculation captures long-term variations in spectral content.\\\\
During environmental compensation, channel bias minimization reduces noise-induced biases across filter channels. Asymmetric noise suppression and temporal masking attenuate background noise, while weighted smoothing and time-frequency normalization ensure consistent spectral representation. Mean power normalization and a power function nonlinearity enhance relevant spectral components.\\\\
In final feature extraction, the Discrete Cosine Transform (DCT) decorrelates features, and Cepstral Mean Normalization (CMN) stabilizes them across frames. The resulting MPNCC features are compact, informative, and robust, making them suitable for automatic speech recognition. The proposed process of MPNCC is explicated in Figure \ref{figure2}.
Key improvements in MPNCC are that IMS reduces noise sensitivity in the power spectrum, and Large-time power calculation captures longer-term spectral trends. And Channel bias minimization corrects noise-induced distortions across channels. The description of the propposed structure of MPNCC is as follows:

\begin{enumerate}
    \item \textbf{IMS (Improved Magnitude Squared):} In the standard PNCC, the power spectral density is acquired through estimating the magnitude square of the resultant frequencies. The major limitation lies in its sensitivity to noise. The squared magnitude can lead to a low signal-to-noise ratio. To address this, IMS adopts a method that multiplies the SNR by the power spectral density.
    \item \textbf{Large Time Power Calculation:} The large time power filter is adopted as a mean low-pass average filter, and this is equivalent to a medium time power filter, but it performs over a higher number of frames. This filter is applied by estimating the running average power $Pw\left[q,l \right]$ for $2Q+1$ successive frames. The large time power calculation is computed using Eq. \ref{eq10}. Here, $l$ indicates the Gammatone channel index, and $q$ indicates the frame index.

\begin{equation} \label{eq10}
    Lt[q,l]=\frac{1}{2Q+1\sum_{q^{'}=q-Q}^{q+Q}}Pw[q^{'},l]
\end{equation}

\item \textbf{Channel Bias Minimizing:} Speech information is concentrated in medium and low frequencies, while noise varies by type. After Gammatone filtering, PSD (Power Spectral Density) values are smoothed using the filter function. Due to differential smoothing in consecutive channels, noise energy spreads more than speech energy, creating a channel bias influenced by noise distribution. This bias varies across channels, especially in environmental noise. The channel bias minimizing is estimated using Eq. \ref{eq11}.
\begin{equation} \label{eq11}
    Lt[q,l]=Lt[l]-b\times min\left( Lt[q,l] \right)
\end{equation}
where b indicates a constant bias factor, and this is based on the coefficient value such that, $0<b<1$.
Thereby, MPNCC-based features extracted from $S^{pre}$ is denoted by $MPNCC^{f}$.
\end{enumerate}
\subsubsubsection{\textbf{MFCC}} \label{MFCC}
MFCCs \cite{valiyavalappil2022emotion} are commonly used in speech recognition methods, especially for feature extraction to recognize speech. To uniquely identify and detect these sounds, a viable approach involves creating a mathematical framework that captures the underlying physical principles of spoken words, which is then encoded using the distinctive characteristics provided by MFCCs. It is worth noting that integrating more features generally enhances accuracy in these applications. Thereby, the MFCC-based feature extracted from $S^{pre}$ is denoted by $Mfcc^{f}$.
\subsubsubsection{\textbf{Chroma}} \label{chroma}
Chroma features \cite{sandhya2020spectral}, also referred to as chromatograms, correspond to $12$ distinct pitch categories commonly found in music. These features, known as pitch class profiles, offer valuable insights into music analysis by categorizing pitches and melodies that follow the equal tempered scale. Chroma-based attributes effectively capture both the chromatic (related to pitch) and melodic elements in music, being sensitive to changes in timbre and accompaniment. They represent the tonal content of music, showcasing the variations and patterns in pitch classes, enabling a thorough assessment of musical compositions. The chroma-based feature extracted from $S^{pre}$ is denoted by $C^{f}$. Thus, the overall features extracted from pre-processed speech signal $S^{pre}$ can be represented as $Speech^{f}$=[$Mpncc^{f}$, $Mfcc^{f}$, $C^{f}$].

\subsection{\textbf{Feature Selection}}\label{featureselection} 
This section details a novel feature selection strategy for the following detection process. From the face feature $Face^{f}$ and speech feature $Speech^{f}$ are extracted. Further, the features are selected using the HFS (Hierarchical Feature Selection) \cite{Mohiuddin2023Hierarchical} approach.
\vspace{2mm}

\begin{figure}[!htbp]
    \centering
    \includegraphics[width=0.6\linewidth]{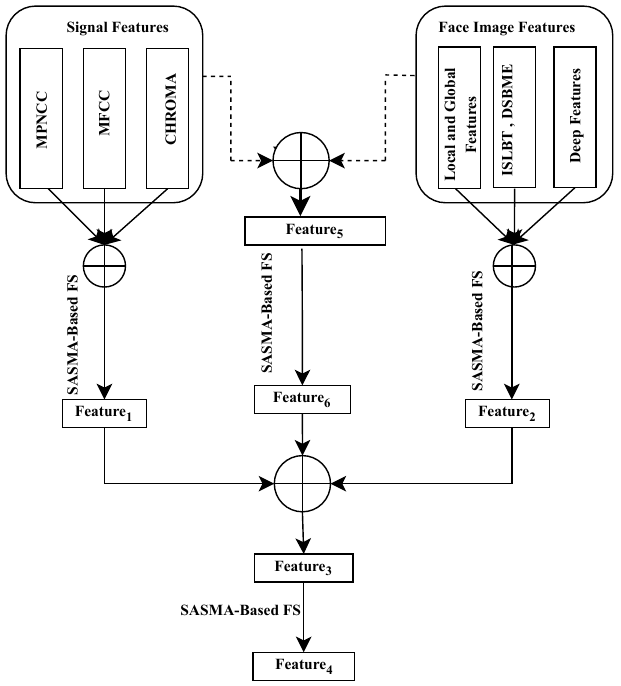}
    \caption{Proposed HFS-Based Feature Selection using SASMA Strategy}
    \label{figure2new}
\end{figure}

\textbf{Hierarchical Feature Selection (HFS):} Figure \ref{figure2new}  illustrates the proposed multimodal feature extraction, fusion, and selection framework designed to detect deepfakes by leveraging both $Speech^{f}$ (audio) and $Face^{f}$ image features. The process consists of two main stages: feature extraction and feature selection, where a hierarchical SASMA-based Feature Selection (FS) method is applied at multiple levels to optimize the feature set. Initially, feature extraction is performed separately for both modalities. \\
The $Speech^{f}$ signal is processed to extract three key feature types: MPNCC, MFCC, and chroma features, which capture frequency and tonal information, further undergoing an initial SASMA-based FS to remove redundant and less relevant features, resulting in $Feature_{1}$. Similarly, $Face^{f}$ image features are extracted, including local and global features, ISLBT, DSBME, and deep features derived from deep learning models. These extracted $Face^{f}$ features also go through SASMA-based FS producing $Feature_{2}$. \\
Following feature extraction, a fusion mechanism is applied to combine information from both modalities. The initially fused representation $Feature_{5}$ integrates the selected $Speech^{f}$ and $Face^{f}$. To further enhance the representation, $Feature_{5}$ undergoes an additional round of SASMA-based FS, resulting in $Feature_{6}$, which contains the most informative features from both sources. The refined multimodal feature set is then processed to form $Feature_{3}$, representing a more structured and optimized representation. \\
Finally, an additional SASMA-based FS step is applied to derive $Feature_{4}$, which serves as the ultimate feature set used for deepfake detection. This approach helps to select the most pre-eminent features, reducing dimensionality while preserving critical information. The integration of $Speech^{f}$ and $Face^{f}$ helps to improve the accuracy of detecting deepfake manipulations effectively. The multi-stage feature selection process ensures that both modalities contribute meaningfully to the final classification model.

\subsubsection{\textbf{SASMA strategy for optimal feature selection}} \label{sasma}
In the HSF technique, the near-optimal features are selected optimally by employing the SASMA strategy. SASMA strategy is a variant of SMA (Slime Mould Algorithm) \cite{li2020slime}, which is adopted by inspiring the searching behavior of food. The SMA draws inspiration from the intricate behaviors of real-world slime molds, particularly the species Physarum polycephalum. Slime molds, despite their simplicity, demonstrate sophisticated and adaptive behavior that provides a compelling basis for algorithmic models. Physarum polycephalum, in particular, exhibits the remarkable capability of discovering the shortest path between food sources in environments. However, they cannot process information or make decisions based on complex reasoning. To address this, the Sandpiper Optimization Algorithm (SOA) \cite{kaur2020sandpiper} is employed by inspiring the collective foraging behavior of sandpipers. Hence, the proposed hybrid optimization is named as Sandpiper Assisted Slime Mould Algorithm (SASMA) strategy. 

\subsubsubsection{\textbf{Solution encoding and objective Function}}
The solution encoded according to the SASMA strategy features includes $Face^{f}$ and $Speech^{f}$. The objective function for the SASMA strategy is the minimization of accuracy, as defined in Eq. \ref{eq12}.

\begin{equation} \label{eq12}
    Of=Min\left[ \frac{1}{Accuracy} \right]
\end{equation}
\subsubsubsection{\textbf{Mathematical Modelling}}

\textbf{Approach Food-Proposed phase:} Slime mould can navigate towards food based on the scent in the air. The mathematical formulation of approaching behavior is defined in Eq. \ref{eq13}. Here, $\overrightarrow{pc}$ indicates linearity from one to zero, $\overrightarrow{S_{b}}$ indicates individual position with the greatest order concentration, $\overrightarrow{pb}$ indicates parameters with the interval [-a, a], $\overrightarrow{\omega}$ indicates slime mould's weight, $t$ indicates current iteration, and $\overrightarrow{S_{A}}$, $\overrightarrow{S_{B}}$ indicates two individual arbitrarily chosen from slime mould.
\begin{equation} \label{eq13}
\overrightarrow{S(t+1)} = 
\begin{cases}
\overrightarrow{S_b(t)} + \overrightarrow{pb} \cdot \left( \overrightarrow{\omega} \cdot \overrightarrow{S_{A}(t)} - \overrightarrow{S_B(t)} \right), & r < k \\
\overrightarrow{pc} \cdot \overrightarrow{S(t)}, & r \geq k
\end{cases}
\end{equation}
The procedure of $p$ is $p=p=tanh\left| A(i)-Bf \right|$; $Bf$ indicates best fitness of current iteration, $A(i)$ indicates fitness of $\overrightarrow{S}$ as per Eq. \ref{eq12} such that $i$ $\epsilon$ 1,2,..n. The procedure of $\overrightarrow{pb}$ is $\overrightarrow{pb}$ =[-a,a]; here $a=arctanh\left( -\left( \frac{t}{t_{max}} \right)+1 \right)$. The procedure of $\overrightarrow{\omega}$ is defined as in Eq. \ref{eq14}. Here, $SmellIndex$ indicates indicates sorted fitness value sequence and their value is $SmellIndex=sort(A)$, $oF$ indicates optimal fitness, $wF$ indicates worst fitness, condition indicates that $A(i)$ ranks first half of the population.
\begin{equation} \label{eq14}
   \overrightarrow{\omega(SmellIndex(i))}= 
   \begin{cases}
       1+r.log\left( \frac{oF-A(i)}{oF-wF} +1\right ) ,& condition\\
        1-r.log\left( \frac{oF-A(i)}{oF-wF} +1\right ) ,& Others
   \end{cases}
\end{equation}

Eq. \ref{eq13} is improved to address the issue of processing information or making decisions based on complex reasoning. The condition in Eq. \ref{eq13} is combined into one formulation as in Eq. \ref{eq15}.
\begin{equation} \label{eq15}
2\overrightarrow{S(t+1)}= \overrightarrow{S_{b}(t)}+\overrightarrow{pb} \cdot\left( \overrightarrow{\omega}\cdot \overrightarrow{S_{A}(t)}-\overrightarrow{S_{B}(t)} \right)+\overrightarrow{pc} \cdot \overrightarrow{S(t)}
\end{equation}
Then, adopt SOA and their updated position of search agent is given in Eq. \ref{eq16} and it is reformulated as in Eq. \ref{eq17}.
\begin{equation} \label{eq16}
    \overrightarrow{S(t+1)}= \left(\overrightarrow{D_{sp}} \times \left( x^{'}+y^{'}+ z^{'} \right)\right) \times \overrightarrow{P_{bst}}(z)
\end{equation}
\begin{equation} \label{eq17}
    \overrightarrow{S(t+1)}=\overrightarrow{D_{sp}} \times x^{'} \times \overrightarrow{P_{bst}}(z)+ \overrightarrow{D_{sp}} \times y^{'} \times \overrightarrow{P_{bst}}(z)+ \overrightarrow{D_{sp}} \times z^{'} \times \overrightarrow{P_{bst}}(z)
\end{equation}
By adding Eq. \ref{eq15} and Eq. \ref{eq17}, the new formulation is given in Eq. \ref{eq18}

\begin{equation} \label{eq18}
    3\overrightarrow{S(t+1)}= \overrightarrow{S_{b}(t)}+\overrightarrow{D_{sp}} \times x^{'} \times \overrightarrow{P_{bst}}(z)+ \overrightarrow{pb} \cdot\left( \overrightarrow{\omega}\cdot \overrightarrow{S_{A}(t)}-\overrightarrow{S_{B}(t)} \right)+\overrightarrow{D_{sp}} \times y^{'} \times \overrightarrow{P_{bst}}(z)+ \overrightarrow{D_{sp}} \times z^{'} \times \overrightarrow{P_{bst}}(z)+ \overrightarrow{pc} \cdot \overrightarrow{S(t)}
\end{equation}
Therefore, 
\begin{equation} \label{eq19}
    \overrightarrow{S(t+1)}=\frac{\overrightarrow{S_{b}(t)}+\overrightarrow{D_{sp}} \times x^{'} \times \overrightarrow{P_{bst}}(z)+ \overrightarrow{pb} \cdot\left( \overrightarrow{\omega}\cdot \overrightarrow{S_{A}(t)}-\overrightarrow{S_{B}(t)} \right)+\overrightarrow{D_{sp}} \times y^{'} \times \overrightarrow{P_{bst}}(z)
    + \overrightarrow{D_{sp}} \times z^{'} \times \overrightarrow{P_{bst}}(z)+ \overrightarrow{pc} \cdot \overrightarrow{S(t)}}{3} 
\end{equation}
Thus, Eq. \ref{eq13} is replaced by using Eq. \ref{eq19}. In addition, $\overrightarrow{S_{A}}$ and $\overrightarrow{S_{B}}$ are improved by employing a cubic chaotic map, and it is expressed as in Eq. \ref{eq20}.

\begin{equation} \label{eq20}
    \overrightarrow{S_{(A,B)+1}}= \gamma \cdot\overrightarrow{S_{(A,B)}} \cdot \left( 1- \overrightarrow{S_{(A,B)}} \right)^{2}
\end{equation}

\textbf{Wrap Food:} 
This section formulates the contraction behavior of venous tissue in slime mould when foraging, establishing a relationship between food concentration and the characteristics of the venous structure. The feedback loop between the vein width and the encountered food concentration is given in Eq. \ref{eq21}. Moreover, the variable $r$ in Eq. \ref{eq14} accounts for the uncertainty in the vein contraction mode. The logarithmic term regulates the numerical fluctuations, ensuring stable contraction frequencies, whereas $condition$ indicates a factor that imitates how slime molds adapt their search patterns based on the food quality. Higher food concentrations increase guiding exploration, the local weight, whereas lower concentrations reduce the region's influence, prompting exploration of other areas. As per the aforementioned principle, the updated position of slime mould is defined as in Eq. \ref{eq14}. Here, $r$ and $rand$ are random values in [$0$,$1$], and $lb$ indicate lower boundaries, and $ub$ indicate upper boundaries.

\begin{equation} \label{eq21}
\overrightarrow{S^{*}} = 
\begin{cases}
rand.\left( ub-lb \right)+lb, & rand < y \\
\overrightarrow{S_b(t)} + \overrightarrow{pb} \cdot \left( \overrightarrow{\omega} \cdot \overrightarrow{S_{A}(t)} - \overrightarrow{S_B(t)} \right), & r < k \\
\overrightarrow{pc} \cdot \overrightarrow{S(t)}, & r \geq k
\end{cases}
\end{equation}

\textbf{Grabble Food:} The slime mould primarily relies on the propagation wave generated by its biological oscillator to alter the flow of cytoplasm in its veins. Moreover, this enables the slime mould to position itself more favourably about the concentration of food. The vector $\overrightarrow{\omega}$ mathematically represents the oscillation frequency of slime mould in proximity to food. This enables the slime mould to expedite its approach to high-quality food sources and decelerate when encountering lower concentrations. By adjusting their speed based on food quality, slime moulds enhance their efficiency in selecting the optimal food source.

Thereby, the features are selected optimally, and the labels of these features are balanced through balancing the minimal count label of features with the size of the maximum label count of features. The selected features are denoted as $Feature_{4}$.

\subsection{\textbf{Detection Phase}} \label{detection}
After the feature selection process, the SqueezeNet-RNN-based deepfake detection (SqN-R-DFD) phase is initiated to determine whether the object in the video is real or fake. The detection model involves two classifiers, namely SqueezeNet and RNN. Both classifiers, SqueezeNet and RNN, take the selected features $Feature_{4}$ as input. These classifiers train these features $Feature_{4}$ to detect the deepfake. The outcome of both classifiers is subjected to take the average, and the analyzed results provide $'0'$ and $'1'$ as the outcome, where $'0'$ indicates fake and $'1'$ indicates real.

\subsubsection{\textbf{SqueezeNet}}
SqueezeNet \cite{bernardo2022modified} is a lightweight neural network designed for image classification that keeps accuracy high while using fewer parameters. Its main building block is the Fire Module, which first “squeezes” the input using a $1\times1$ convolution to reduce channels, then “expands” it with a mix of $1\times1$ and $3\times3$ convolutions to capture more features. The network begins with a convolution and pooling layer, passes through several Fire Modules with occasional pooling and dropout, and ends with global average pooling and a softmax layer for classification. ReLU activation is applied throughout to introduce non-linearity. The final output from SqueezeNet provides the image-based predictions.
Thereby, the outcome from the SqueezeNet model is denoted by $Sqz^{out}$. 

\subsubsection{\textbf{RNN}}
RNNs \cite{traver2022egocentric} are a type of neural network designed to work with sequential data such as speech or text. They remember previous information through a hidden state, which helps the network understand patterns over time. The model is trained to classify inputs using a standard sparse categorical cross-entropy loss function, and an Adam optimizer ensures faster and more stable learning. Finally, the predictions from multiple RNN streams are combined to decide whether an input is ‘Real’ or ‘Fake’. Thereby, the outcome from the RNN model is denoted by $RNN^{out}$.

\begin{equation} \label{eq22}
    y_{t}=\psi_{1}h_{t}+\beta_{y}
\end{equation}

By taking an average of both outcomes, such as $Sqz^{out}$ and $RNN^{out}$, it provides the detected output as 'real' or 'fake'.

\section{Experimental Results and Findings} \label{result}
This section describes the experimentation followed by performance evaluations carried out on the World Leader Dataset (WLDR) \cite{agarwal2019protecting} and the DeepfakeTIMIT dataset \cite{deepfaketimit}. The description of the dataset is illustrated in subsections \ref{dataset1} and \ref{dataset2}, respectively.
\subsection{\textbf{Simulation Procedure}}
The proposed deepfake detection with multimodalities was simulated using PYTHON, specifically version 3.7. The processor employed was Intel(R) Core(TM) I5-4210U CPU @ 1.70 GHz, and the installed RAM size was 8.00 GB. Furthermore, the deepfake detection was analyzed using WLDR (World Leader Dataset) \cite{agarwal2019protecting}, and the DeepfakeTIMIT Dataset \cite{deepfaketimit}.

\subsection{\textbf{Dataset1 Desription (World Leader Dataset)}}\label{dataset1}
This compilation features extensive footage of authentic videos showcasing five U.S. political figures, their impressionists in the political realm, and deepfake videos employing a face-swapping approach to depict all political figures alongside their respective impersonators.
For our investigation, we employed a total of $1710$ video samples with two diverse classes, such as Fake (Label $0$) and Real (Label $1$). Each class consists of $855$ instances.

\subsection{\textbf{Dataset2 Description (DeepfakeTIMIT Dataset)}} \label{dataset2}
In this dataset, they had manually picked $16$ similarly looking pairs of people from the VidTIMIT database. For each of 32 subjects, they trained two diverse strategies: a lower quality ($LQ$) with $64 \times 64$ input/output size model, and a higher quality ($HQ$) with $128 \times 128$ size model. Subsequently, there were $10$ videos per person in the VidTIMIT database; hence, they created $320$ videos consistent with a separate version, resulting in $620$ total videos with faces swapped. Regarding audio, they sustained the original audio track of all videos, i.e., no experimentation was done to the audio channel.
In this research, we employed a dataset containing $640$ data, divided into two categories such as Fake (Label $'0'$) and Real (Label $'1'$). Further, each class contains $320$ instances.

\subsection{\textbf{Performance Analysis}}
A comprehensive comparison was conducted between the SqN-R-DFD method and established techniques for multimodal deepfake detection. The assessment included various performance metrics such as sensitivity, NPV, specificity, F-measure, FNR, precision, FPR, MCC, and accuracy. Furthermore, ablation tests, K-fold cross-validation, ROC curves, training and test loss, and statistical analyses were conducted. The SqN-R-DFD approach is compared with state-of-the-art methods such as DBN \cite{suganthi2022deep}, SVM \cite{tu2024face}, and traditional classifiers, including SqueezeNet, RNN, Bi-LSTM, LinkNet, and DCNN. Further, the SqN-R-DFD method and conventional approaches were evaluated using the World Leader, DeepfakeTIMIT datasets. 

\begin{figure}[h!]
    \centering
    \begin{subfigure}{0.13\textwidth}
        \includegraphics[width=\linewidth]{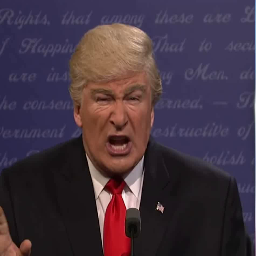}
    \end{subfigure}\hfill
    \begin{subfigure}{0.13\textwidth}
        \includegraphics[width=\linewidth]{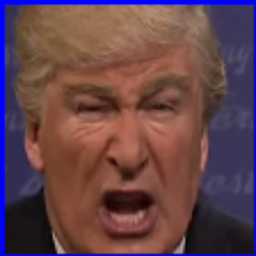}
    \end{subfigure}\hfill
    \begin{subfigure}{0.13\textwidth}
        \includegraphics[width=\linewidth]{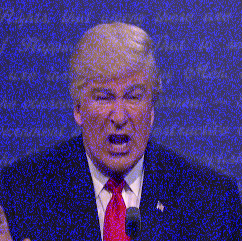}
    \end{subfigure}\hfill
    \begin{subfigure}{0.13\textwidth}
        \includegraphics[width=\linewidth]{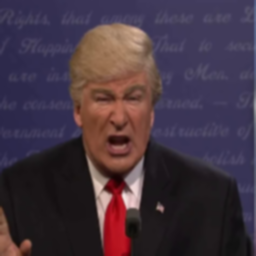}
    \end{subfigure}\hfill
    \begin{subfigure}{0.13\textwidth}
        \includegraphics[width=\linewidth]{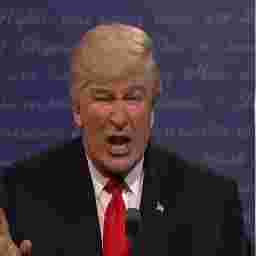}
    \end{subfigure}\hfill
    \begin{subfigure}{0.13\textwidth}
        \includegraphics[width=\linewidth]{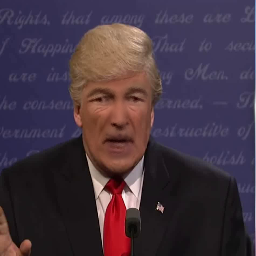}
    \end{subfigure}\hfill
    \begin{subfigure}{0.13\textwidth}
        \includegraphics[width=\linewidth]{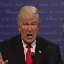}
    \end{subfigure}
    
     \begin{subfigure}{0.13\textwidth}
        \includegraphics[width=\linewidth]{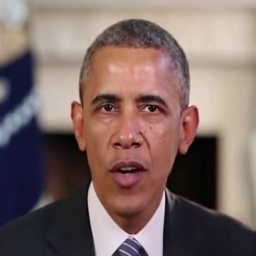}
        \caption{}
    \end{subfigure}\hfill
    \begin{subfigure}{0.13\textwidth}
        \includegraphics[width=\linewidth]{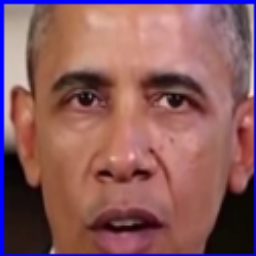}
        \caption{}
    \end{subfigure}\hfill
    \begin{subfigure}{0.13\textwidth}
        \includegraphics[width=\linewidth]{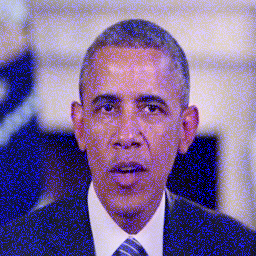}
        \caption{}
    \end{subfigure}\hfill
    \begin{subfigure}{0.13\textwidth}
        \includegraphics[width=\linewidth]{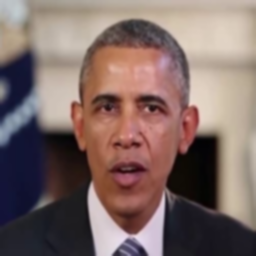}
        \caption{}
    \end{subfigure}\hfill
    \begin{subfigure}{0.13\textwidth}
        \includegraphics[width=\linewidth]{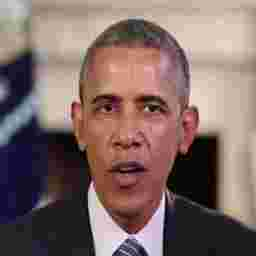}
        \caption{}
    \end{subfigure}\hfill
    \begin{subfigure}{0.13\textwidth}
        \includegraphics[width=\linewidth]{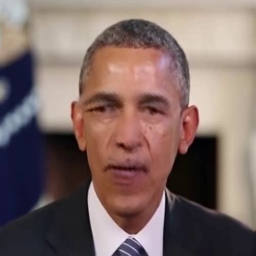}
        \caption{}
    \end{subfigure}\hfill
    \begin{subfigure}{0.13\textwidth}
        \includegraphics[width=\linewidth]{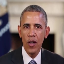}
        \caption{}
    \end{subfigure}
    \caption{Images for Deepfake Detection using Dataset1 (a) Original Images (b) Face Cascade using Pre-processed Images (c) Noise Case Images (d) Blurred Case Images (e) Compressed Case images (f) Face Swapped Case images and (g) Resized Case Images}
    \label{fig4}
\end{figure}

\begin{figure}[!htbp]
    \centering
    \begin{subfigure}{0.13\textwidth}
            \includegraphics[width=\linewidth]{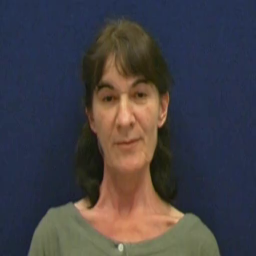}
    \end{subfigure}\hfill
    \begin{subfigure}{0.13\textwidth}
        \includegraphics[width=\linewidth]{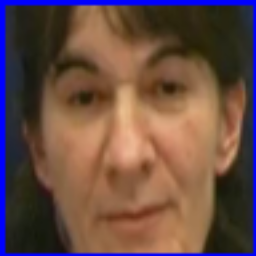}
    \end{subfigure}\hfill
    \begin{subfigure}{0.13\textwidth}
        \includegraphics[width=\linewidth]{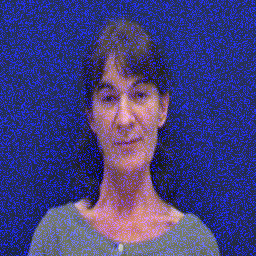}
    \end{subfigure}\hfill
    \begin{subfigure}{0.13\textwidth}
        \includegraphics[width=\linewidth]{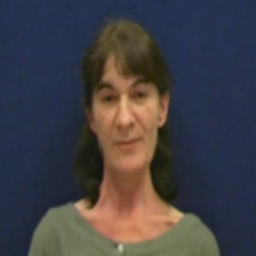}
    \end{subfigure}\hfill
    \begin{subfigure}{0.13\textwidth}
        \includegraphics[width=\linewidth]{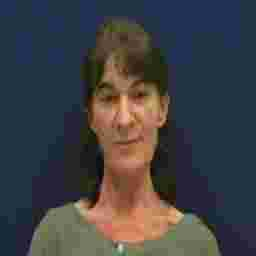}
    \end{subfigure}\hfill
    \begin{subfigure}{0.13\textwidth}
        \includegraphics[width=\linewidth]{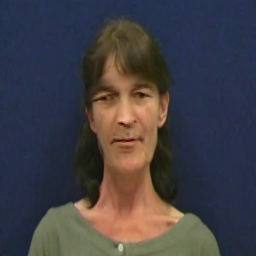}
    \end{subfigure}\hfill
    \begin{subfigure}{0.13\textwidth}
        \includegraphics[width=\linewidth]{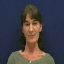}
    \end{subfigure}
     \begin{subfigure}{0.13\textwidth}
        \includegraphics[width=\linewidth]{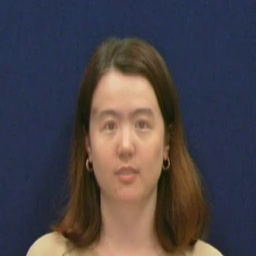}
    \end{subfigure}\hfill
    \begin{subfigure}{0.13\textwidth}
        \includegraphics[width=\linewidth]{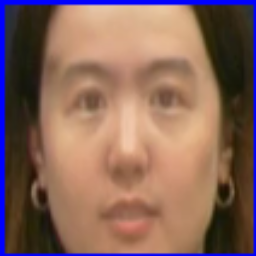}
    \end{subfigure}\hfill
    \begin{subfigure}{0.13\textwidth}
        \includegraphics[width=\linewidth]{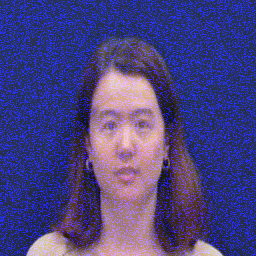}
    \end{subfigure}\hfill
    \begin{subfigure}{0.13\textwidth}
        \includegraphics[width=\linewidth]{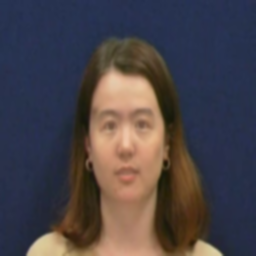}
    \end{subfigure}\hfill
    \begin{subfigure}{0.13\textwidth}
        \includegraphics[width=\linewidth]{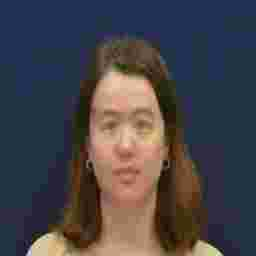}
    \end{subfigure}\hfill
    \begin{subfigure}{0.13\textwidth}
        \includegraphics[width=\linewidth]{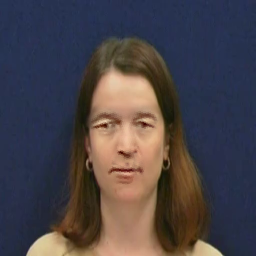}
    \end{subfigure}\hfill
    \begin{subfigure}{0.13\textwidth}
        \includegraphics[width=\linewidth]{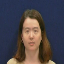}
    \end{subfigure}
    \caption{Images for Deepfake Detection using Dataset2 (a) Original Images (b) Face Cascade using Pre-processed Images (c) Noise Case Images (d) Blurred Case Images (e) Compressed Case images (f) Face Swapped Case images and (g) Resized Case Images}
    \label{fig5}
\end{figure}

Figures \ref{fig4} and \ref{fig5} display the original images alongside their respective face cascade outcomes using pre-processed images. Also, it includes noise case images (with Gaussian Noise applied), blurred case images (processed using the Gaussian method), compressed case images (compression quality = $10$), face-swapped case images, and resized case images (scaled to a $64\times64$ resolution) employing the WLDR dataset (Dataset1) \cite{agarwal2019protecting} and DeepfakeTIMIT dataset (Dataset2) \cite{deepfaketimit} each. 

\subsection{\textbf{Comparative Assessment for Dataset1}} \label{completeassessment1}
The following section demonstrates the comparative analysis among conventional approaches against seven deformation cases such as original, noise, blurred, compressed, wav2lip, faceswapped, and resize for WLDR dataset \cite{agarwal2019protecting} in terms of accuracy, F-measure, analysis based on the ROC curve, and training-testing loss curve discussed in subsubsection \ref{comparative1}, \ref{roc1} and \ref{training1} respectively. Further, in this study, we integrate seven different cases allowing us to perform a comprehensive analysis including statistical analysis, an ablation study to analyze the impact of each component, cross-dataset analysis, and K-fold cross-validation for model robustness, demonstrated in subsection \ref{featureselectionwldr}, \ref{ablation1}, \ref{crossdataset1}, and \ref{kfold1}, respectively.
\subsubsection{\textbf{Comparative Analysis}} \label{comparative1}
The performance of the proposed SqN-R-DFD model was evaluated across multiple scenarios: original, noisy, blurred, compressed, Wave2Lip, face-swapped, and resized cases. Accuracy and F-measure metrics were used to compare the method against conventional approaches, including SqueezeNet, RNN, Bi-LSTM, LinkNet, DCNN, DBN \cite{suganthi2022deep}, and SVM \cite{tu2024face}.

\begin{figure}[!htbp]
    \centering
    \begin{subfigure}[t]{0.45\textwidth}
        \includegraphics[width=\linewidth]{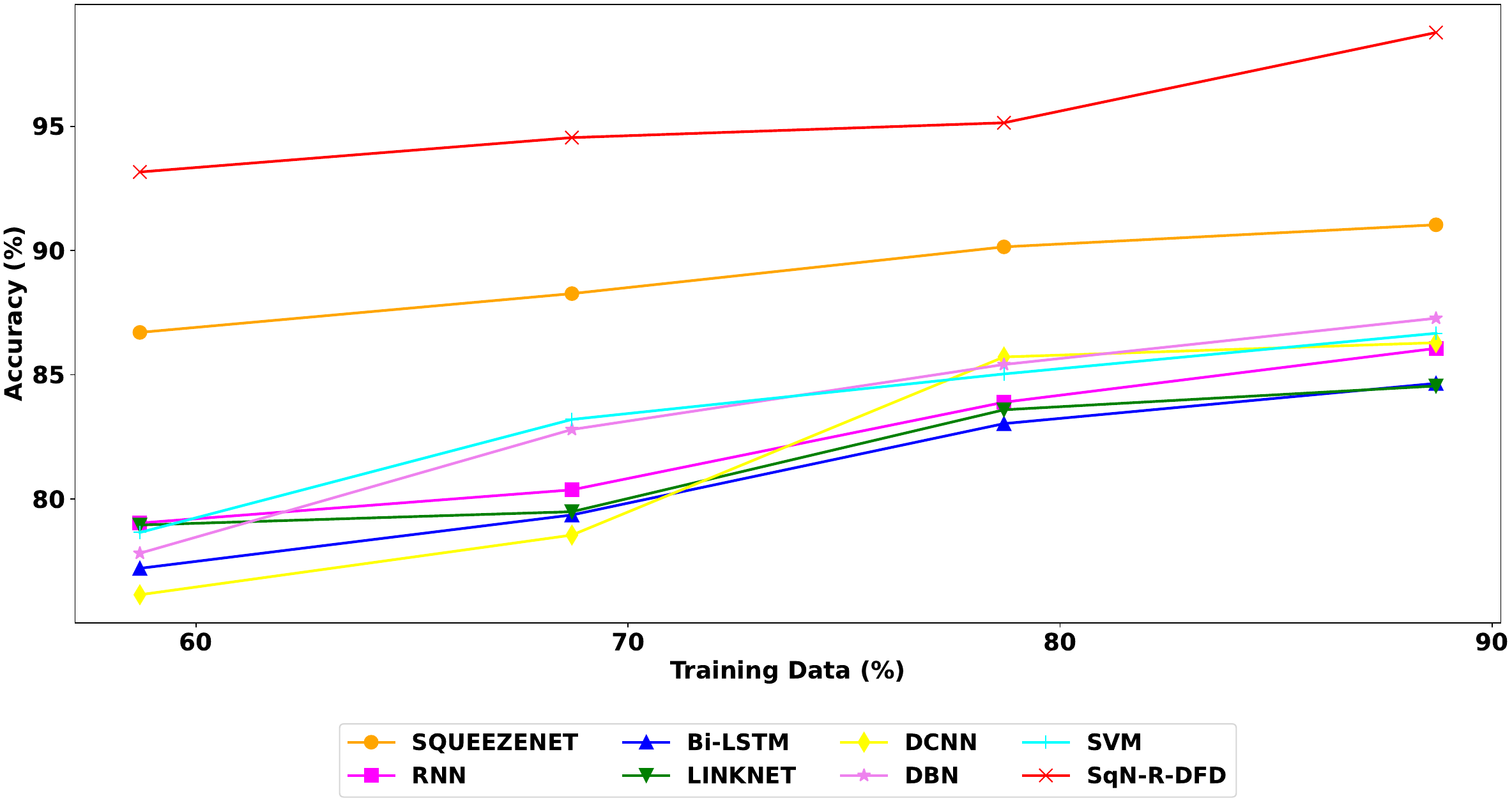}
        \caption{Accuracy for Original Case using Dataset1
        \label{figaccuracy_a}}
    \end{subfigure}%
    \hfill
       \begin{subfigure}[t]{0.45\textwidth}
        \includegraphics[width=\linewidth]{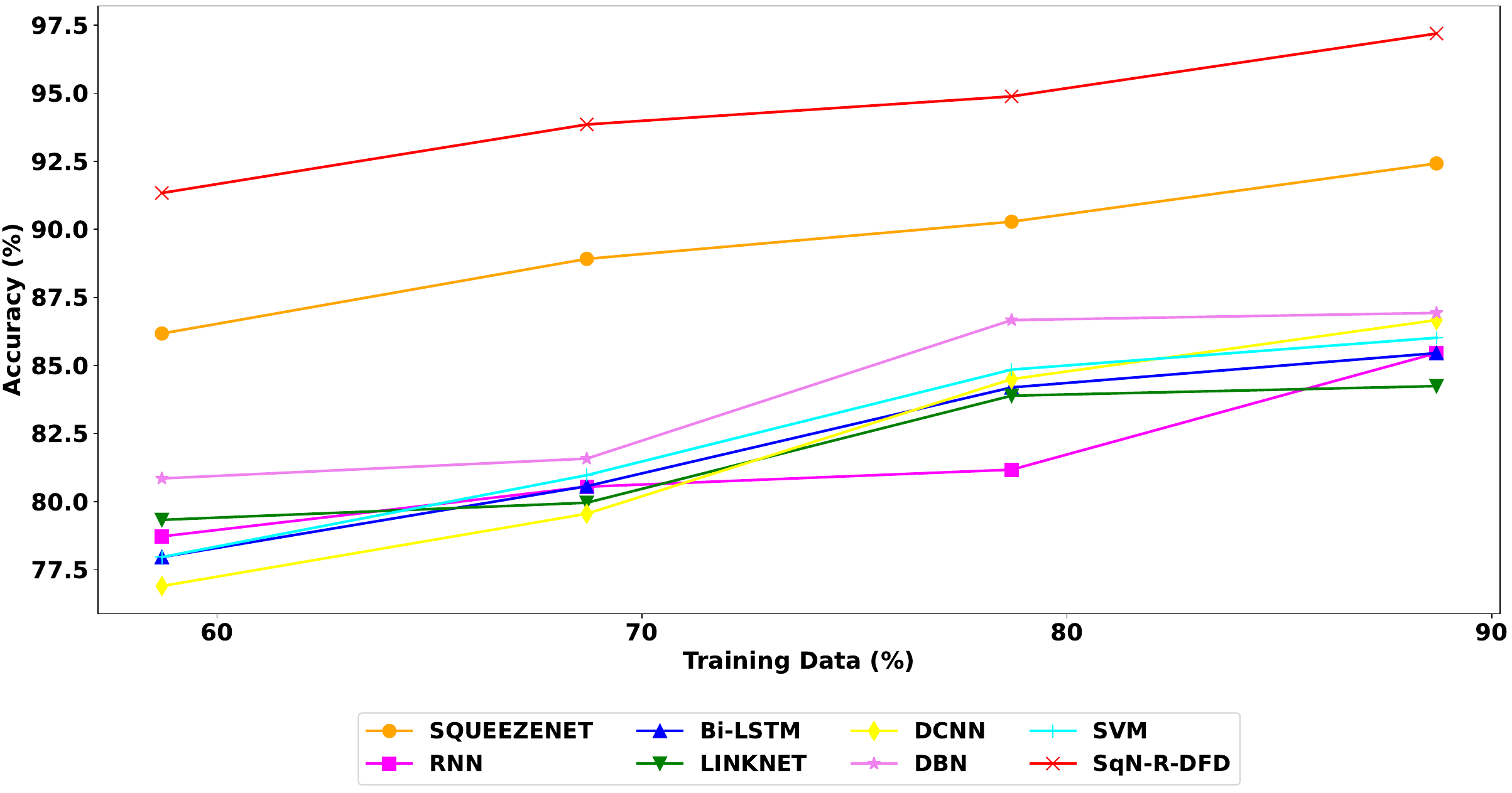}
        \caption{Accuracy for Noise Case using Dataset1\label{figaccuracy_b}}        
    \end{subfigure}

    \begin{subfigure}{0.45\textwidth}
        \includegraphics[width=\linewidth]{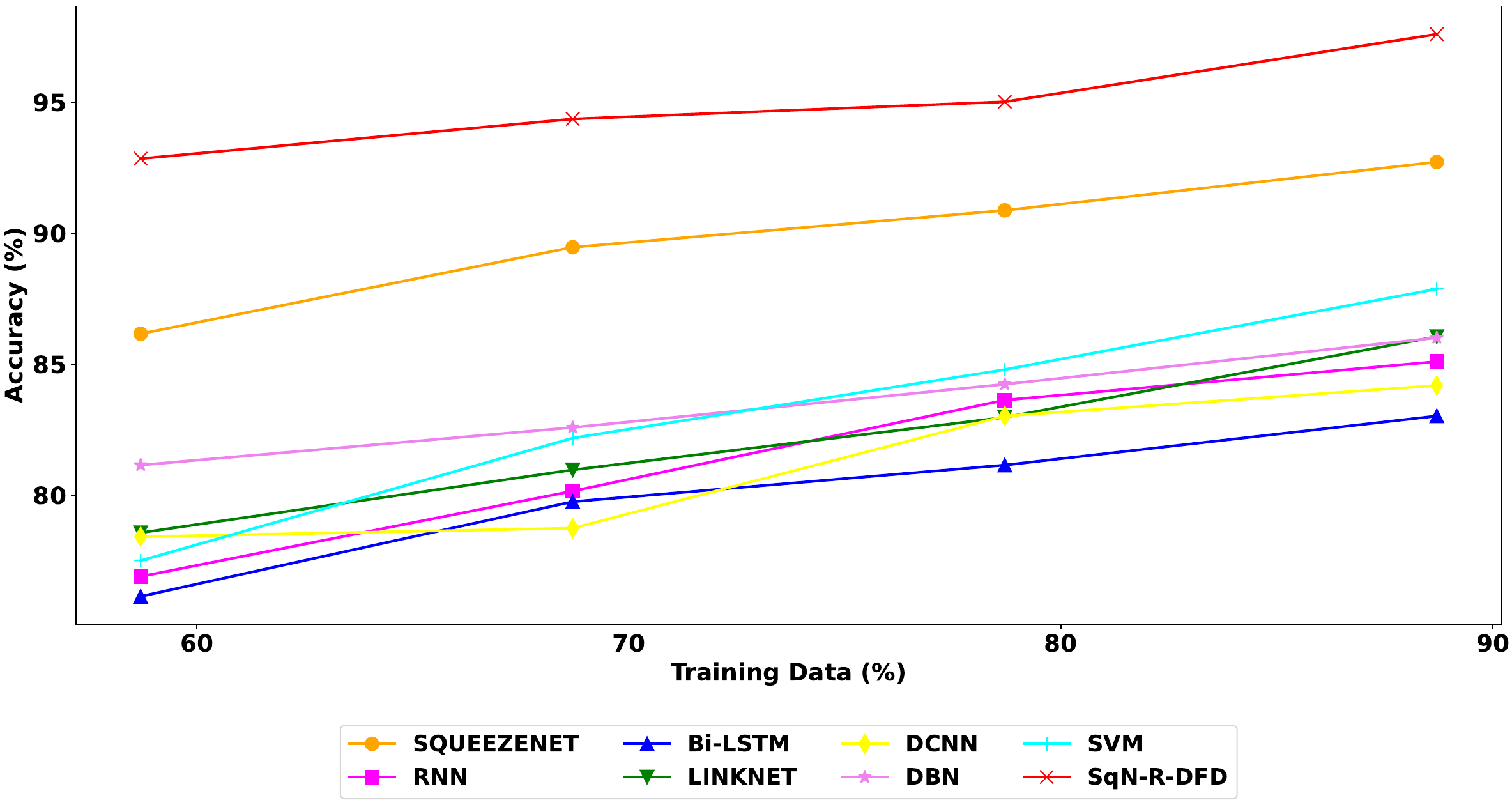}
        \caption{Accuracy for Blurred Case using Dataset1 \label{figaccuracy_c}}
    \end{subfigure} 
    \hfill
    \begin{subfigure}{0.45\textwidth}
        \includegraphics[width=\linewidth]{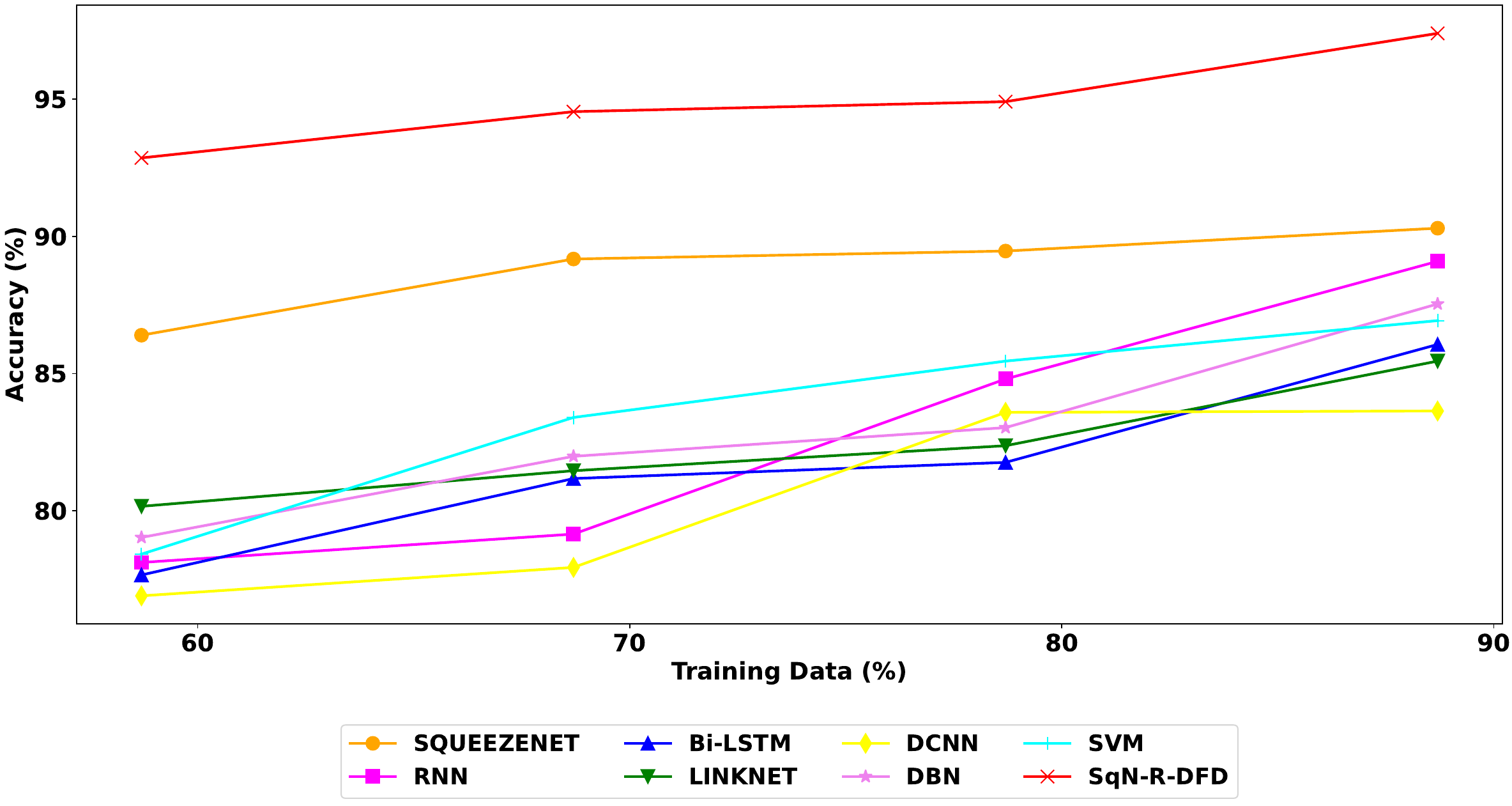}
        \caption{Accuracy for Compressed Case using Dataset1 \label{figaccuracy_d}}
    \end{subfigure} 
    \begin{subfigure}{0.45\textwidth}
        \includegraphics[width=\linewidth]{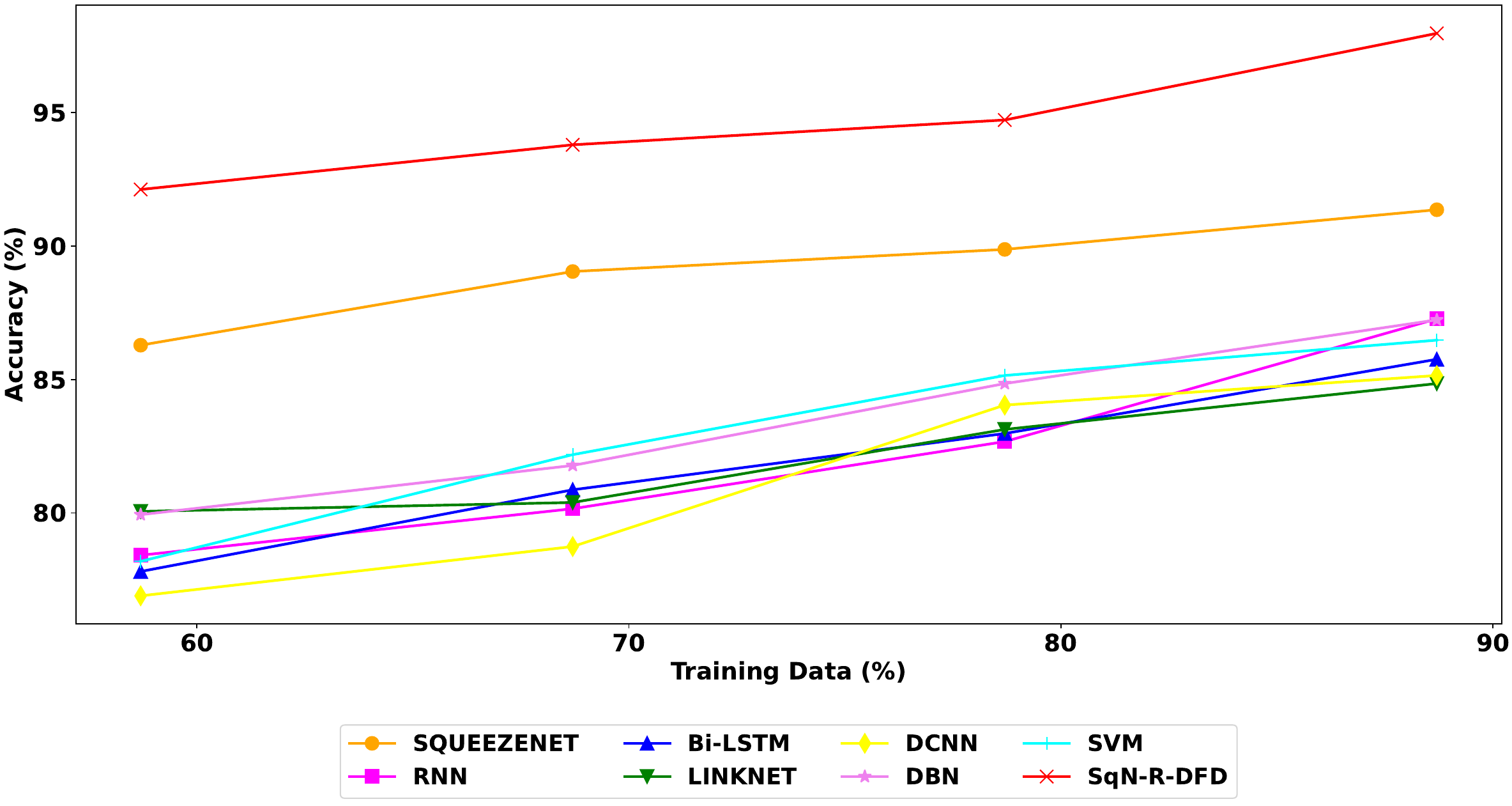}
        \caption{Accuracy for Wav2lip Case using Dataset1 \label{figaccuracy_e}}
    \end{subfigure}     
    \hfill
     \begin{subfigure}{0.45\textwidth}
        \includegraphics[width=\linewidth]{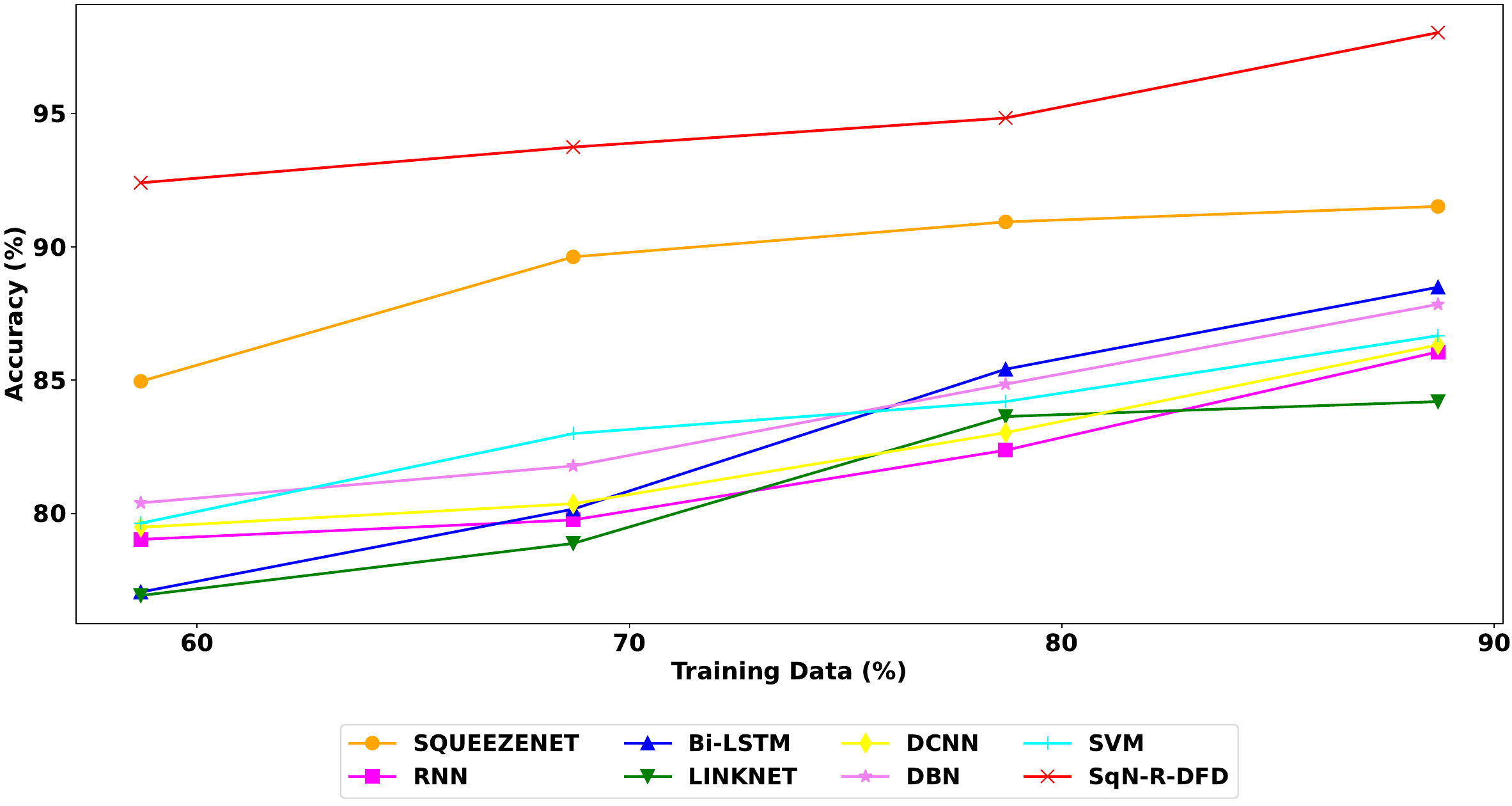}
        \caption{Accuracy for Faceswapped Case using Dataset1 \label{figaccuracy_f}}
    \end{subfigure}
         \begin{subfigure}{0.45\textwidth}
        \includegraphics[width=\linewidth]{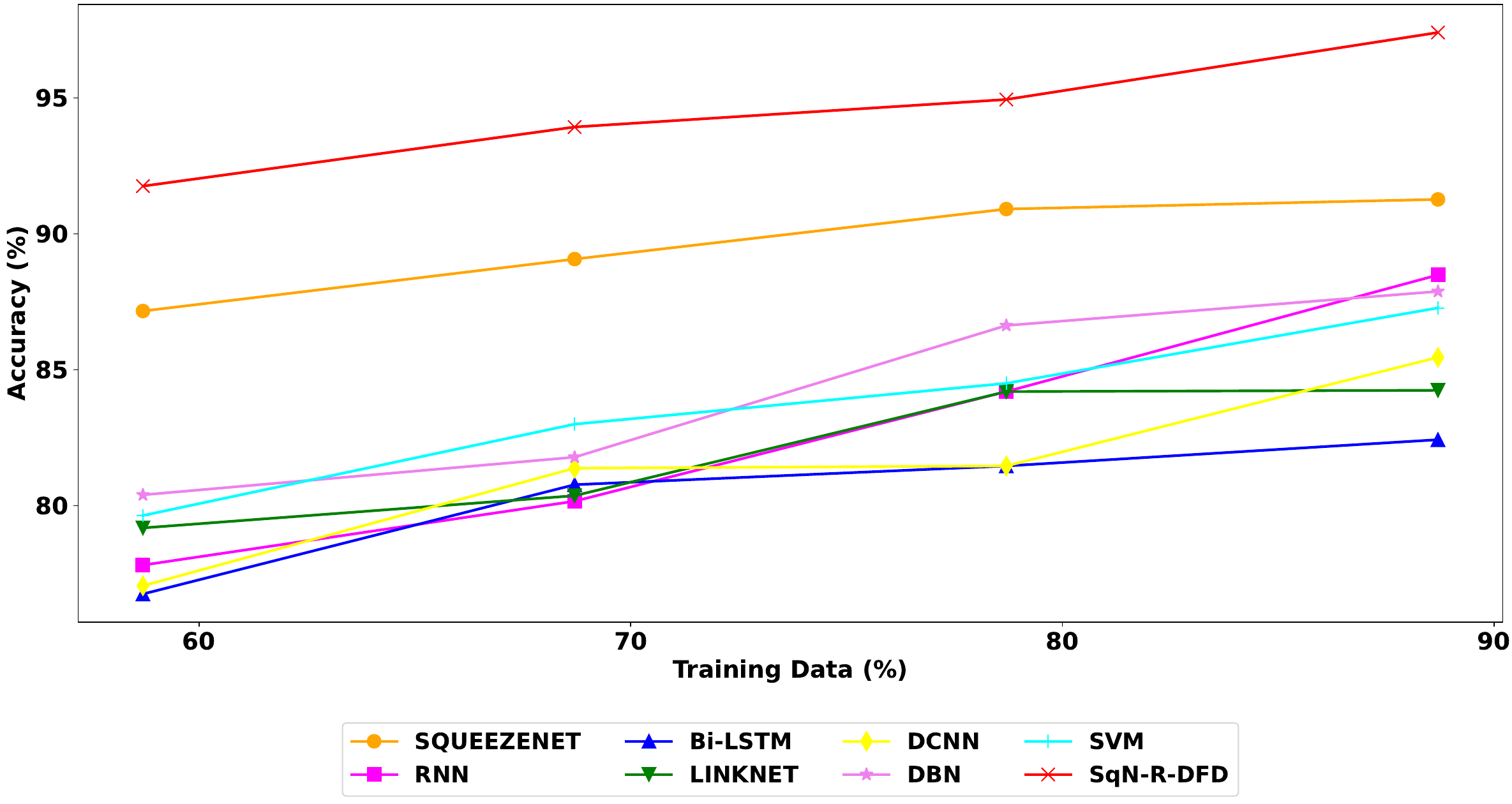}
        \caption{Accuracy for Resized Case using Dataset1 \label{figaccuracy_g}}
    \end{subfigure}
    \caption{Comparison on SqN-R-DFD Versus Conventional Methods for Dataset1 (a) Accuracy for Original case (b) Accuracy for Noise Case (c) Accuracy for Blurred case (d) Accuracy for Compressed Case  (e) Accuracy for Wav2lip Case (f) Accuracy for Faceswapped Case (g) Accuracy for Resized Case}
    \label{fig14accuracy}
    \end{figure}

\textbf{ Accuracy:} Upon analyzing Figure \ref{figaccuracy_b}, which is a noise case, the SqN-R-DFD consistently outperforms the traditional approaches in terms of F-measure and Accuracy. At $90\%$ training data, the SqN-R-DFD model accomplished the peak accuracy rate of $98.765$, signifying enhanced performance in detecting deepfakes. In contrast, the conventional methods established moderately lower accuracy values. Further, for all the other cases also the proposed SqN-R-DFD model consistently outperforms all the existing methods. Thus, the HFS approach is introduced to optimally select features using the SASMA model, and these selected features are balanced.

\begin{figure}[!htbp]
    \centering
    \begin{subfigure}[t]{0.45\textwidth}
        \includegraphics[width=\linewidth]{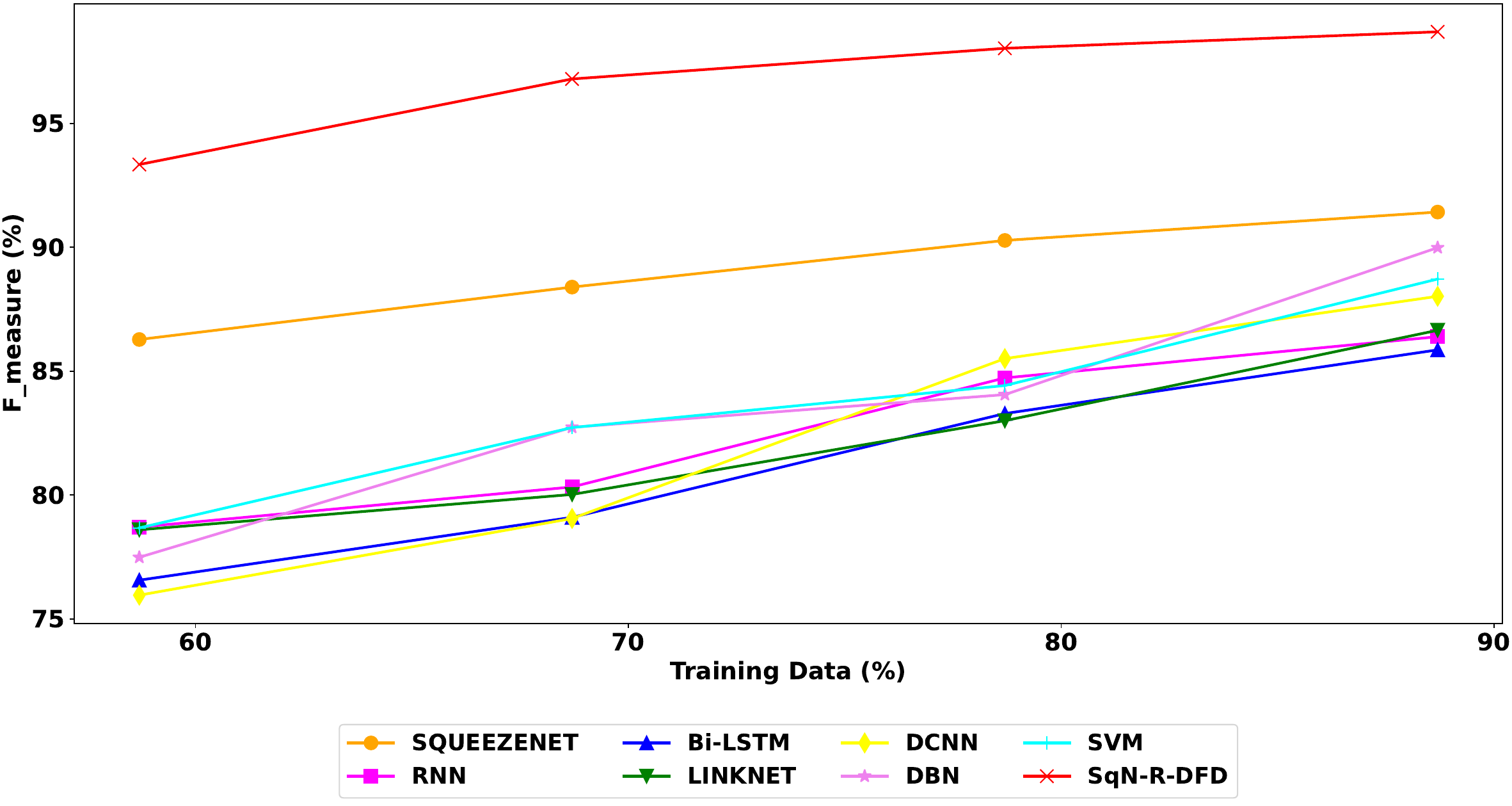}
        \caption{F-measure for Original Case using Dataset1 \label{fig14fmeasure_a}}
    \end{subfigure}%
    \hfill
    \begin{subfigure}[t]{0.45\textwidth}
        \includegraphics[width=\linewidth]{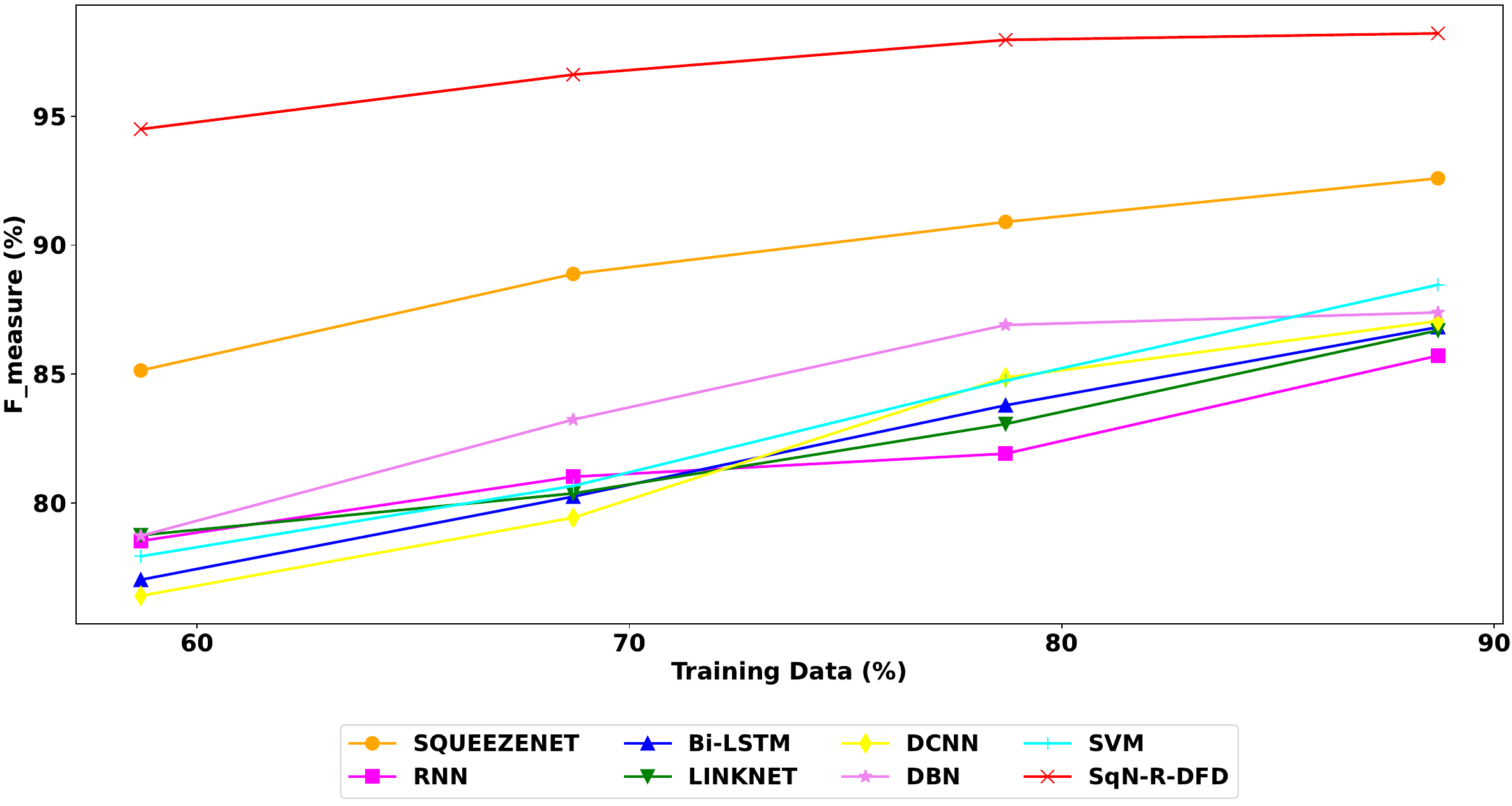}
        \caption{F-measure for Noise Case using Dataset1 \label{fig14fmeasure_b}}
    \end{subfigure}
    \begin{subfigure}{0.45\textwidth}
        \includegraphics[width=\linewidth]{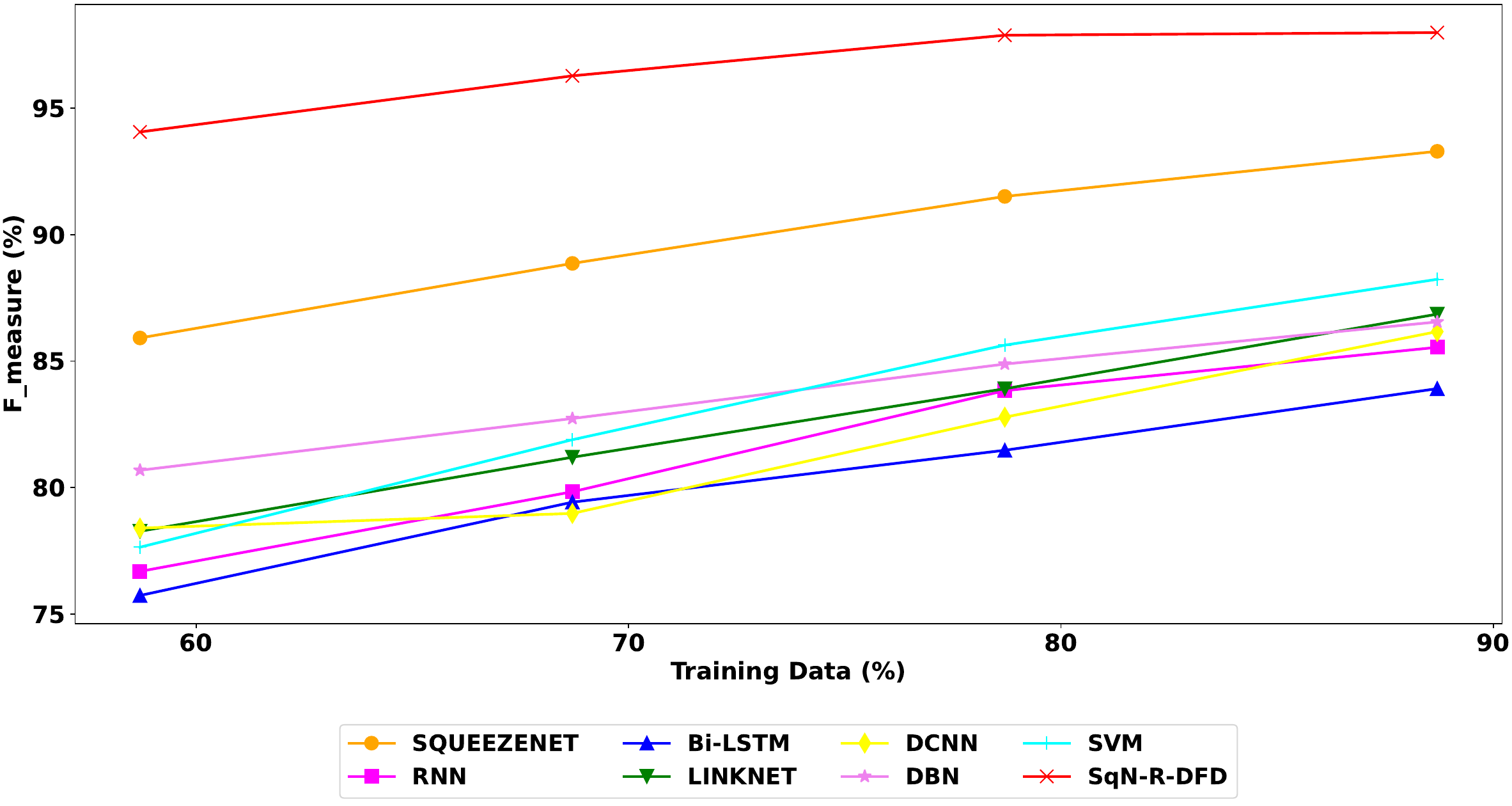}
        \caption{F-measure for Blurred Case using Dataset1 \label{fig14fmeasure_c}}
    \end{subfigure}%
    \hfill
    \begin{subfigure}{0.45\textwidth}
        \includegraphics[width=\linewidth]{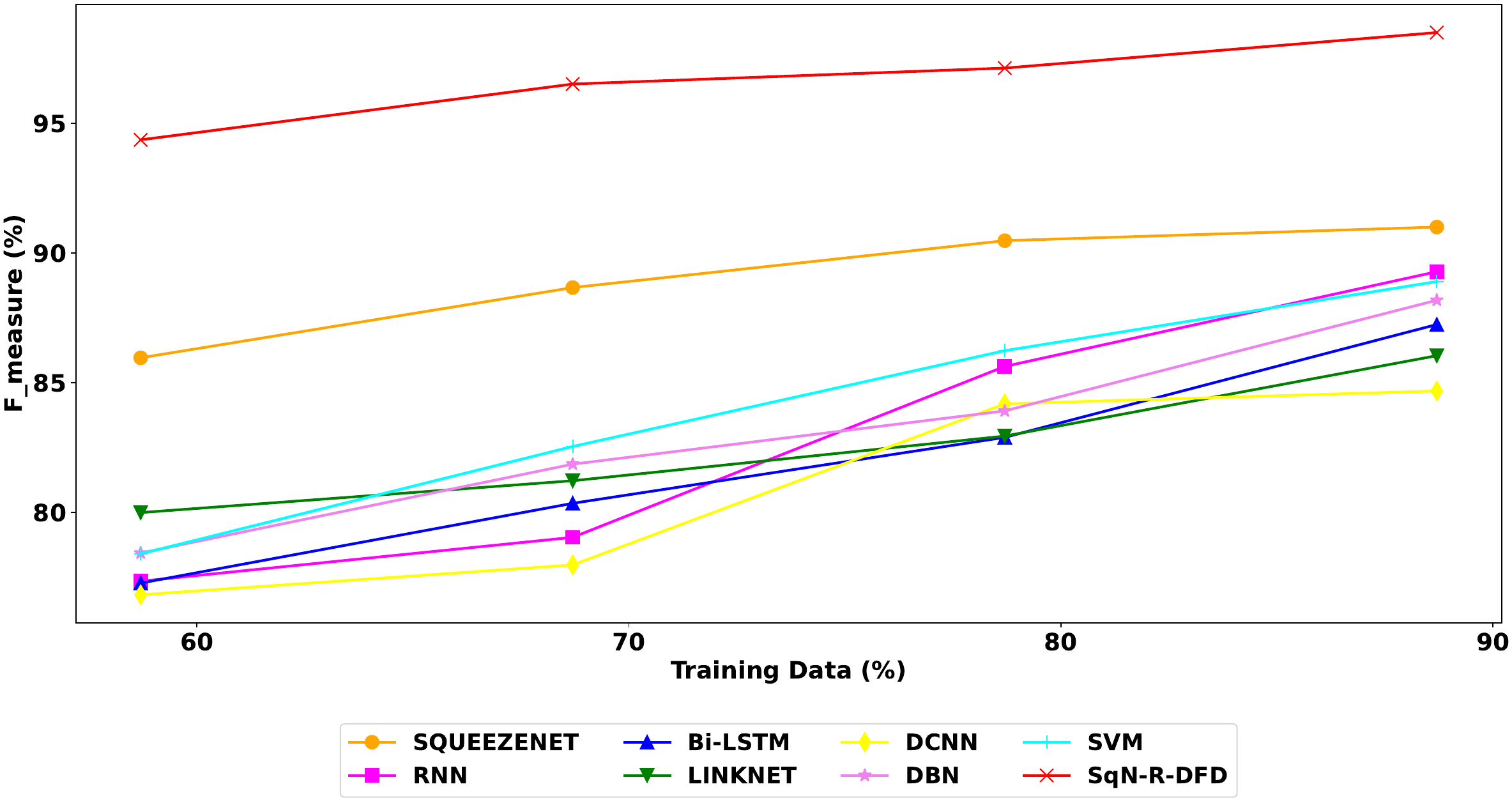}
        \caption{F-measure for Compressed Case using Dataset1 \label{fig14fmeasure_d}}
    \end{subfigure}
             
    \begin{subfigure}{0.45\textwidth}
        \includegraphics[width=\linewidth]{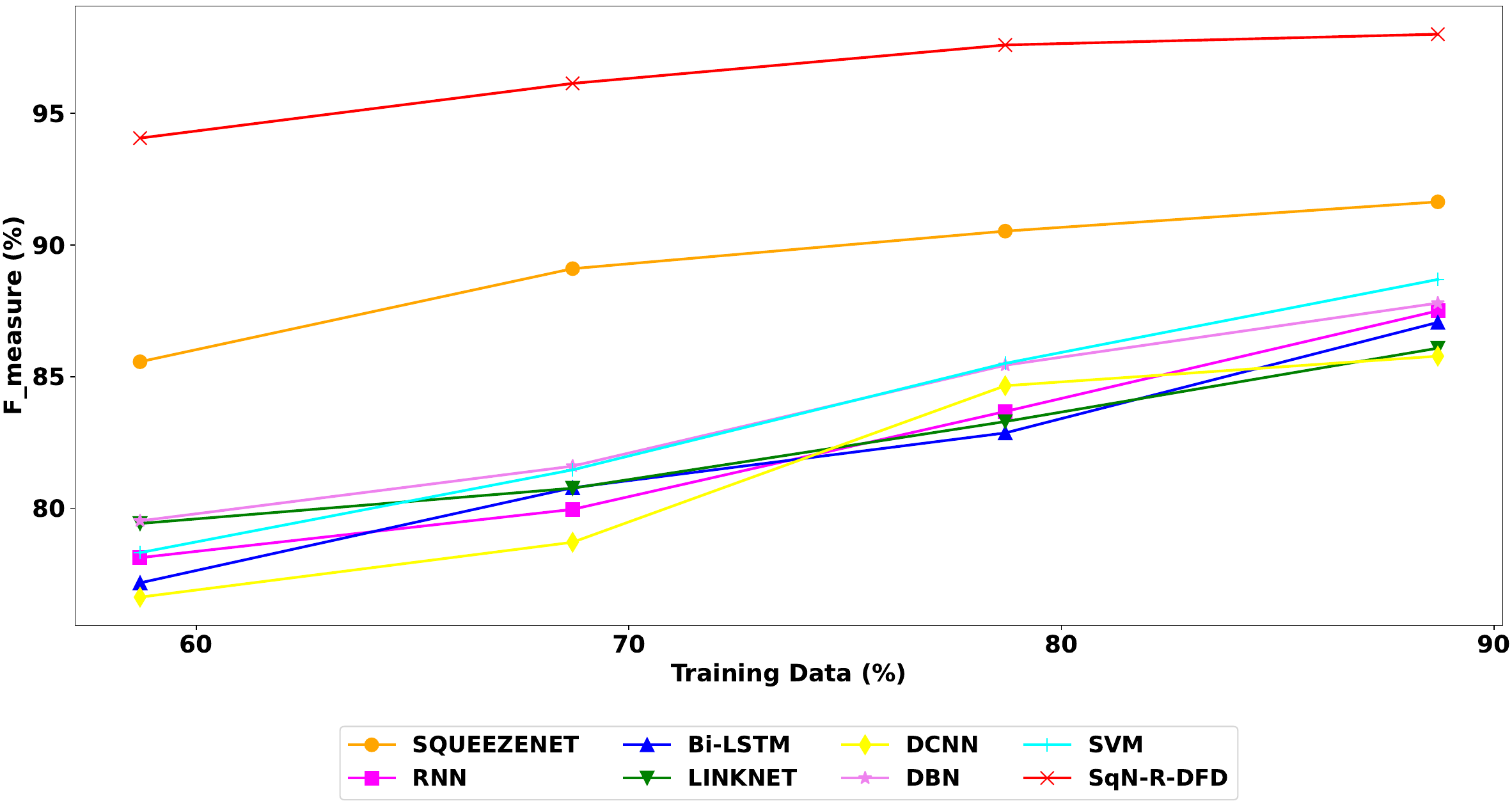}
        \caption{F-measure for Wav2lip Case using Dataset1 \label{fig14fmeasure_e}}
    \end{subfigure}%
     \hfill
    \begin{subfigure}{0.45\textwidth}
        \includegraphics[width=\linewidth]{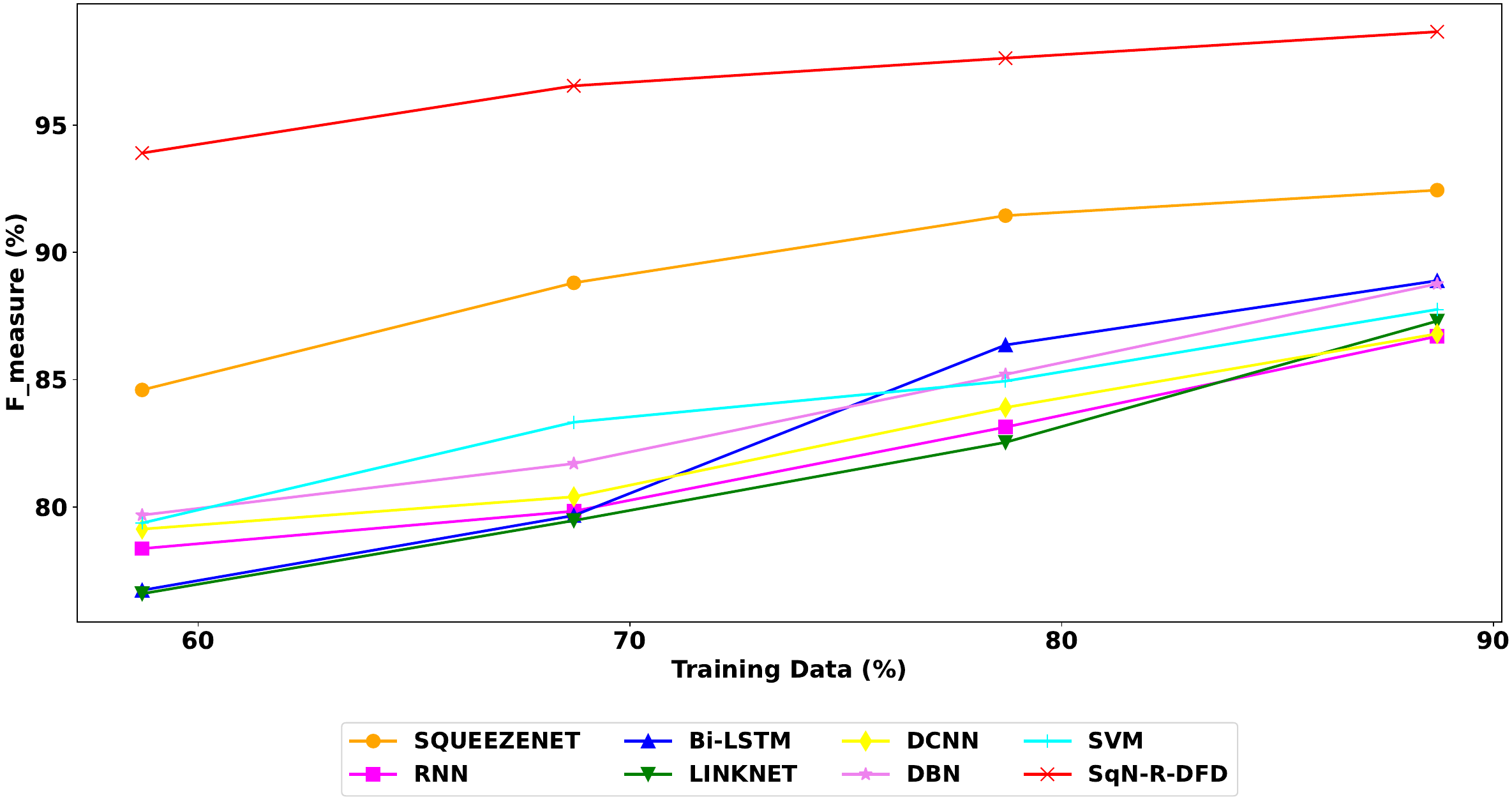}
        \caption{F-measure for Faceswapped Case using Dataset1 \label{fig14fmeasure_f}}
    \end{subfigure}
             
    \begin{subfigure}{0.45\textwidth}
        \includegraphics[width=\linewidth]{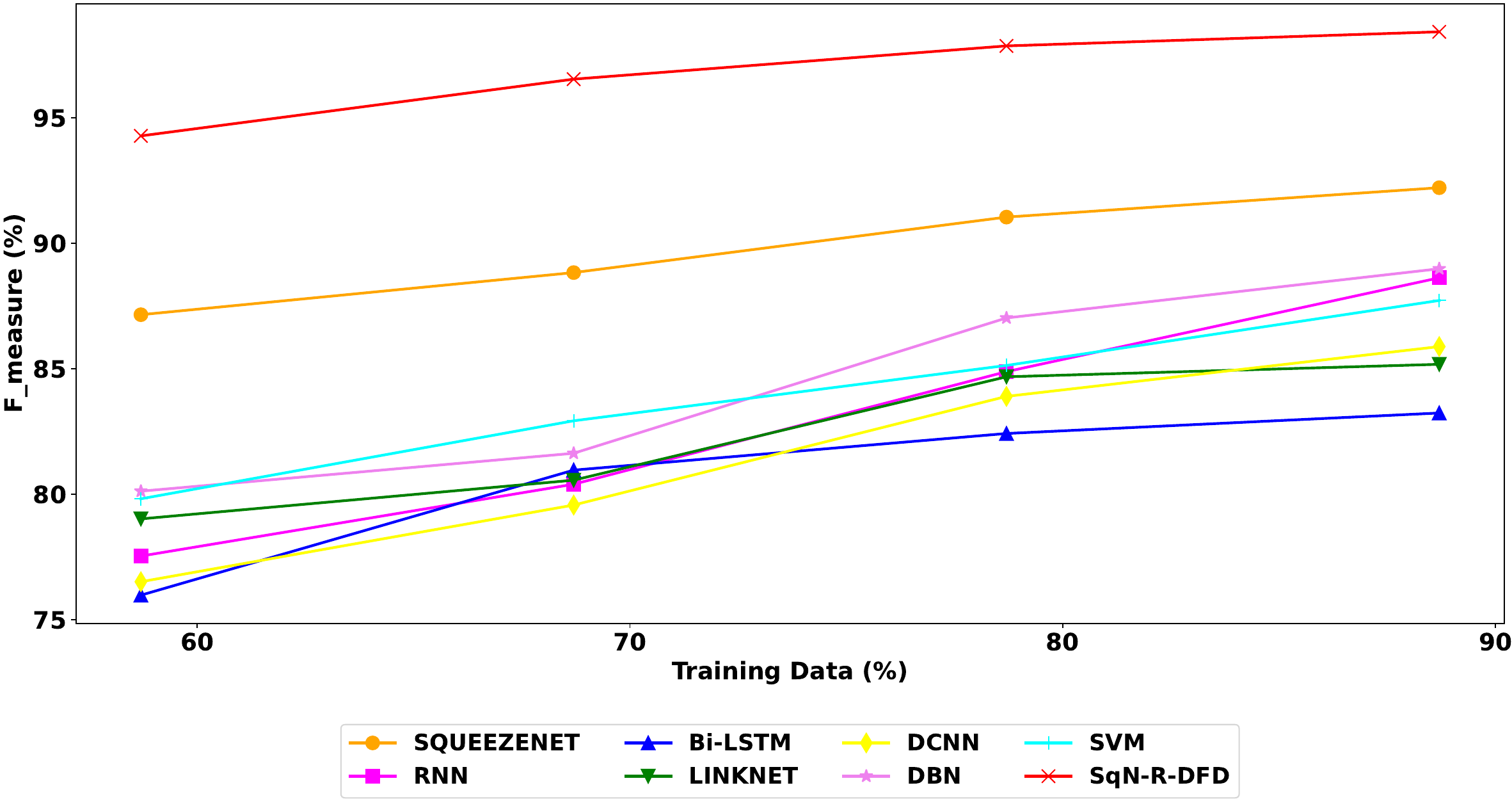}
        \caption{F-measure for Resized Case using Dataset1 \label{fig14fmeasure_g}}
    \end{subfigure}
      \caption{Comparison on SqN-R-DFD Versus Conventional Methods for Dataset1 (a) F-Measure for Original Case (b)  F-Measure for Noisy Case (c)  F-Measure for Blurred Case (d) F-Measure for Compressed Case (e) F-Measure for Wav2lip Case (f) F-Measure for Faceswapped Case (g) F-Measure for Resized Case}
      \label{fig14fmeasure}
\end{figure}

\textbf{ F-measure:}
 Observing the original case of Figure \cref{fig14fmeasure_a}, the SqN-R-DFD scheme obtained the highest F-measure rate of $90.145$ in the training data $60\%$, outperforming conventional methods. As the training data increased to $70\%$ and $80\%$, the SqN-R-DFD scheme continues its lead with an F-measure value of $94.256$ and $95.754$, again surpassing the traditional approaches. Finally, at $90\%$ training data, the highest F-measure value is achieved using the SqN-R-DFD approach is $97.453$, whereas the conventional methods attained the least F-measure values with SqueezeNet at $90.543$, RNN at $83.621$, Bi-LSTM at $82.156$, LinkNet at $84.532$, DCNN at $85.954$, DBN \cite{suganthi2022deep} at $88.521$, and SVM \cite{tu2024face} at $87.413$, respectively. From the examination, the SqN-R-DFD approach reliably outperforms the existing methods, accomplishing the highest F-measure values across all training data in all cases. As the training data improves, the performance of all algorithms enhances. However, the SqN-R-DFD scheme maintains a significant lead. This emphasizes the ability and generalization of the SqN-R-DFD approach in deepfake detection.

\subsubsection{\textbf{ROC Curve Analysis}} \label{roc1}
\begin{figure}[!htbp]
    \centering
    \begin{subfigure}[t]{0.45\textwidth}
        \includegraphics[width=\linewidth]{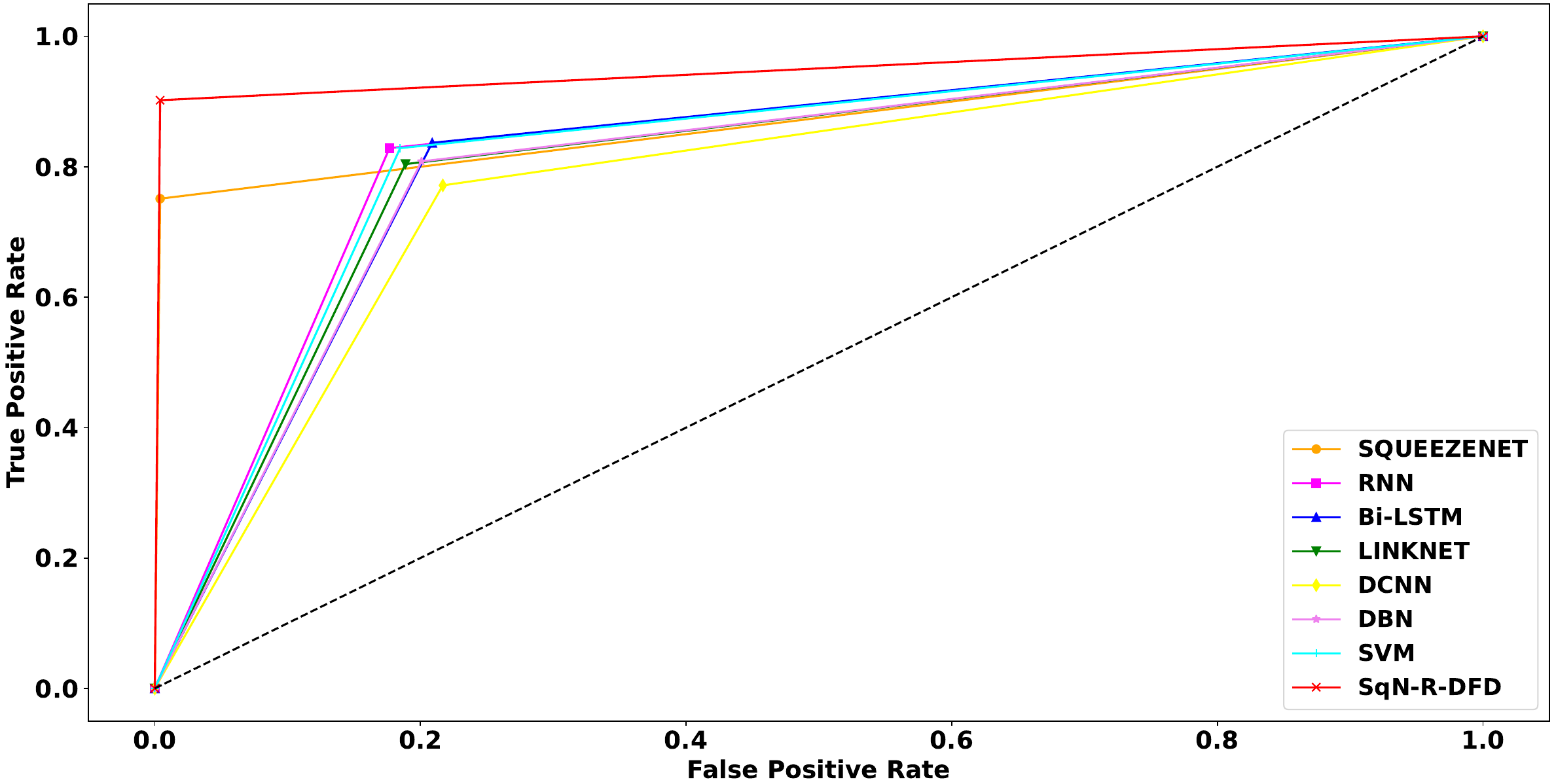}
        \caption{ROC Curve for Original case using Dataset1\label{rocoriginal}}
    \end{subfigure}
    \hfill
    \begin{subfigure}[t]{0.45\textwidth}
        \includegraphics[width=\linewidth]{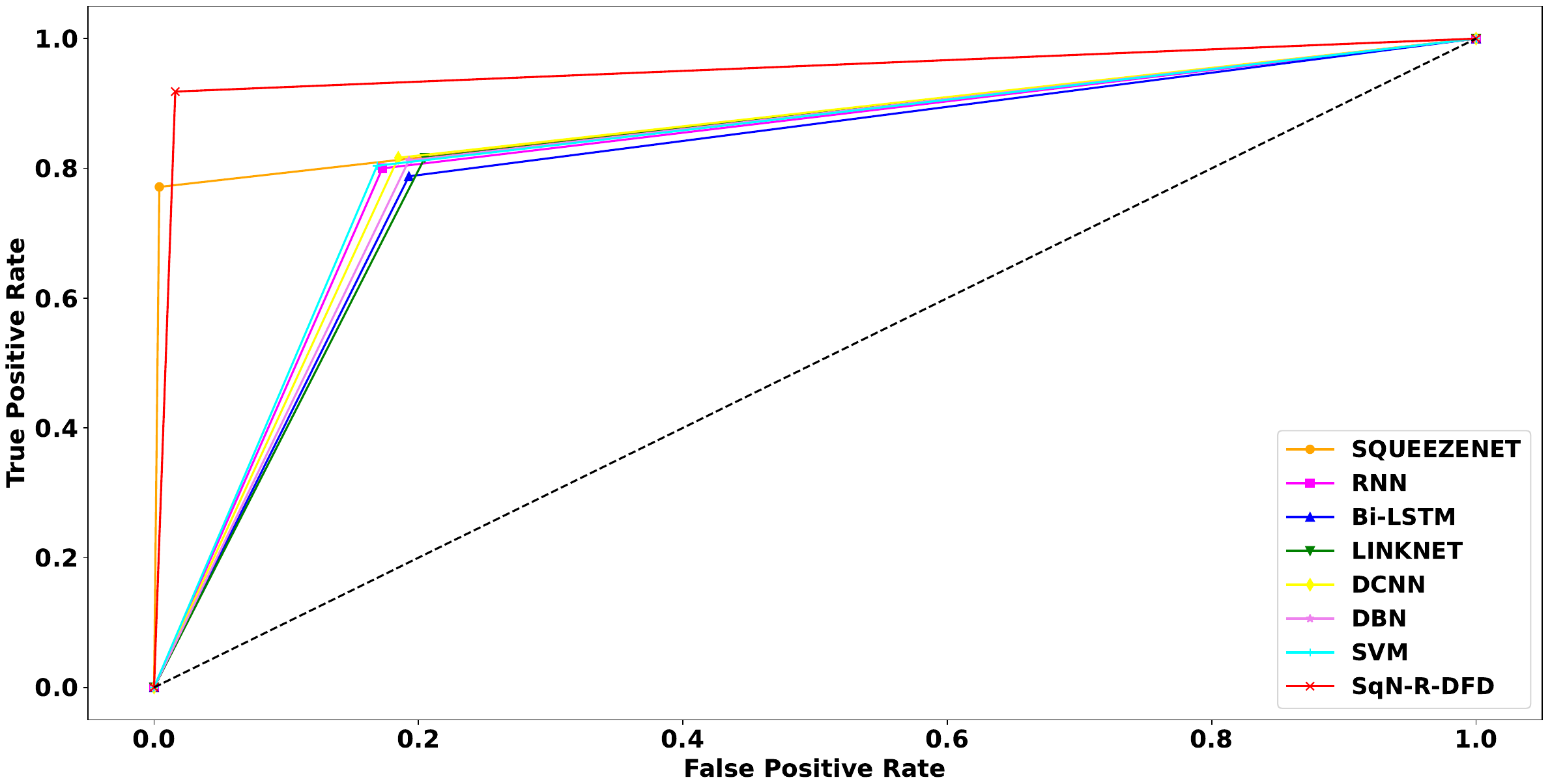}
        \caption{ROC Curve for Noise case using Dataset1 \label{rocnoisy}}
    \end{subfigure}
    \begin{subfigure}{0.45\textwidth}
        \includegraphics[width=\linewidth]{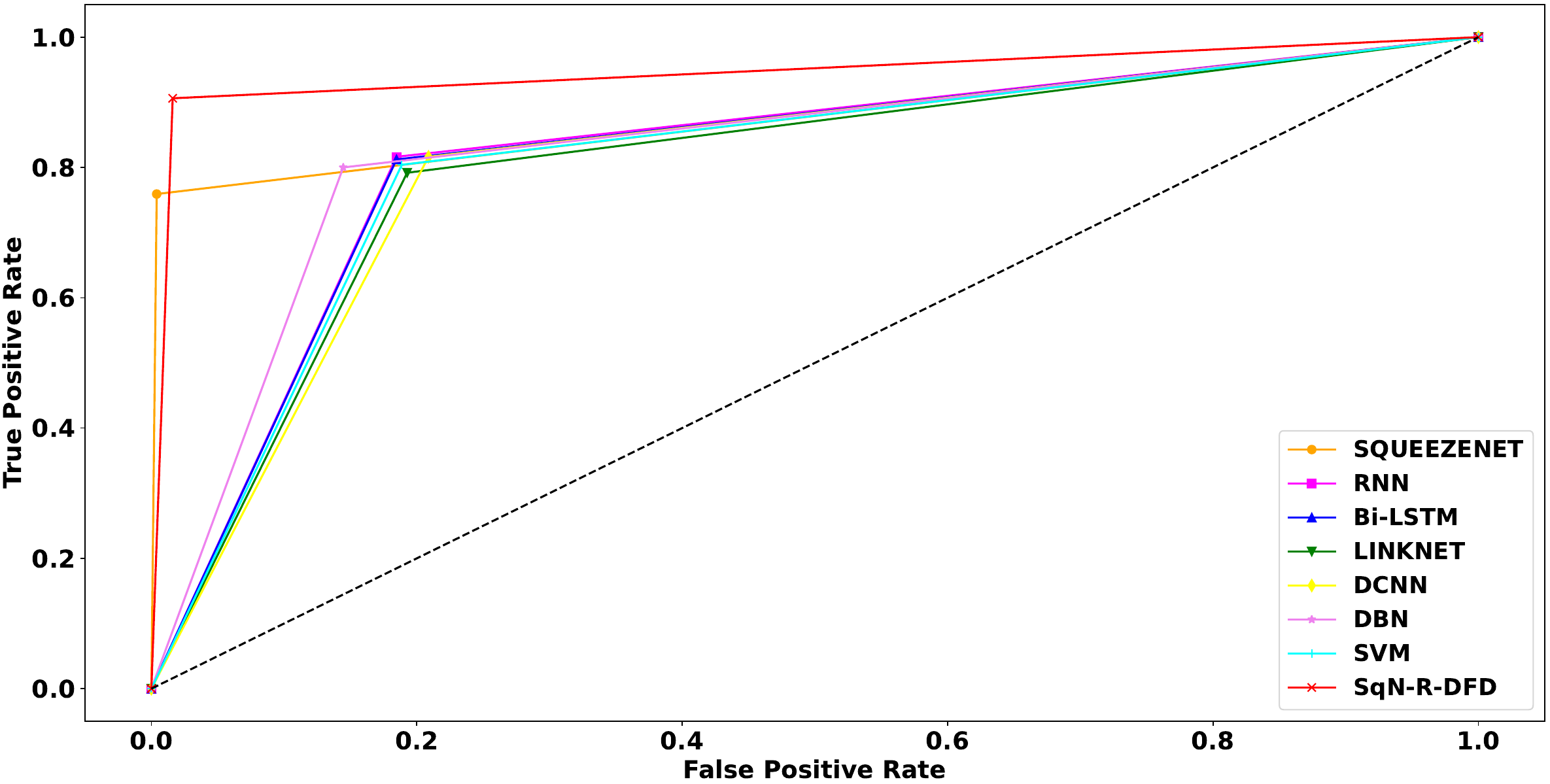}
        \caption{ROC Curve for Blurred case using Dataset1 \label{rocblurred}}
    \end{subfigure}
    \hfill
    \begin{subfigure}{0.45\textwidth}
        \includegraphics[width=\linewidth]{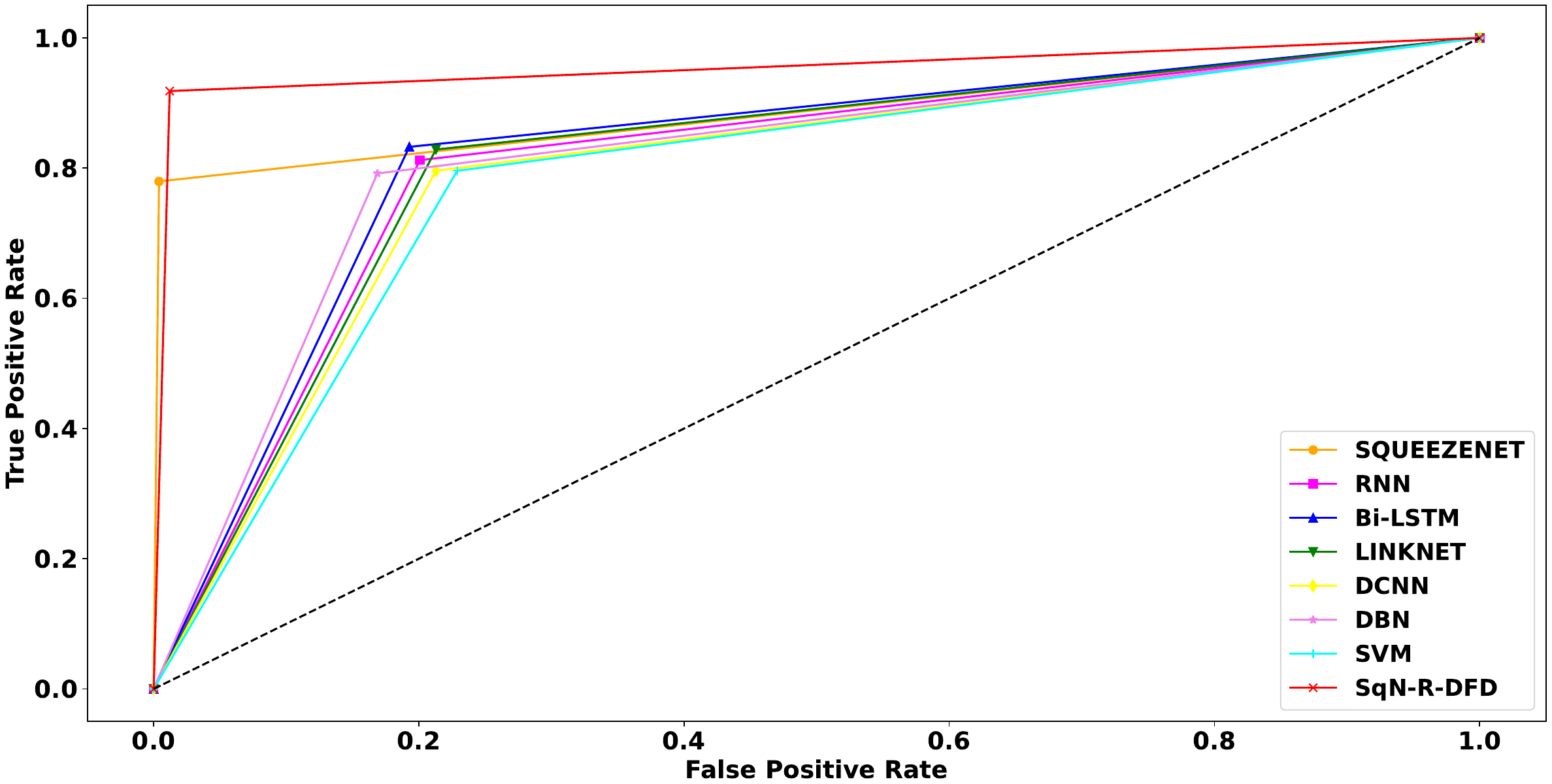}
        \caption{ROC Curve for Compressed case using Dataset1 \label{roccompressed}}
    \end{subfigure}    
    \begin{subfigure}{0.45\textwidth}
        \includegraphics[width=\linewidth]{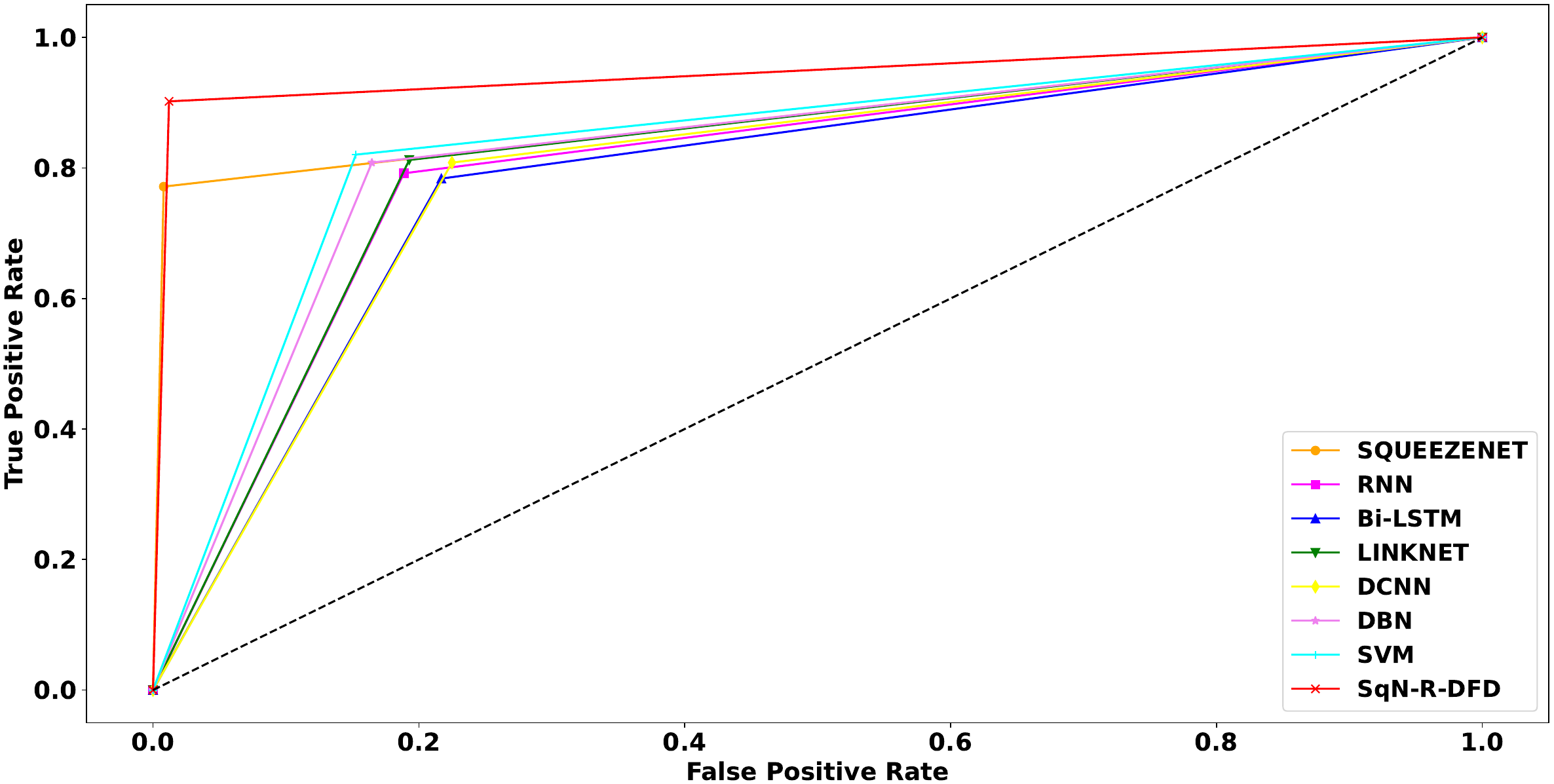}
        \caption{ROC Curve for Wave2lip case using Dataset1 \label{rocwav2lip}}
    \end{subfigure}
    \hfill
    \begin{subfigure}{0.45\textwidth}
        \includegraphics[width=\linewidth]{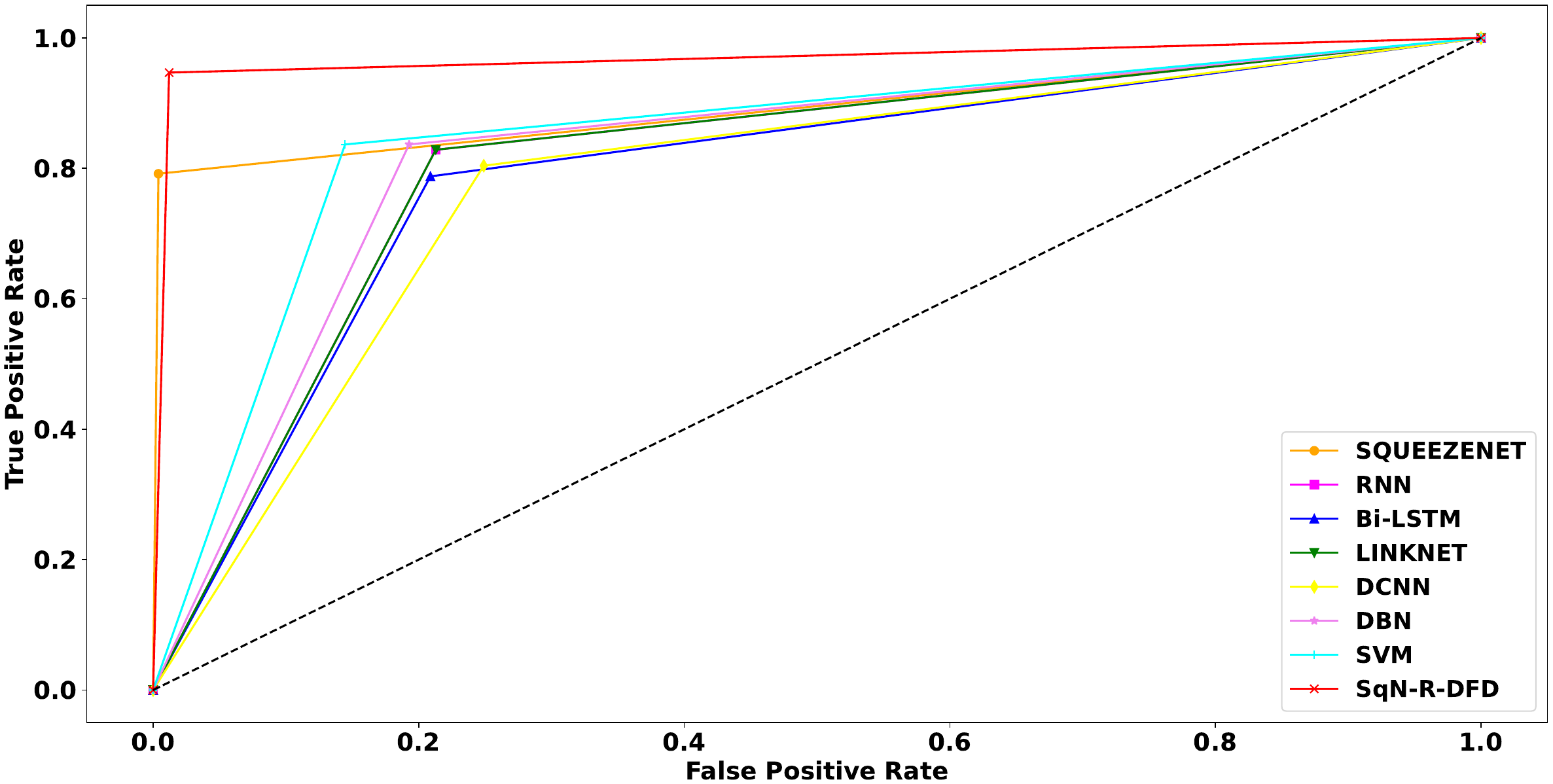}
        \caption{ROC Curve for Faceswapped case using Dataset1\label{rocfaceswapped}}
    \end{subfigure}
    \begin{subfigure}{0.45\textwidth}
        \includegraphics[width=\linewidth]{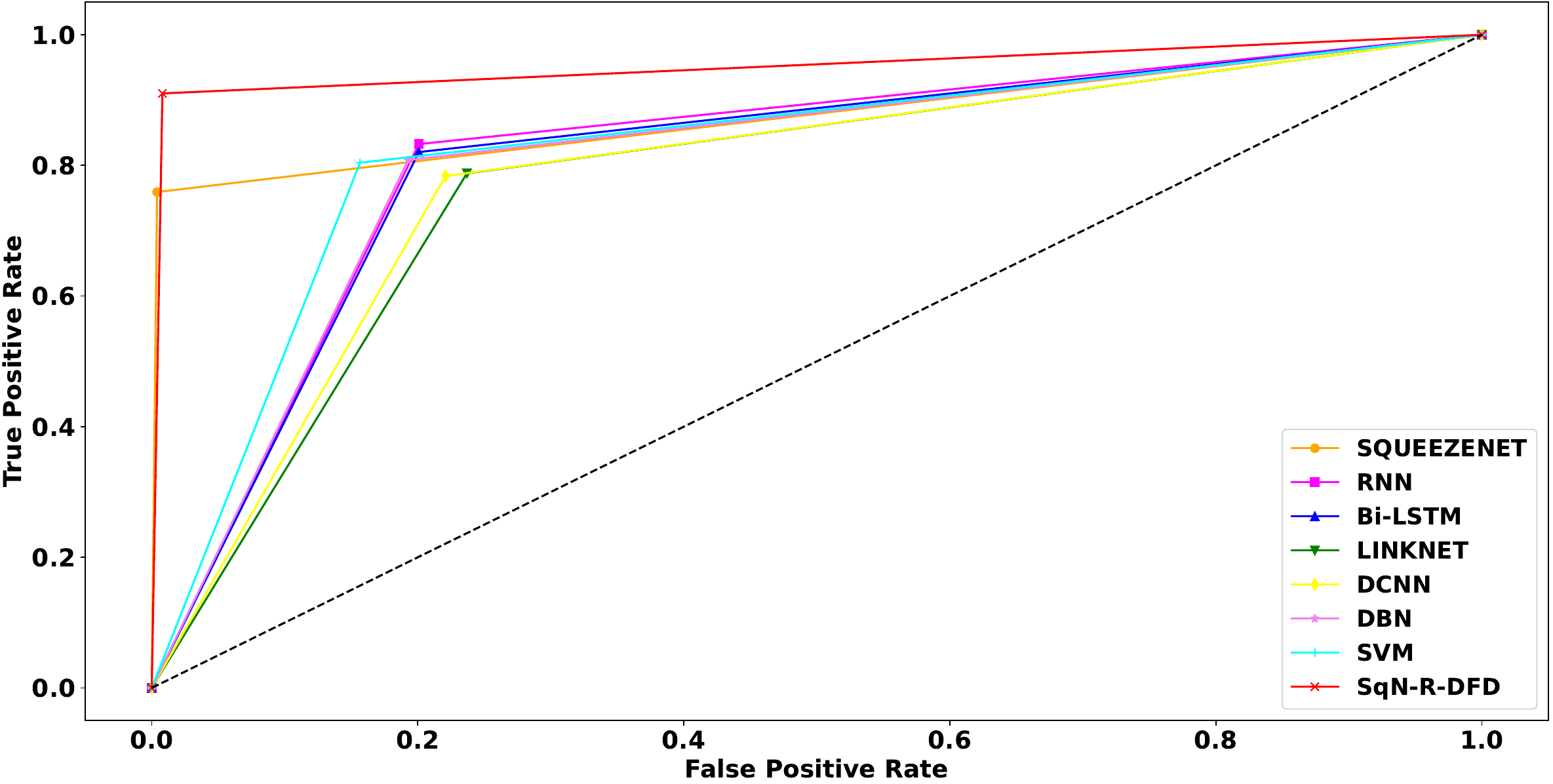}
        \caption{ROC Curve for Resized Case using Dataset1\label{rocresized}}
    \end{subfigure}
\caption{Comparison on SqN-R-DFD Versus Conventional Methods for Original Case on Dataset1 a) ROC Curve for Original case  b) ROC Curve for Noise case c) ROC curve for Blurred case d) ROC Curve for Compressed case e) ROC Curve for Wav2lip case f) ROC curve for Face-swapped case g) ROC Curve for Resized case} \label{Fig7}
\end{figure}

A ROC curve is a graph that shows how well a model performs across different thresholds. The ROC curve is the plot of the TPR against the FPR at each threshold setting. Figure \ref{rocoriginal} depicts the ROC curve analysis on the SqN-R-DFD methodology compared with conventional methods such as SqueezeNet, RNN, Bi-LSTM, LinkNet, DCNN, DBN \cite{suganthi2022deep}, and SVM \cite{tu2024face} for deepfake detection with multimodalities. An AUC score of $0.7$ to $0.8$ is considered adequate, $0.8$ to $0.9$ is reflected as exceptional, and greater than $0.9$ is considered outstanding. More specifically, the SqN-R-DFD scheme achieved the highest AUC rate of $0.927$, suggesting outstanding performance. In comparison, the RNN, SVM \cite{tu2024face}, Bi-LSTM, and LinkNet all attained AUC rates exceeding $0.8$, indicating excellent performance. The SqueezeNet and DCNN demonstrated a lower AUC rate of $0.781$ and $0.793$, signifying acceptable performance. 

\subsubsection{\textbf{Training-Testing Loss Analysis}}\label{training1}
\begin{figure}[!htbp]
    \centering
    \begin{subfigure}{0.45\textwidth}
        \includegraphics[width=\linewidth]{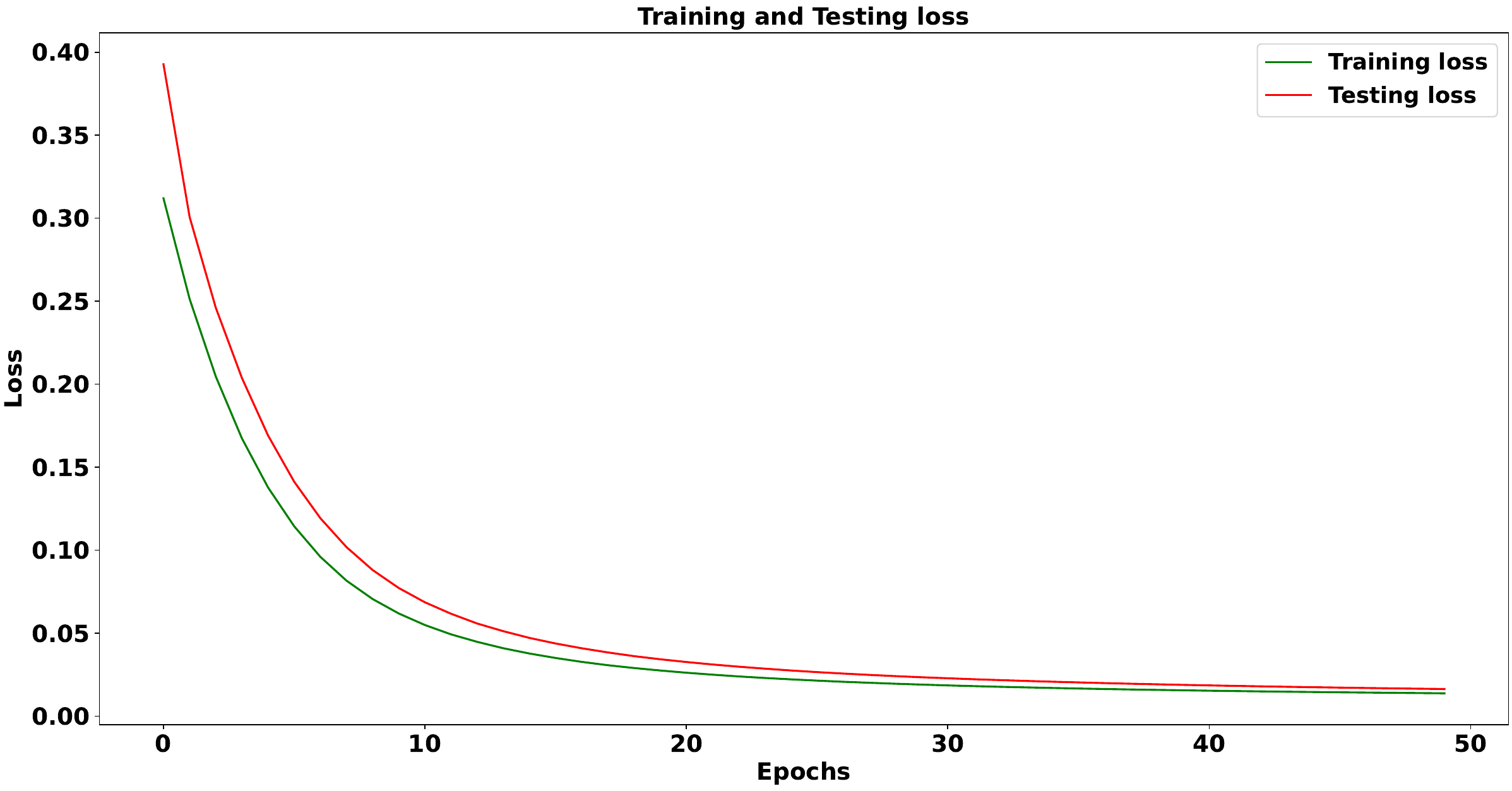}        \caption{Training-Testing Loss curve for Original case using Dataset1 \label{training_original}}
    \end{subfigure}
    \hfill
    \begin{subfigure}{0.45\textwidth}
        \includegraphics[width=\linewidth]{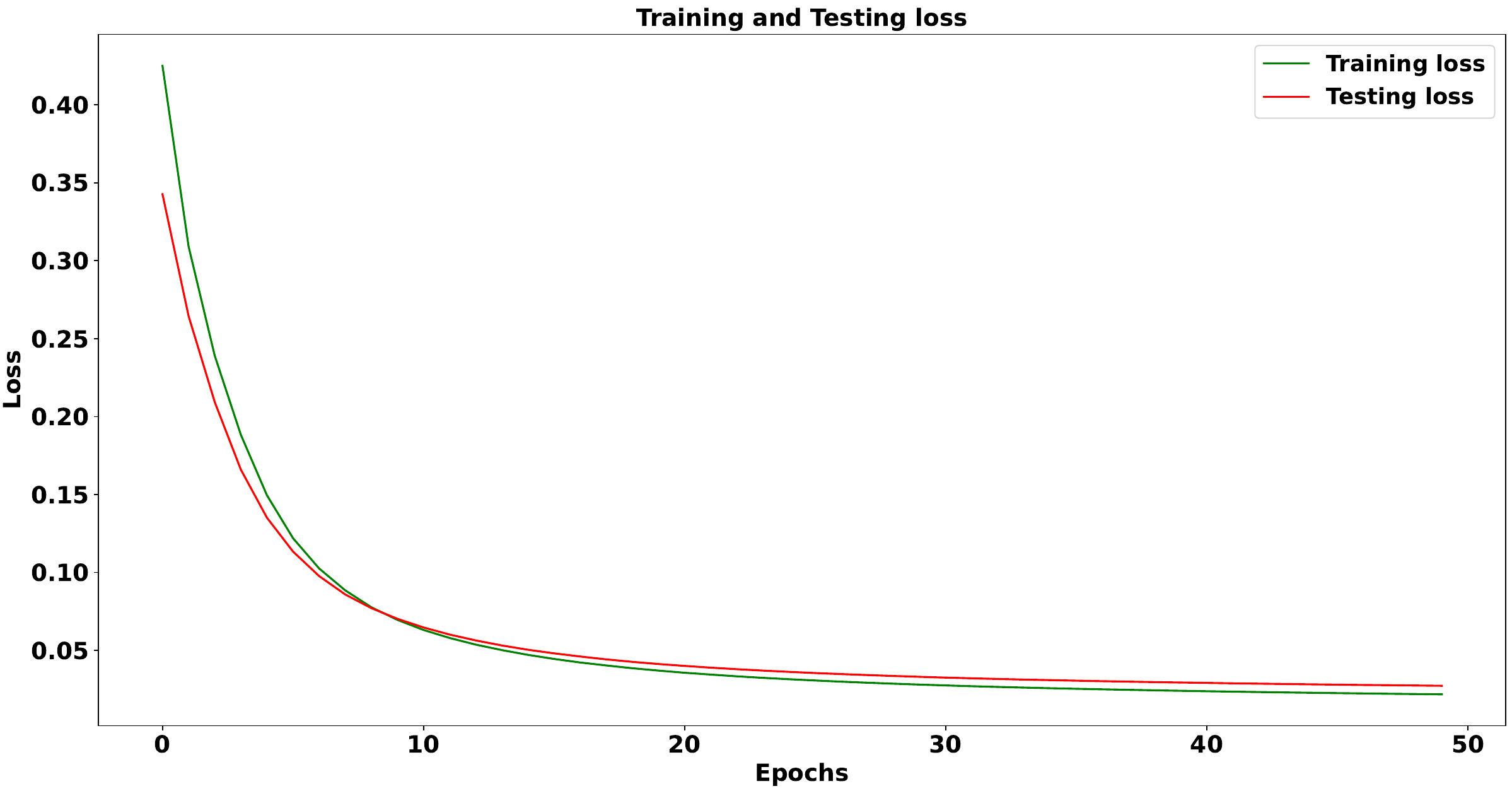}
        \caption{Training-Testing Loss curve for Noise case using Dataset1 \label{trainingnoise}}
    \end{subfigure}    
    \begin{subfigure}{0.45\textwidth}
        \includegraphics[width=\linewidth]{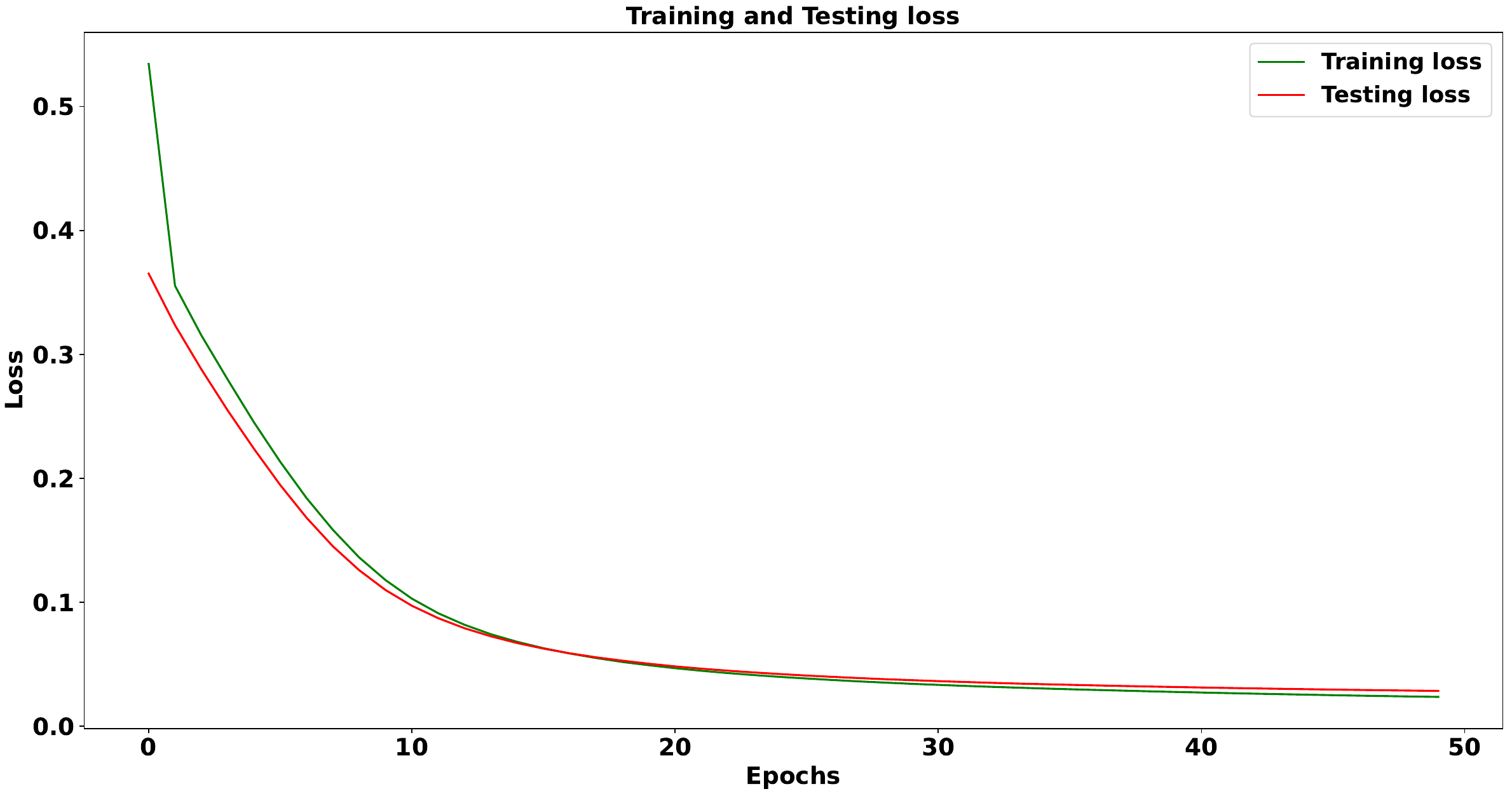}
        \caption{Training-Testing Loss curve for Blurred case using Dataset1\label{trainingblurred}}
    \end{subfigure}
    \hfill
    \begin{subfigure}{0.45\textwidth}
        \includegraphics[width=\linewidth]{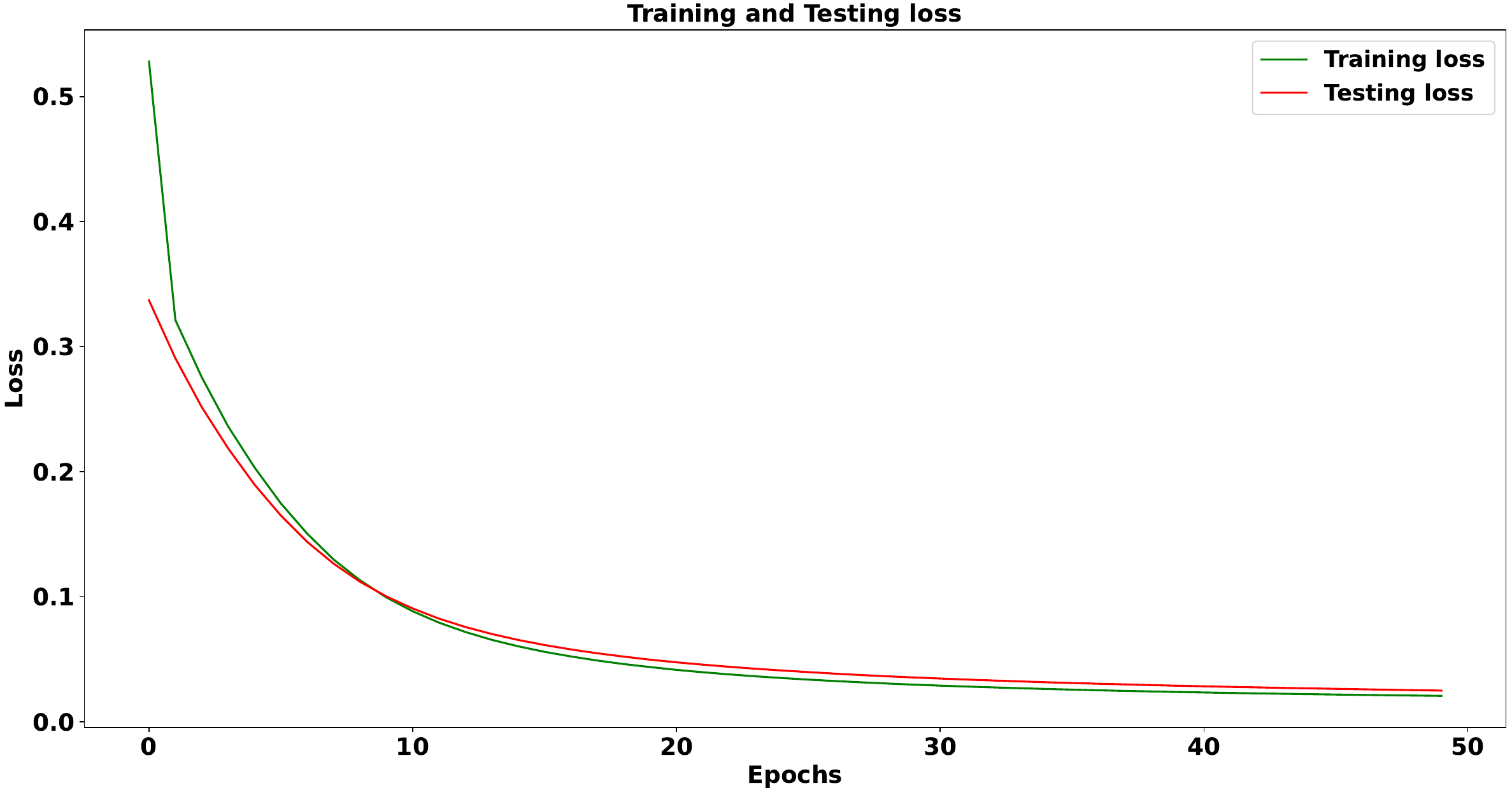}
        \caption{Training-Testing Loss curve for Compressed case using Dataset1\label{trainingcompressed}}
    \end{subfigure} 
    \begin{subfigure}{0.45\textwidth}
        \includegraphics[width=\linewidth]{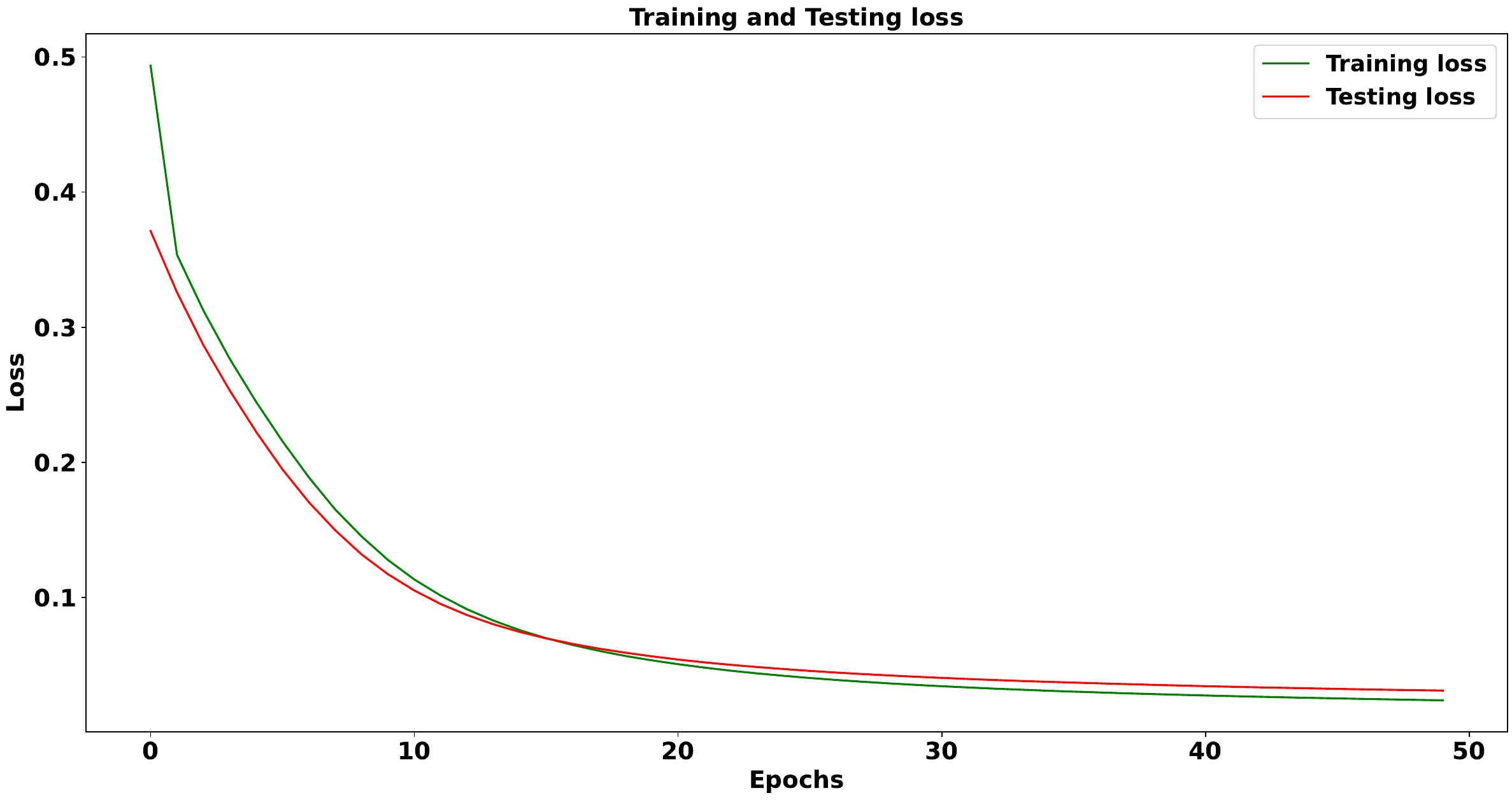}       \caption{Training-Testing Loss curve for Wav2lip case using Dataset1 \label{trainingwav2lip}}
    \end{subfigure}
    \hfill
    \begin{subfigure}{0.45\textwidth}
        \includegraphics[width=\linewidth]{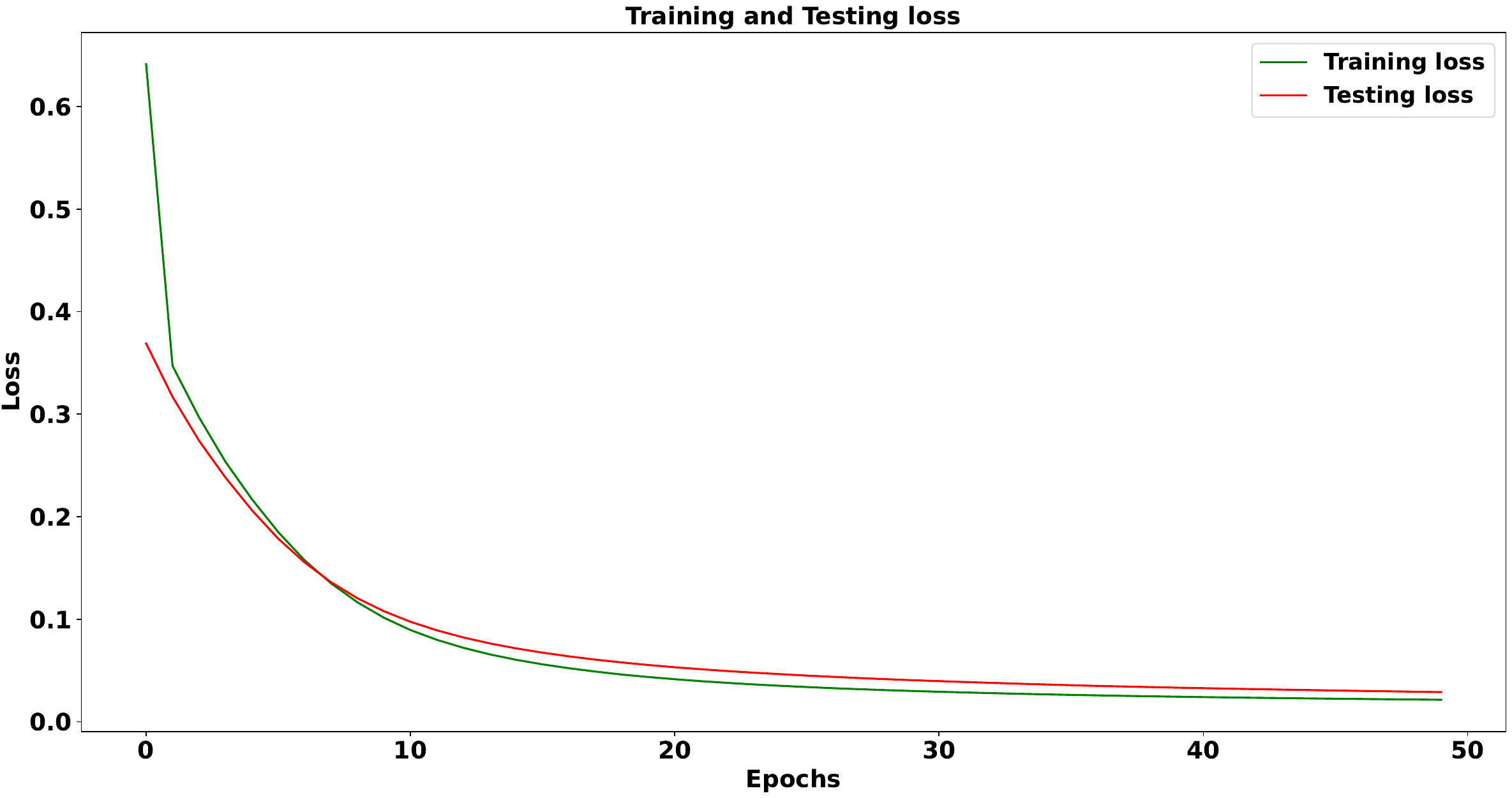}
        \caption{Training-Testing Loss Curve for Faceswapped case using Dataset1 \label{trainingfaceswap}}
    \end{subfigure}
    \begin{subfigure}{0.45\textwidth}
        \includegraphics[width=\linewidth]{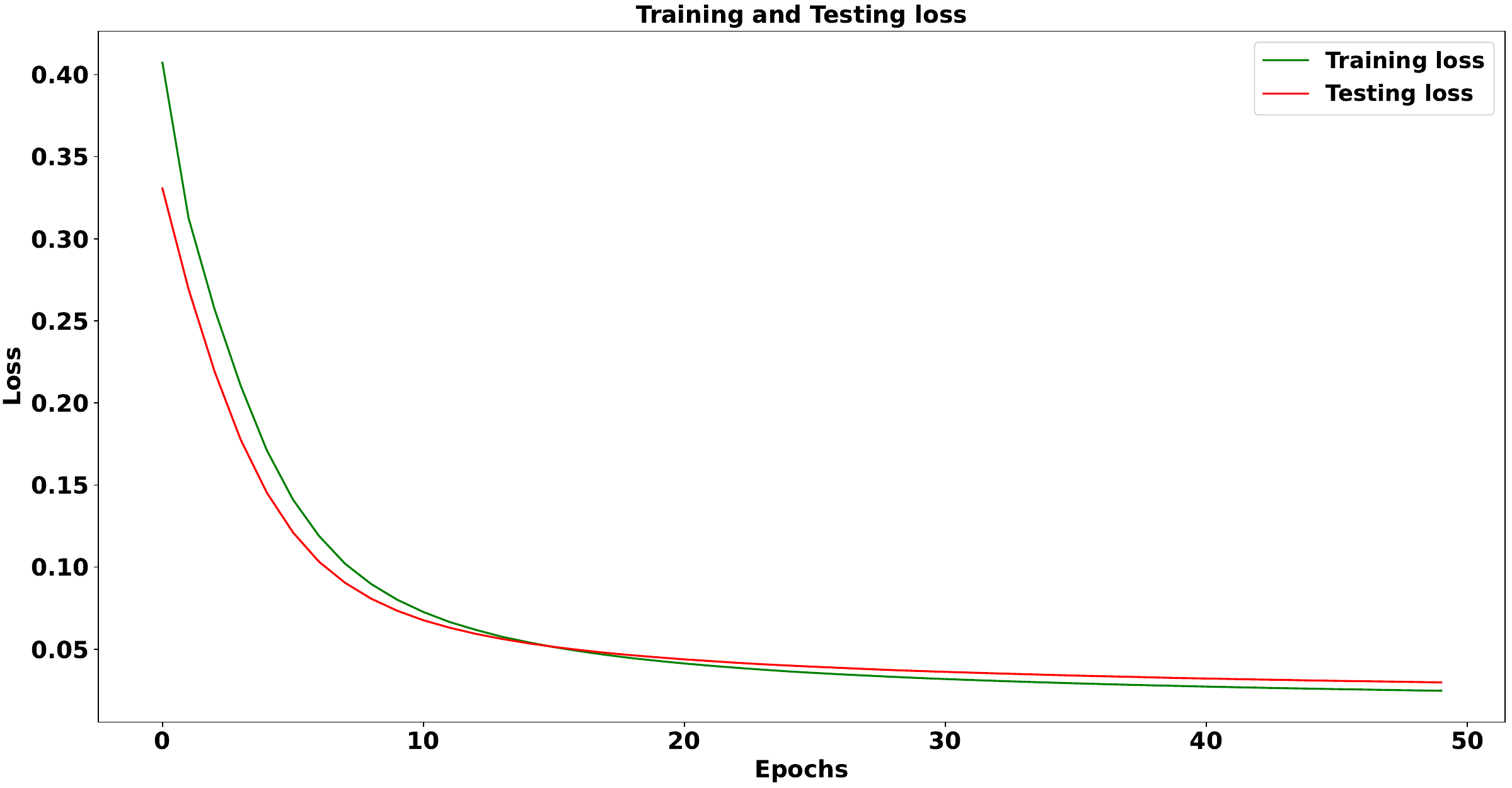}
        \caption{Training-Testing Loss curve for Resized case using Dataset1 \label{trainingresize}}
    \end{subfigure}
\caption{Comparison on SqN-R-DFD Versus Conventional Methods for Training-Testing Loss Curve on Dataset1 a) Training-Testing Loss Curve for Original case  b) Training-Testing Loss Curve for Noise case c)Training-Testing Loss Curve for Blurred case d)Training-Testing Loss Curve for Compressed case e) Training-Testing Loss Curve for Wav2lip case f)Training-Testing Loss Curve for Face-swapped case g) Training-Testing Loss Curve for Resized case} \label{Fig7new}
\end{figure}

Figure \ref{Fig7new} shows the training and testing loss curves for the SqN-R-DFD approach in the context of Deepfake detection. This study exposes the learning performance of the SqN-R-DFD model over $50$ epochs. Primarily, training and testing loss displays a severe decline, suggesting quick learning and adaptation to data. This decrease steadies nearby epoch $20$, demonstrating that the model has converged and is no longer making substantial enhancements. The nominal gap among the training and testing loss curves throughout the procedure indicates the model is generalizing well to unseen data and non-overfitting. This indicates that the low values at the end of training recommend a well-performing model. This enhancement is largely due to the incorporation of ISLBT-based features, in which the standard form of LBP pattern is applied to the shape-free patches to extract features from the preprocessed face image, which are modified.\\	
 A training and testing loss of the SqN-R-DFD model is depicted in Figure \ref{trainingnoise}. Firstly, training and testing losses are moderately higher, starting at $0.42$ and $0.35$. As the epochs advanced, a reduction in both training and testing was noticed. At the $10^{th}$ epoch, both losses have decreased significantly, converging below $0.1$. The minor gap between the two curves specifies that the SqN-R-DFD approach is not overfitting the training data and preserves an effective generalization capability.  
 \enlargethispage{-\baselineskip}

\subsubsection{\textbf{Comparative Analysis on Feature Selection}} \label{featureselectionwldr}
Table \ref{table2} presents a comparative analysis of the accuracy and F-measure of different feature selection methods on the WLDR Dataset \cite{agarwal2019protecting}. This comparison includes SMA-SOA, PCA, RFE-RF (Recursive Feature Elimination with Random Forest), SOA, KOA (Kookubura Optimization Algorithm), and SMA feature selection methods for Deepfake detection with multimodalities, illustrated in Table \ref{table2}. The Accuracy score achieved using the SMA-SOA approach is $0.945$, which is extremely higher compared to other models. In contrast, the KOA model obtained the least accuracy rate of $0.917$, RFE-RF and SOA scored the accuracies of $0.921$ and $0.924$, respectively. Further, the PCA and SMA accomplished a slightly improved accuracy of $0.931$. However, the outcomes exhibit that the SMA-SOA surpasses all others in terms of accuracy metrics.

\begin{table}[!htbp]
 \centering
    \begin{tabular}{|l|l|l|l|l|}
     \hline
 \textbf{Methods} & \begin{tabular}[c]{@{}c@{}} \textbf{accuracy} \end{tabular} & \begin{tabular}[c]{@{}c@{}}\textbf{F-measure}\\ \end{tabular} \\ 
 \hline    
PCA & 0.931 & 0.945 \\ \hline
RFE-RF & 0.921 & 0.928 \\ \hline 
SOA  & 0.924 & 0.942 \\ \hline
KOA & 0.917 & 0.922 \\ \hline
SMA & 0.931 & 0.928 \\ \hline
SMA-SOA & 0.945 & 0.968 \\ \hline
\end{tabular}%
\caption{Analysis on Feature Selection Methods for Dataset1}
    \label{table2}
\end{table}

\subsubsection{\textbf{Ablation Study}} \label{ablation1}
Table \ref{table3} describes the ablation assessment on the SqN-R-DFD scheme in comparison to distinct variants, such as SqN-R-DFD with existing SLBT and SqN-R-DFD with existing feature selection for deepfake detection. The least FPR value attained using the SqN-R-DFD model is $0.012$, signifying superiority in deepfake detection. Conversely, the SqN-R-DFD with existing feature selection and the SqN-R-DFD with existing PNCC attained the FPR values of $0.114$, and 
$0.112$, indicating higher error ratings. The SqN-R-DFD with existing SLBT recorded a slightly lower FPR score of $0.096$, but did not outperform the SqN-R-DFD approach.
\begin{table}[H]
 \centering
    \resizebox{\textwidth}{!}{%
    \begin{tabular}{|l|l|l|l|l|}
     \hline
 \textbf{Metrics} & \begin{tabular}[c]{@{}c@{}}\textbf{SqN-R-DFD with}\\ \textbf{Existing PNCC} \end{tabular} & \begin{tabular}[c]{@{}c@{}}\textbf{SqN-R-DFD with}\\ \textbf{Existing SLBT}\end{tabular}  & \begin{tabular}[c]{@{}c@{}}\textbf{SqN-R-DFD with}\\ \textbf{Existing Feature}\\ \textbf{Selection} \end{tabular}  & \begin{tabular}[c]{@{}c@{}}\textbf{SqN-R-DFD} \end{tabular}       \\ 
 \hline    
Accuracy & 0.924 & 0.934 & 0.931 & 0.945 \\ \hline
FPR & 0.040 & 0.037 & 0.024 & 0.012 \\ \hline 
F-Measure & 0.922 & 0.932 & 0.928 & 0.968 \\ \hline
Sensitivity & 0.888 & 0.904 & 0.886 & 0.944 \\ \hline
FNR & 0.112 & 0.096 & 0.114 & 0.056  \\ \hline
Specificity & 0.960 & 0.963 & 0.976 & 0.988  \\ \hline
NPV & 0.895 & 0.908 & 0.896 & 0.932  \\ \hline
Precision & 0.959 & 0.962 & 0.973 & 0.993  \\ \hline
MCC & 0.871 & 0.894 & 0.866 & 0.894 \\ \hline
    \end{tabular}%
    }
    \caption{Ablation Analysis on SqN-R-DFDApproach, SqN-R-DFD with Existing PNCC, SqN-R-DFD with Existing SLBT and SqN-R-DFD with Existing Feature Selection using Dataset1}
    \label{table3}
\end{table}
\subsubsection{\textbf{Comparison of SqN-R-DFD Model with Cross-Dataset Performance and Unbalanced Data}} \label{crossdataset1}
The performance comparison of the SqN-R-DFD approach with cross-dataset analysis, where it is trained using Dataset1 and tested using Dataset2, as well as unbalanced data, is illustrated in Table \ref{table4}. The SqN-R-DFD approach obtained the highest NPV score of $0.932$, while the cross dataset scored $0.921$ and the unbalanced data acquired $0.914$. In this context, the SqN-R-DFD model, which we trained on the balanced dataset, surpasses both cross-dataset analysis and unbalanced data across all metrics.
\begin{table}[H]
 \centering
    \resizebox{\textwidth}{!}{%
    \begin{tabular}{|l|l|l|l|l|}
     \hline
 \textbf{Metrics} & \begin{tabular}[c]{@{}c@{}}\textbf{Cross Dataset Analysis}\\ \textbf{(Trained using Dataset1}\\ \textbf{and tested on Dataset2)} \end{tabular} & \begin{tabular}[c]{@{}c@{}}\textbf{Unbalanced Data} \end{tabular}  & \begin{tabular}[c]{@{}c@{}}\textbf{SqN-R-DFD}\\ \end{tabular} \\ 
 \hline    
Accuracy & 0.937 & 0.943 & 0.945 \\ \hline
Sensitivity & 0.889 & 0.906 & 0.945 \\ \hline 
Specificity & 0.972 & 0.980 & 0.988 \\ \hline
Precision & 0.966 & 0.978 & 0.993 \\ \hline
F-Measure & 0.926 & 0.941 & 0.968  \\ \hline
MCC & 0.871 & 0.889 & 0.894  \\ \hline
NPV & 0.921 & 0.914 & 0.932  \\ \hline
FPR & 0.028 & 0.020 & 0.013  \\ \hline
FNR & 0.111 & 0.094 & 0.056 \\ \hline
    \end{tabular}%
    }
    \caption{Performance Analysis for SqN-R-DFD Model compared to Cross-Dataset Analysis and Unbalanced Data using Dataset1}
    \label{table4}
\end{table}
\subsubsection{\textbf{Analysis on K-Fold Cross Validation}} \label{kfold1}
In K-fold cross-validation, the data is initially divided into K uniform folds. Consequently, k iterations of training and validation are completed, such that within each iteration, an assorted fold of the data is held out for validation while the residual $K-1$ folds are applied for learning. The K-fold cross-validation investigation on the SqN-R-DFD approach is contrasted with conventional methods for deepfake detection, as depicted in Table \ref{table5}. The SqN-R-DFD approach acquired the greatest accuracy values with $0.927$ at $2^{nd}$ fold, $0.949$ at $3^{rd}$ fold, $0.954$ at $4^{th}$ fold, $0.955$ at $5^{th}$ fold and $0.965$ at $6^{th}$ fold, respectively. The smaller variation in the SqN-R-DFD method's accuracy values across folds signifies an enhanced generalization ability, demonstrating that it is less prone to overfitting. In comparison, traditional approaches such as SqueezeNet, RNN, Bi-LSTM, LinkNet, DCNN, DBN \cite{suganthi2022deep}, and SVM \cite{tu2024face} exhibit greater variations in accuracy between the folds.

\begin{table}[H]
 \centering
    \resizebox{\textwidth}{!}{%
    \begin{tabular}{|l|l|l|l|l|l|l|l|l|}
     \hline
 \textbf{K-Folds} & \begin{tabular}[c]{@{}c@{}}\textbf{SqueezeNet}\\  \end{tabular} & \begin{tabular}[c]{@{}c@{}}\textbf{RNN} \end{tabular}  & \begin{tabular}[c]{@{}c@{}}\textbf{Bi-LSTM}\\ \end{tabular} & \begin{tabular}[c]{@{}c@{}}\textbf{LinkNet} \end{tabular}  & \begin{tabular}[c]{@{}c@{}}\textbf{DCNN} \end{tabular}  & \begin{tabular}[c]{@{}c@{}}\textbf{DBN \cite{suganthi2022deep}} \end{tabular}  & \begin{tabular}[c]{@{}c@{}}\textbf{SVM\cite{tu2024face}} \end{tabular}  & \begin{tabular}[c]{@{}c@{}}\textbf{Sqn-R-DFD} \end{tabular}\\ 
 \hline    
2 & 0.869 & 0.779 & 0.783 & 0.797 & 0.769 & 0.793 & 0.786 & 0.927 \\ \hline
3 & 0.870 & 0.803 & 0.792 & 0.798 & 0.775 & 0.810 & 0.807 & 0.949 \\ \hline
4 & 0.879 & 0.808 & 0.831 & 0.807 & 0.789 & 0.830 & 0.834 & 0.954 \\ \hline
5 & 0.888 & 0.822 & 0.850 & 0.811 & 0.794 & 0.859 & 0.862 & 0.955 \\ \hline
6 & 0.909 & 0.873 & 0.851 & 0.846 & 0.835 & 0.868 & 0.864 & 0.965 \\ \hline
\end{tabular}%
    }
 \caption{K-Fold Cross Validation on SqN-R-DFD and Conventional Methods in terms of Accuracy using Dataset1}
\label{table5}
\end{table}

\subsection{\textbf{Comparative Assessment on Dataset2}} \label{result1}
This subsection presents a comparative analysis of the DeepfakeTIMIT dataset \cite{deepfaketimit} for detecting deepfakes against various cases, such as original, noise, blurred, compressed, wav2lip, faceswapped, and resize demonstrated in Subsubsections \ref{comparative2}, \ref{roc2}, and \ref{training2} respectively, in terms of accuracy, F-measure, analysis based on the ROC curve, and training-testing loss curve. In addition, to ensure a thorough evaluation, we combine seven different cases into a single study. This evaluation framework encompasses statistical analysis, an ablation study, cross-dataset analysis, and K-fold cross-validation, as illustrated in Subsections \ref{featureselection2}, \ref{ablation2}, \ref{crossdataset2}, and \ref{kfold2}, respectively.
\subsubsection{\textbf{Comparative Analysis}} \label{comparative2}
\begin{figure}[!htbp]
    \centering    
    \begin{subfigure}[t]{0.45\textwidth}
        \includegraphics[width=\linewidth]{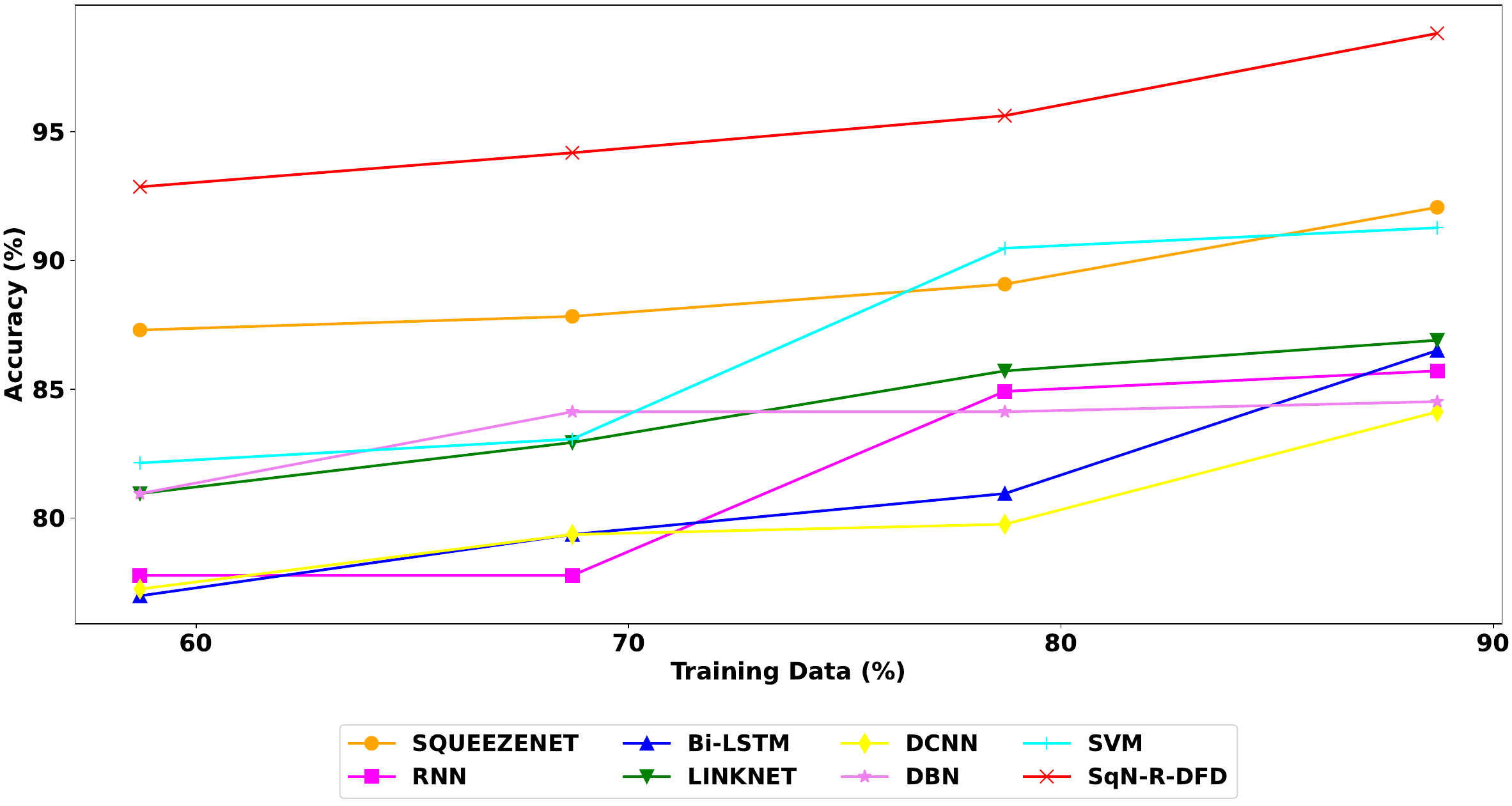}
        \caption{Accuracy for Original Case using Dataset2  \label{fig14a}}   
    \end{subfigure}%
 \hfill
       \begin{subfigure}[t]{0.45\textwidth}
        \includegraphics[width=\linewidth]{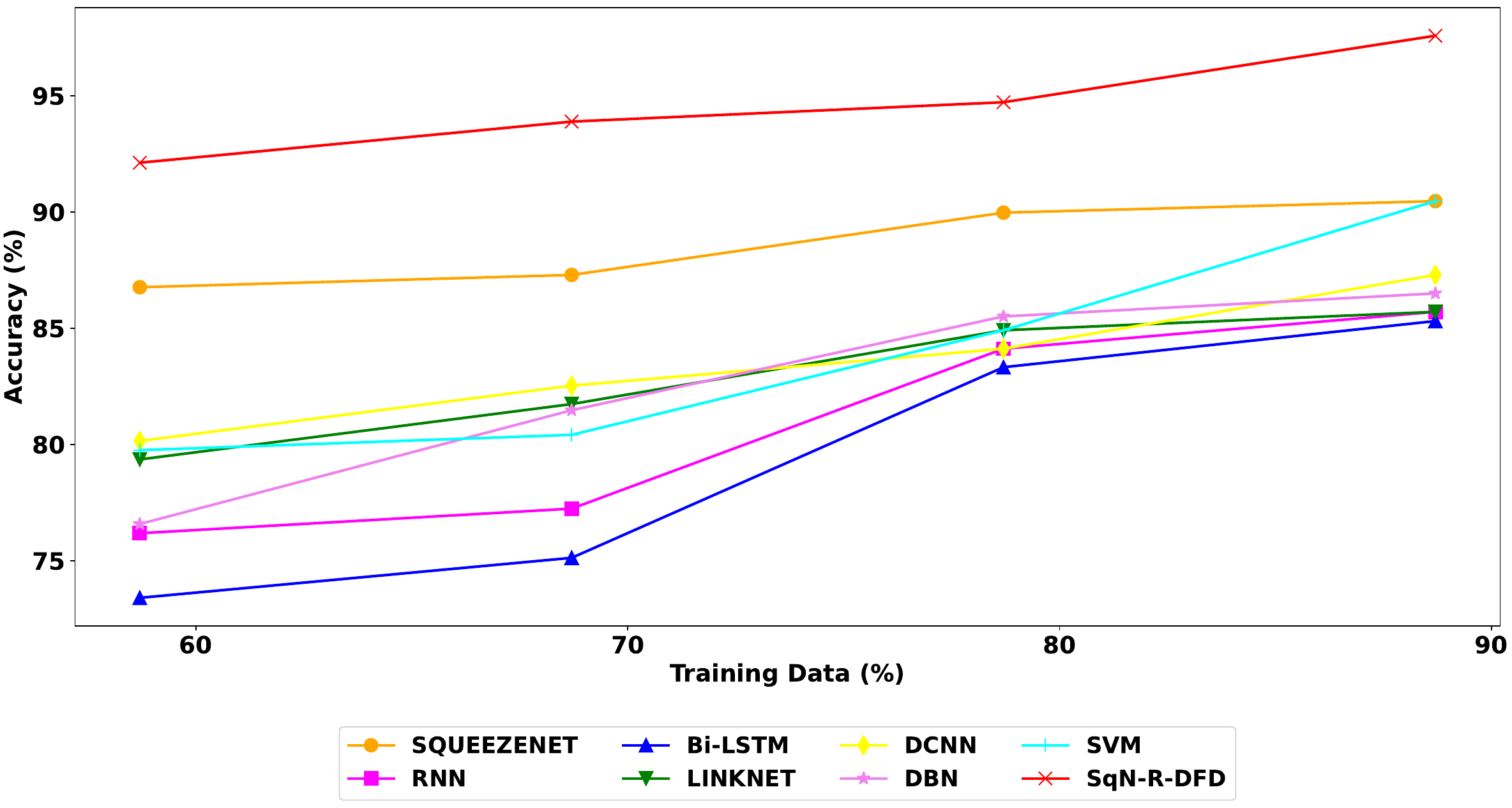}
        \caption{Accuracy for Noise Case using Dataset2 \label{fig14b}}  
    \end{subfigure}
      
    \begin{subfigure}{0.45\textwidth}
        \includegraphics[width=\linewidth]{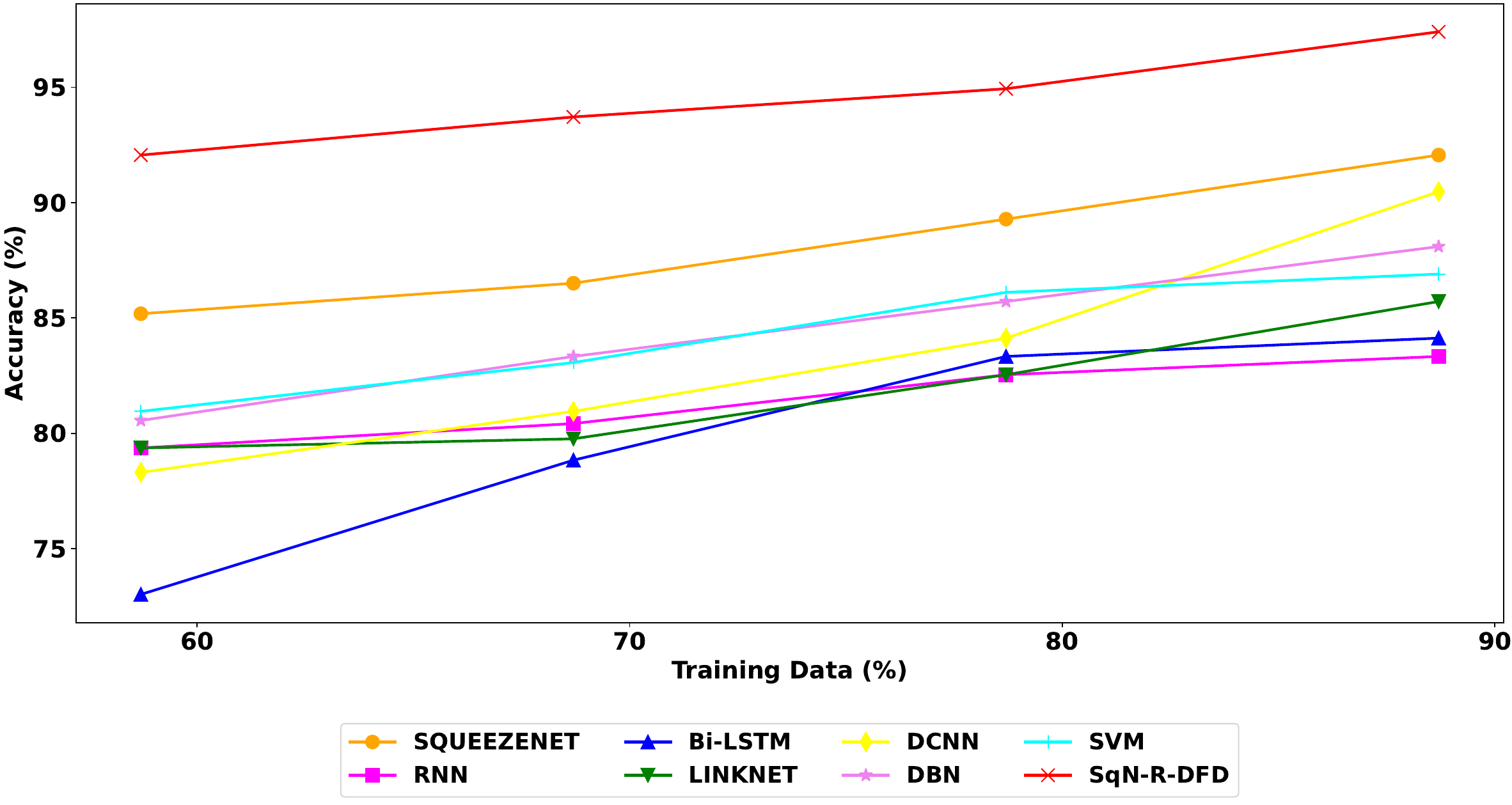}
        \caption{Accuracy for Blurred Case using Dataset2  \label{fig14c}}   
    \end{subfigure}%
    \hfill
    \begin{subfigure}{0.45\textwidth}
        \includegraphics[width=\linewidth]{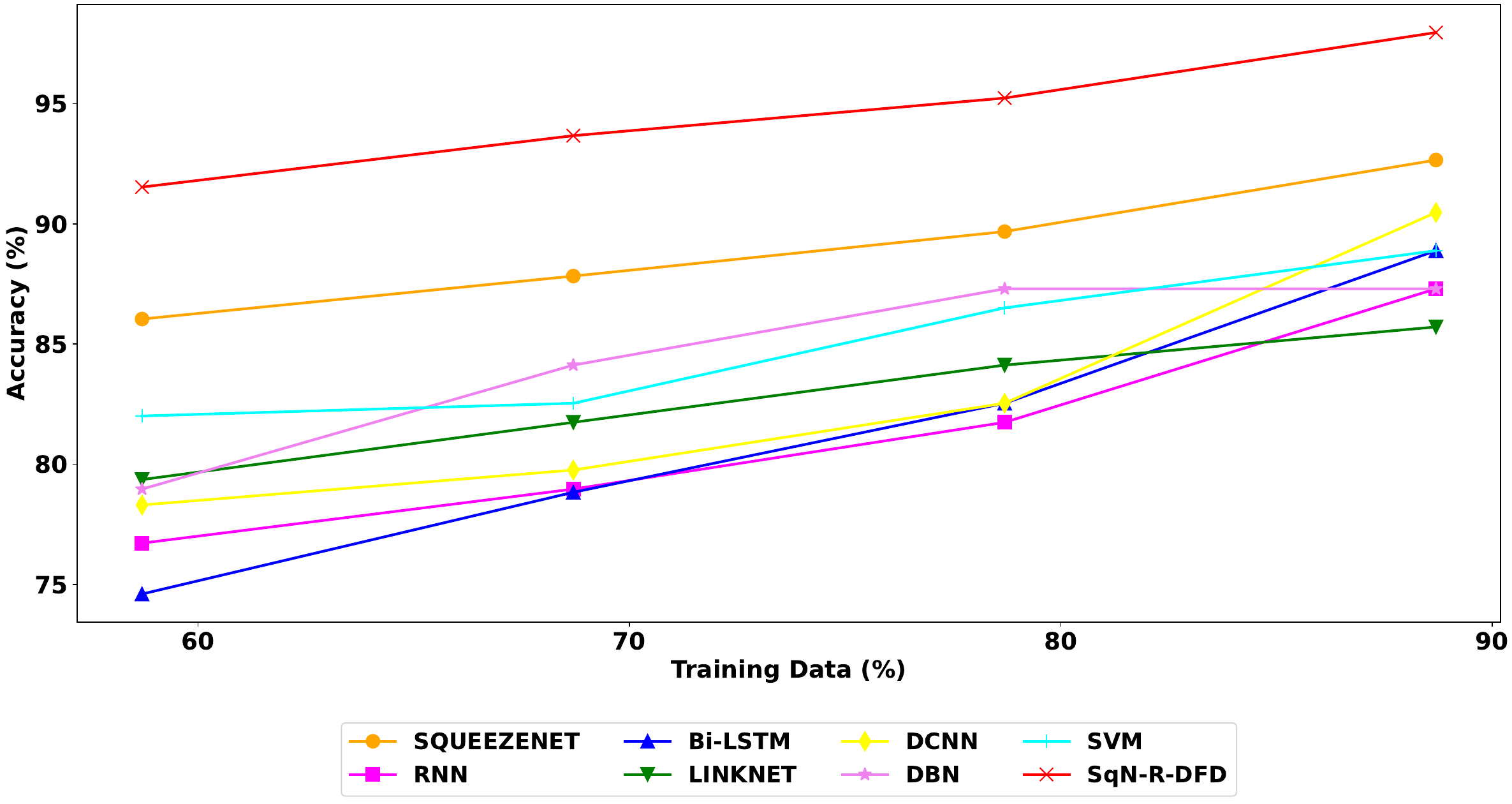}
        \caption{Accuracy for Compressed Case using Dataset2 \label{fig14d}}    
  \end{subfigure}
  \vspace{3mm}
    \begin{subfigure}{0.45\textwidth}
        \includegraphics[width=\linewidth]{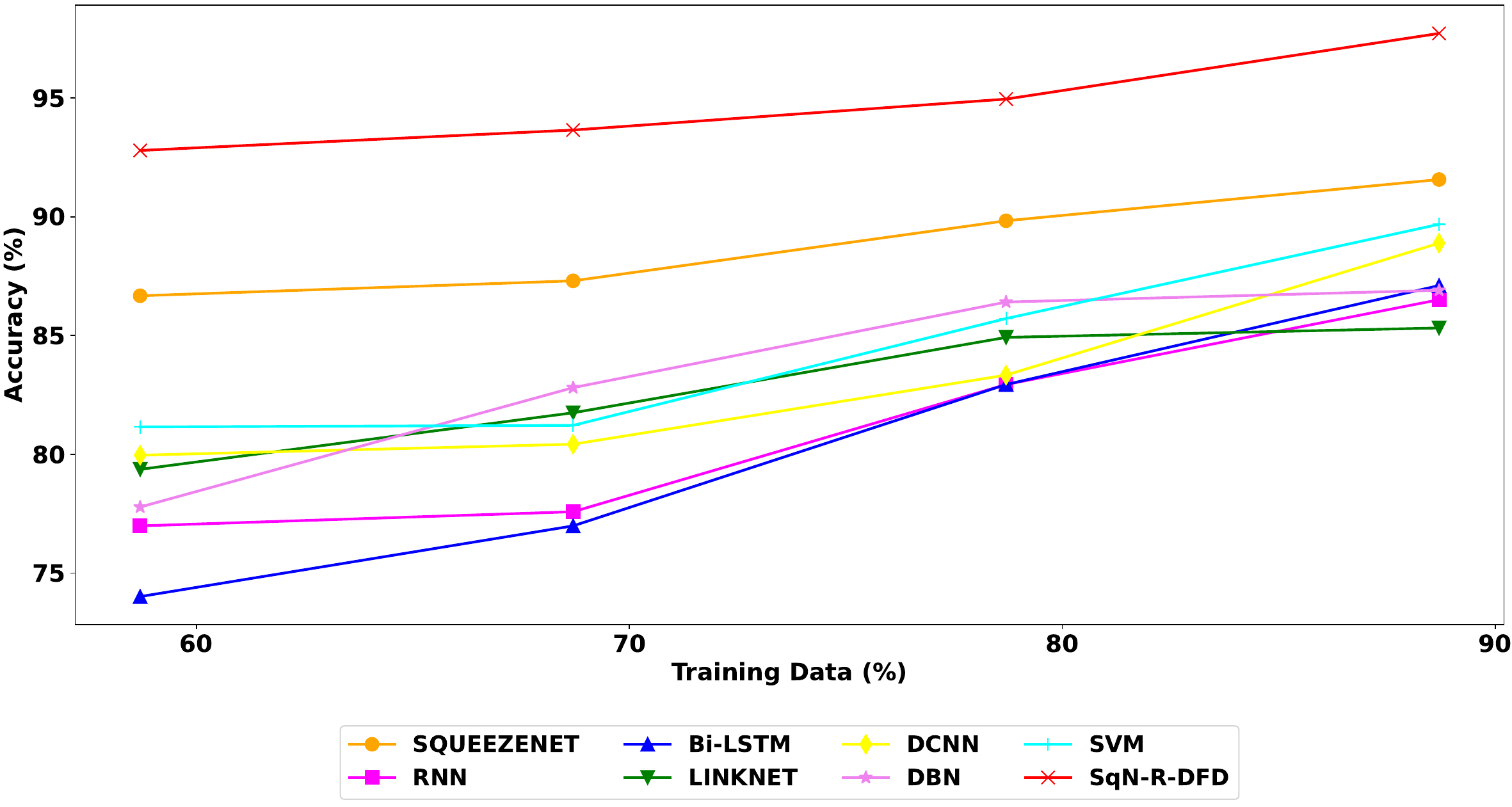}
        \caption{Accuracy for Wav2lip Case using Dataset2} \label{fig14e}
     \end{subfigure}%
    \hfill
     \begin{subfigure}{0.45\textwidth}
        \includegraphics[width=\linewidth]{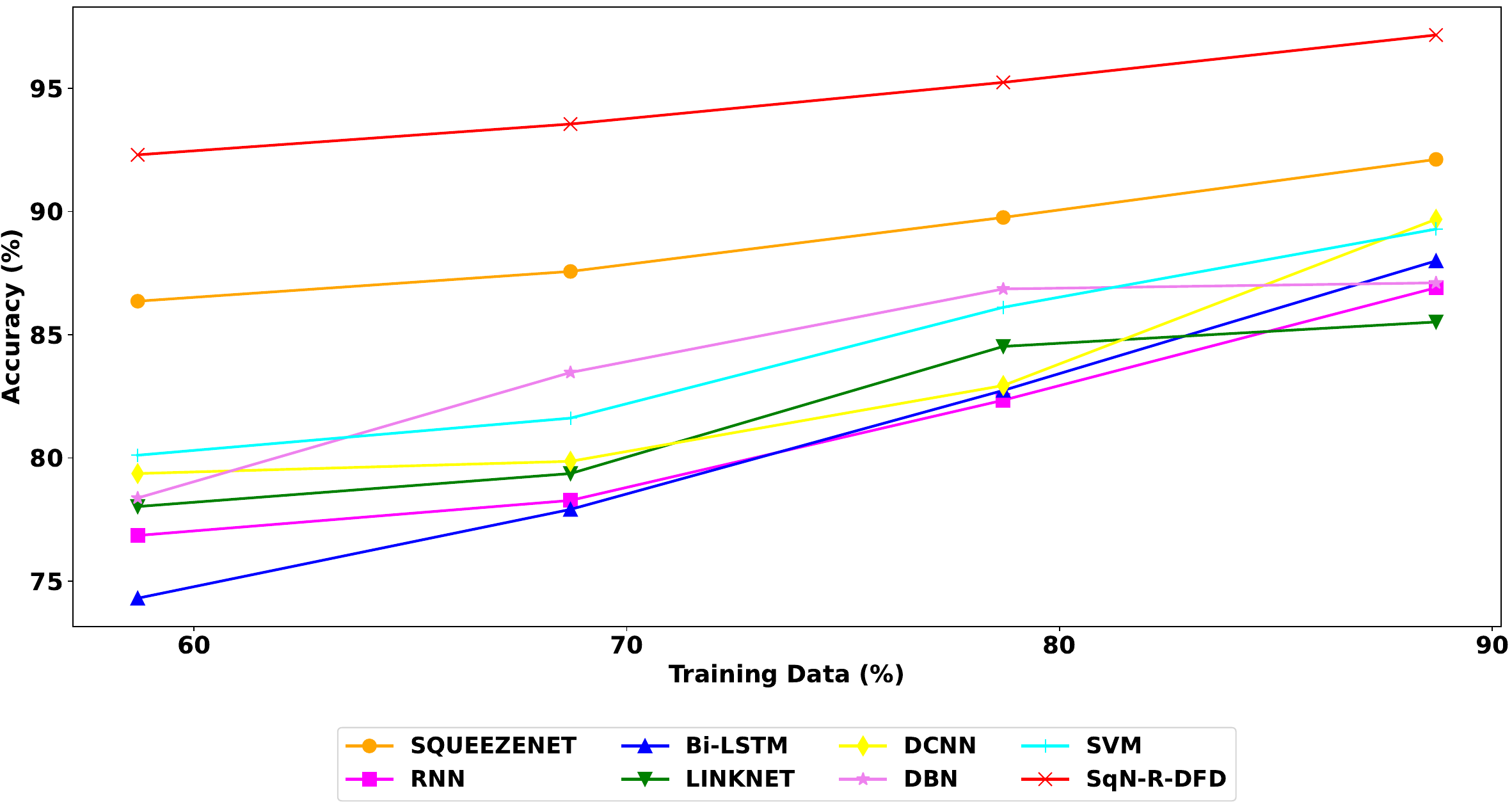}
      \caption{Accuracy for Faceswapped Case using Dataset2 \label{fig14f}}
    \end{subfigure}
         \begin{subfigure}{0.45\textwidth}
        \includegraphics[width=\linewidth]{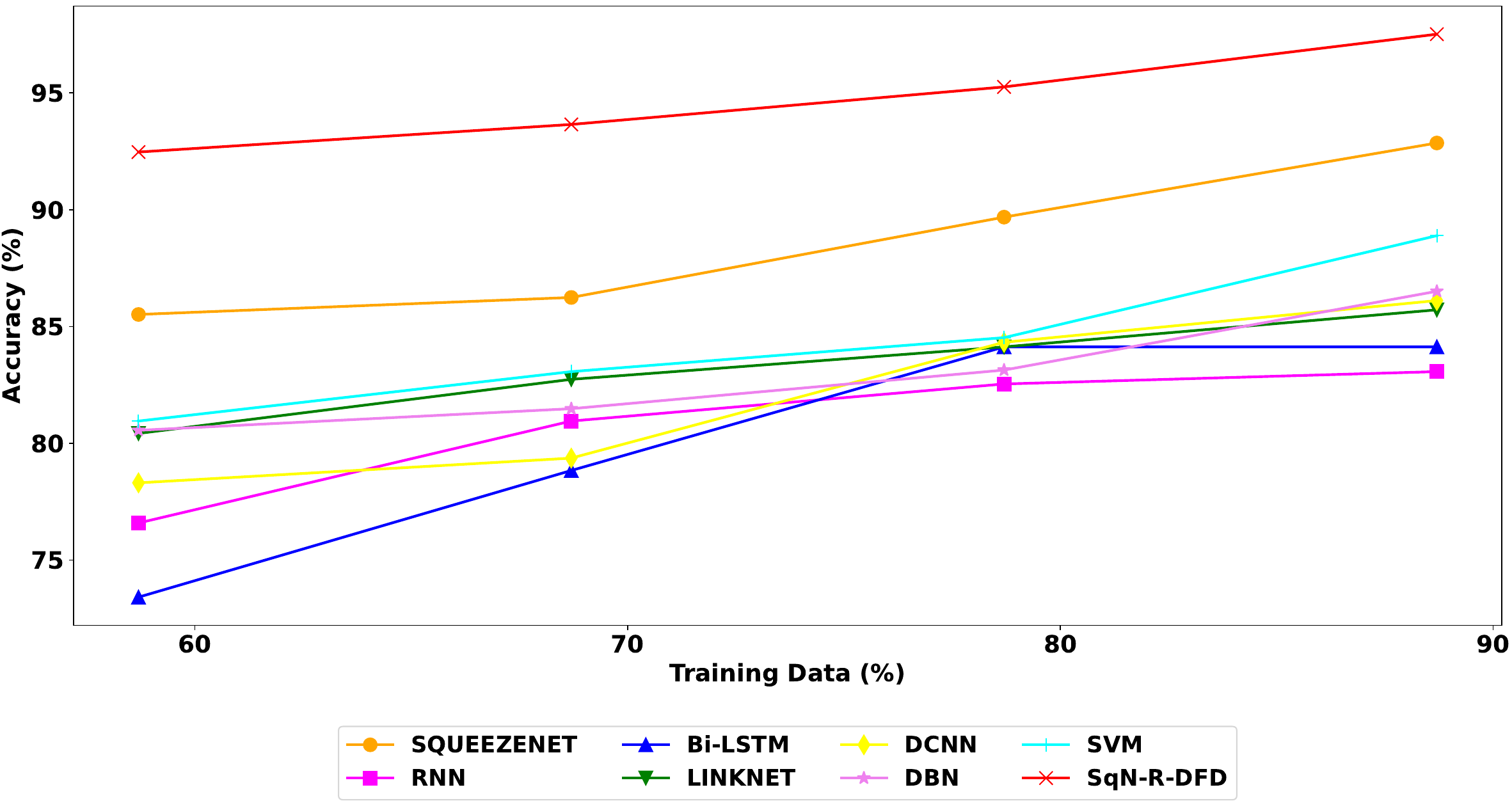}
        \caption{Accuracy for Resized Case using Dataset2  \label{fig14g}}   
        \end{subfigure}
\caption{Comparison on SqN-R-DFD Versus Conventional Methods for  Dataset2 (a)Accuracy for Original Case (b) Accuracy for Noise Case (c) Accuracy for Blurred Case (d) Accuracy for Compressed Case  (e)  Accuracy for Wav2lip Case (f) Accuracy for Faceswapped Case (g) Accuracy for Resized Case}
    \label{figaccuracynew}
    \end{figure}

\begin{figure}[!htbp]
    \centering   
    \begin{subfigure}[t]{0.45\textwidth}
        \includegraphics[width=\linewidth]{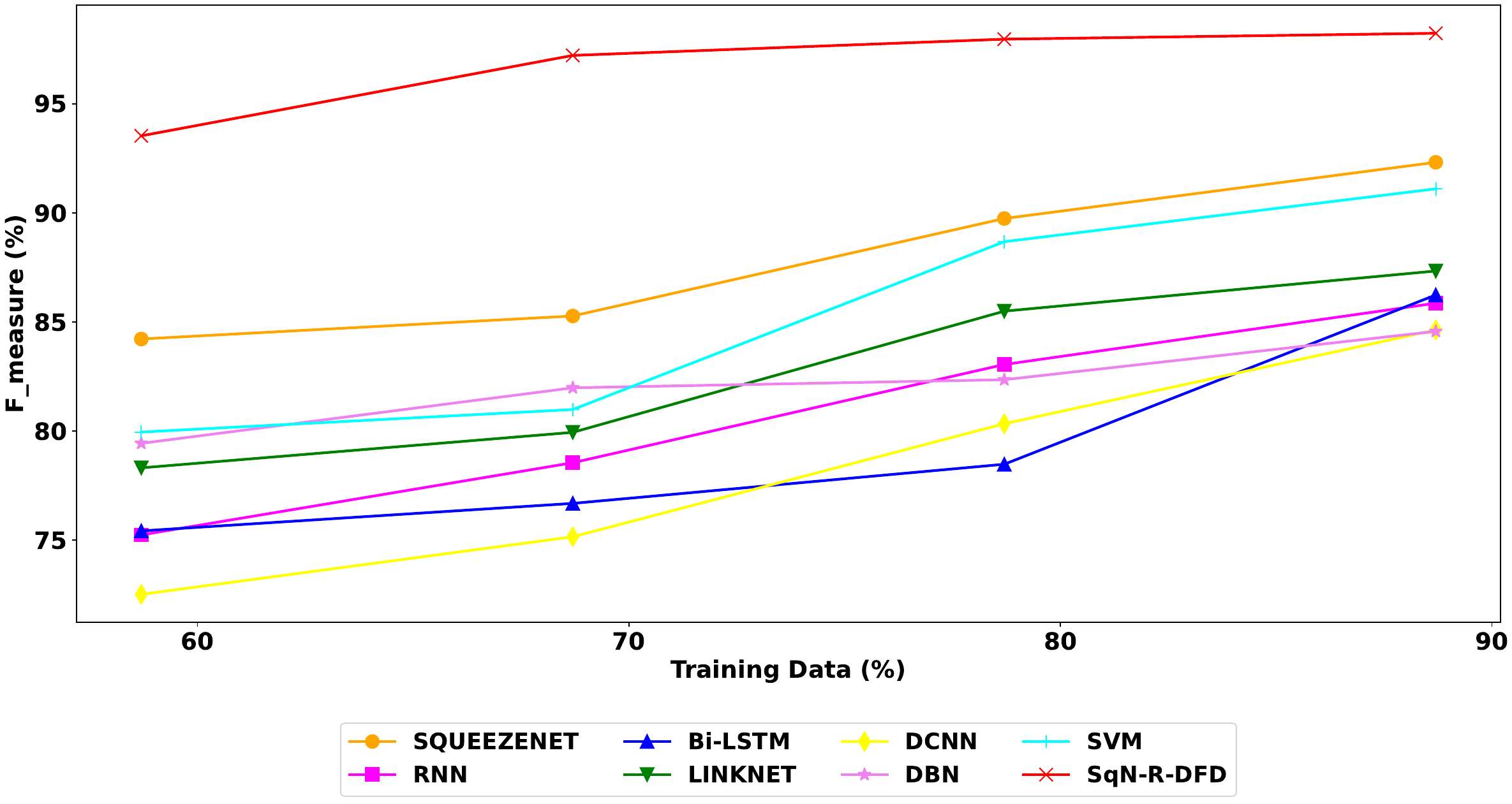}
        \caption{F-measure for Original Case using Dataset2 \label{figfmeasure_a}}    
    \end{subfigure}
    \hfill
    \begin{subfigure}[t]{0.45\textwidth}
        \includegraphics[width=\linewidth]{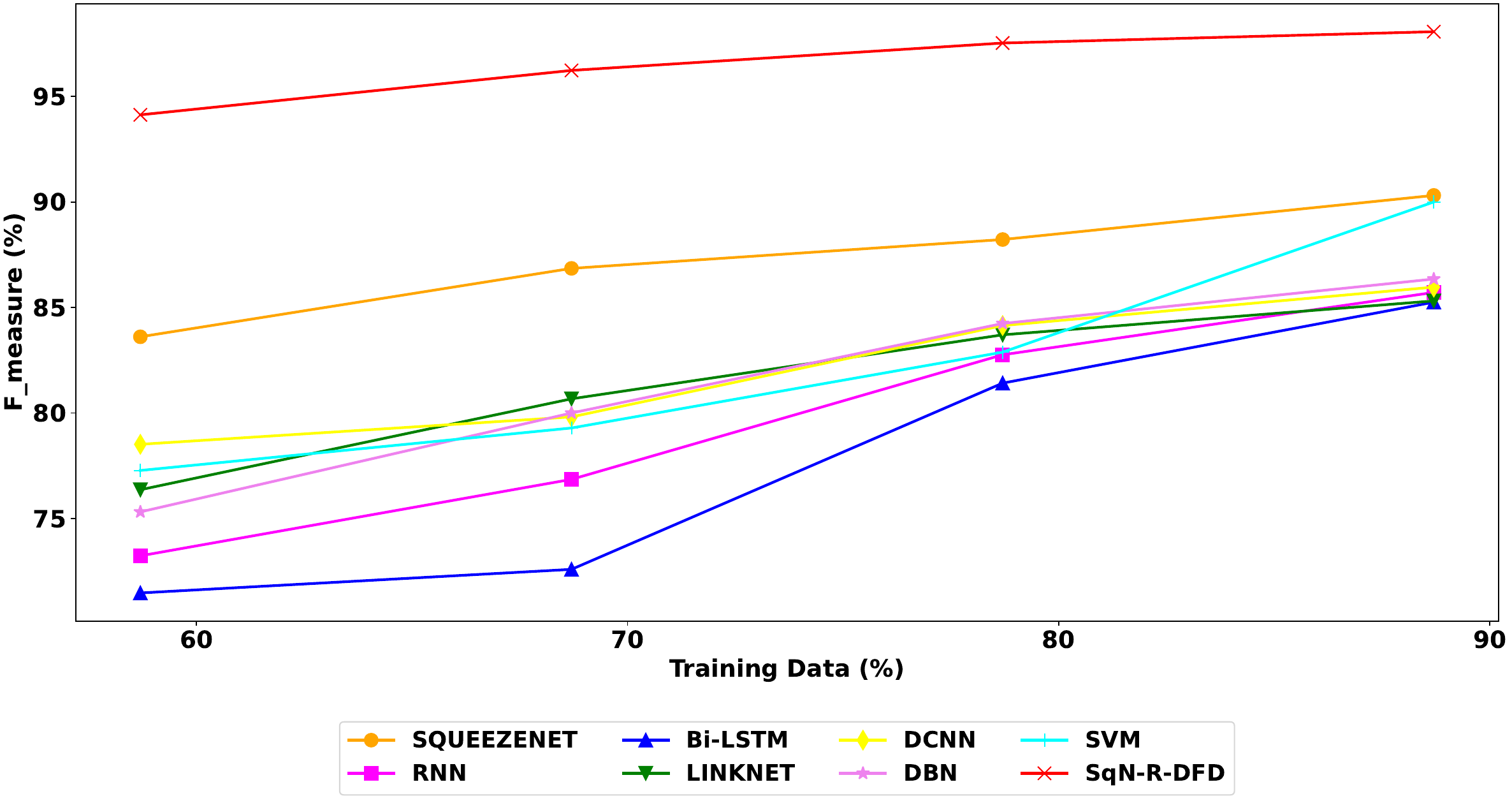}
        \caption{F-measure for Noise Case using Dataset2   \label{figfmeasure_b}}  
    \end{subfigure}
    \begin{subfigure}{0.45\textwidth}
        \includegraphics[width=\linewidth]{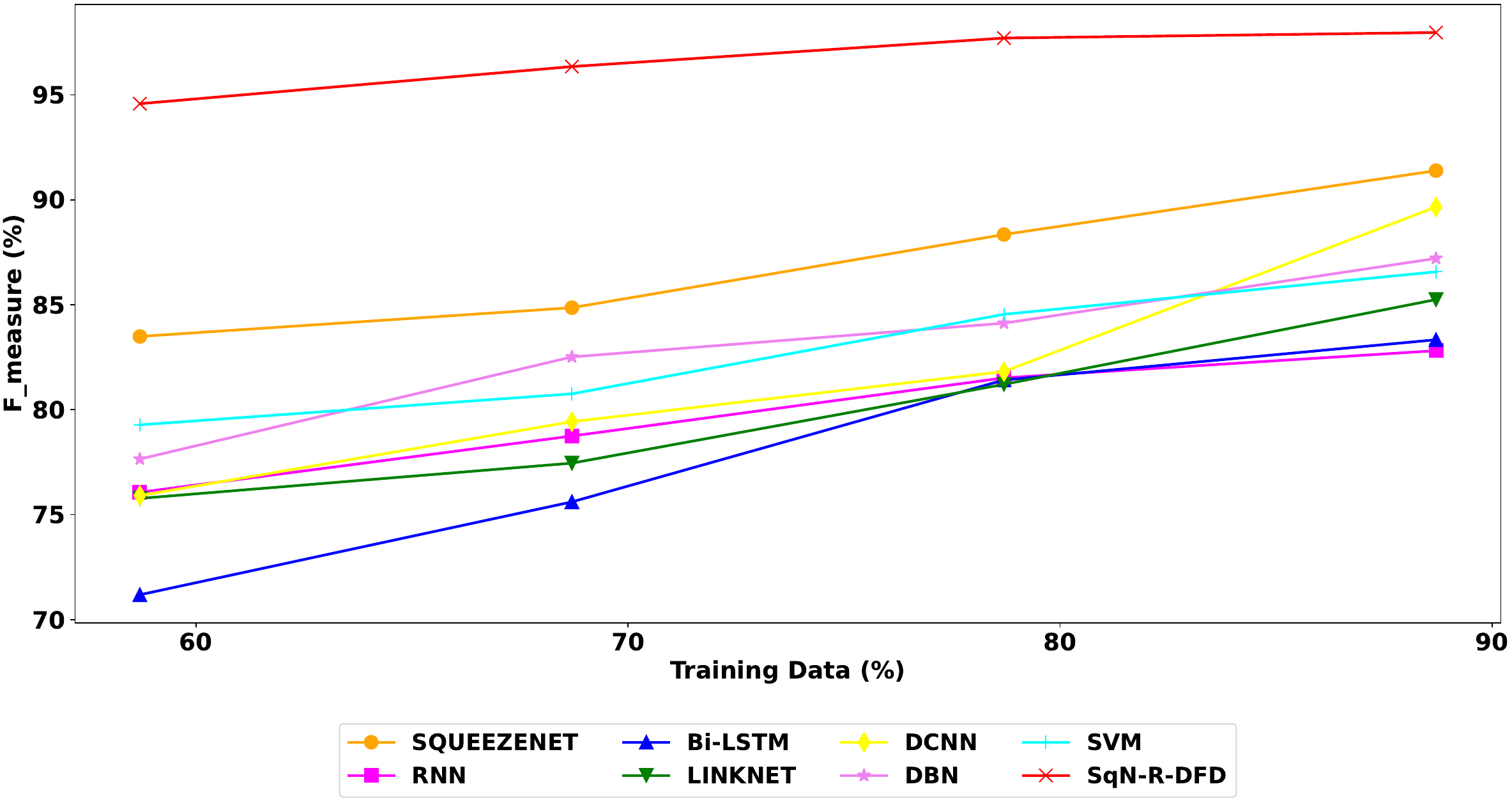}
        \caption{F-measure for Blurred Case using Dataset2   \label{figfmeasure_c}}

    \end{subfigure}
    \hfill
    \begin{subfigure}{0.45\textwidth}
        \includegraphics[width=\linewidth]{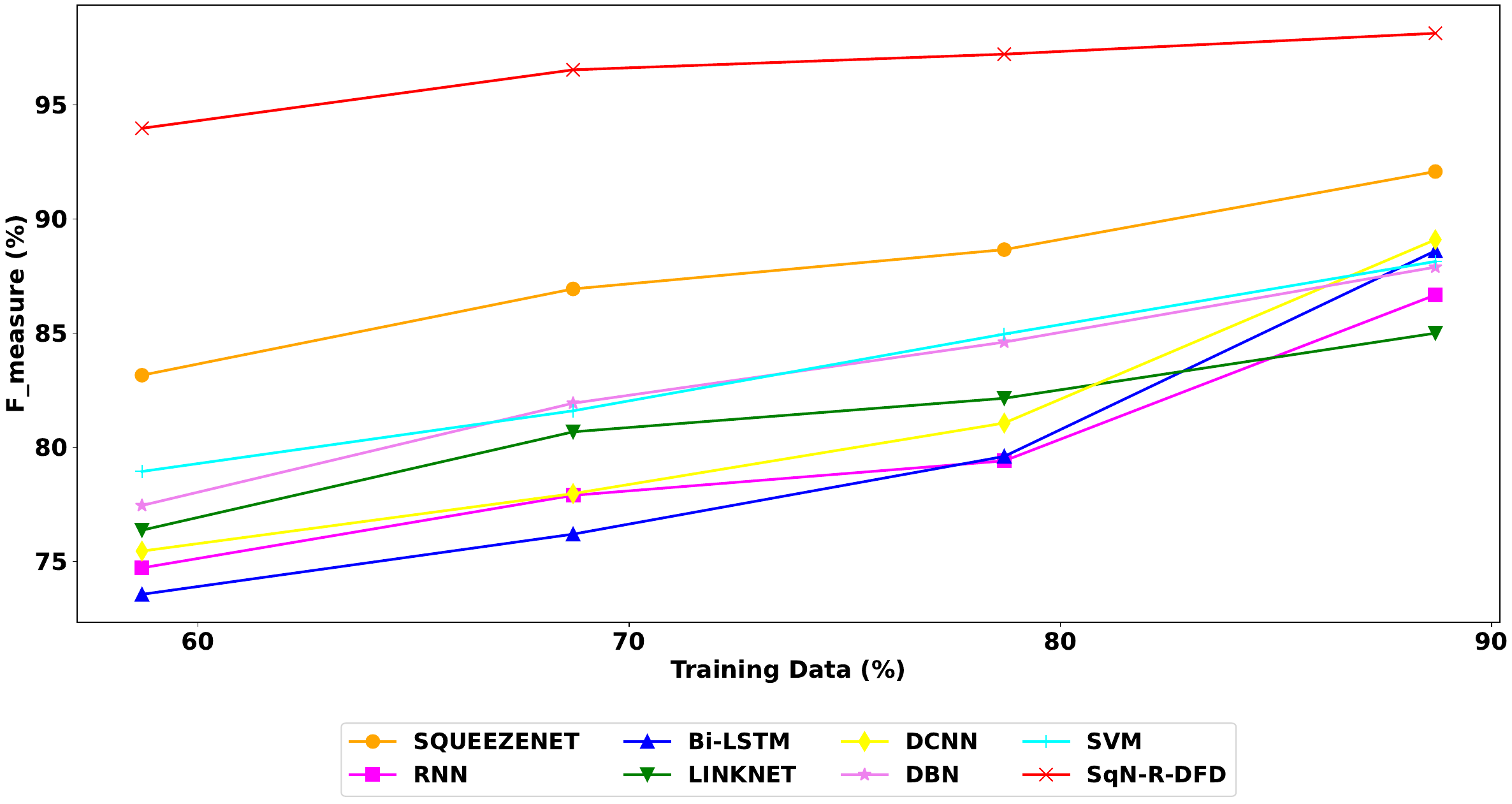}
        \caption{F-measure for Compressed Case using Dataset2 \label{figfmeasure_d}}

    \end{subfigure}
    \begin{subfigure}{0.45\textwidth}
        \includegraphics[width=\linewidth]{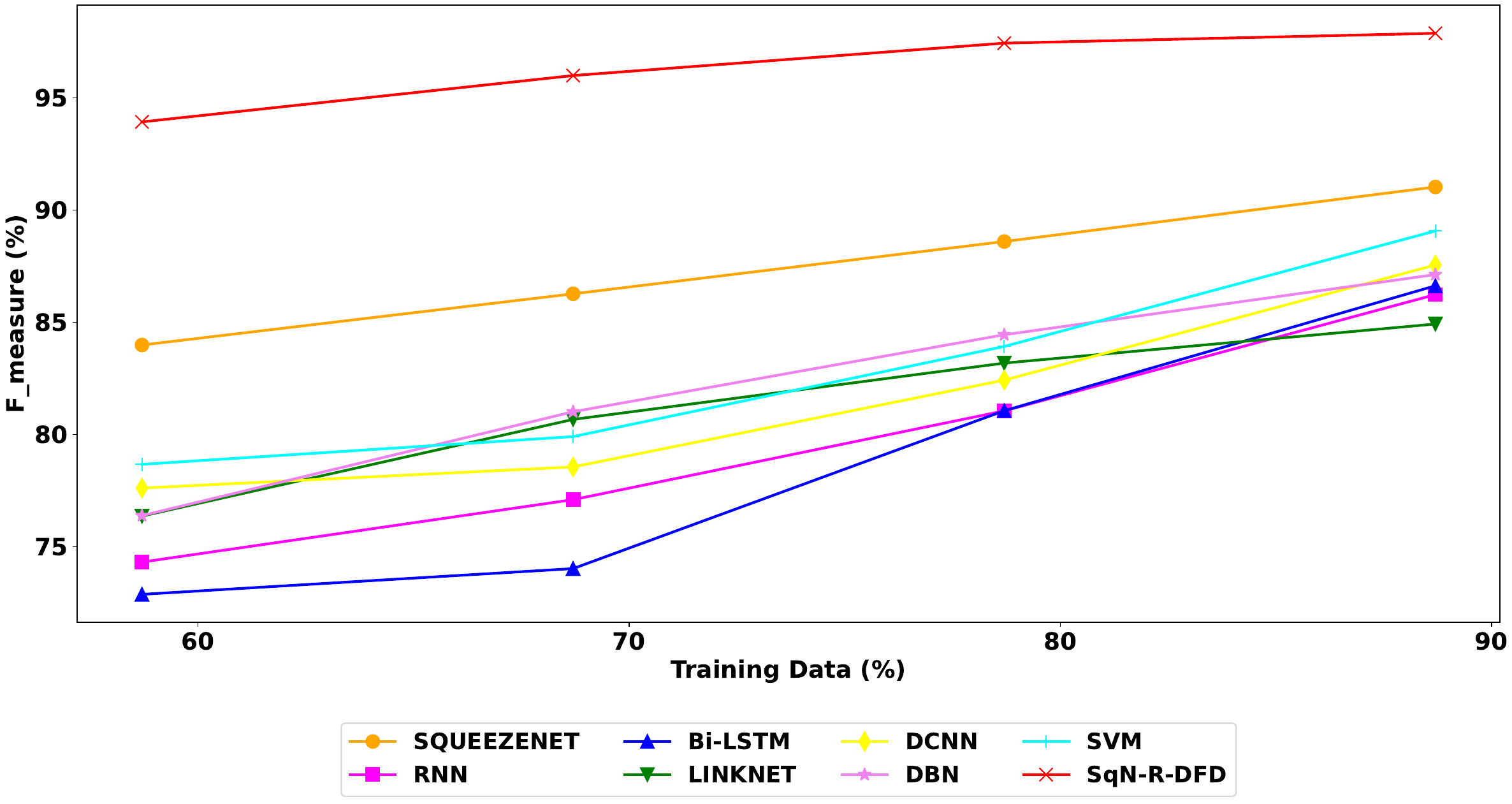}
        \caption{F-measure for Wav2lip Case using Dataset2 \label{figfmeasure_e}}
    \end{subfigure}
     \hfill
    \begin{subfigure}{0.45\textwidth}
        \includegraphics[width=\linewidth]{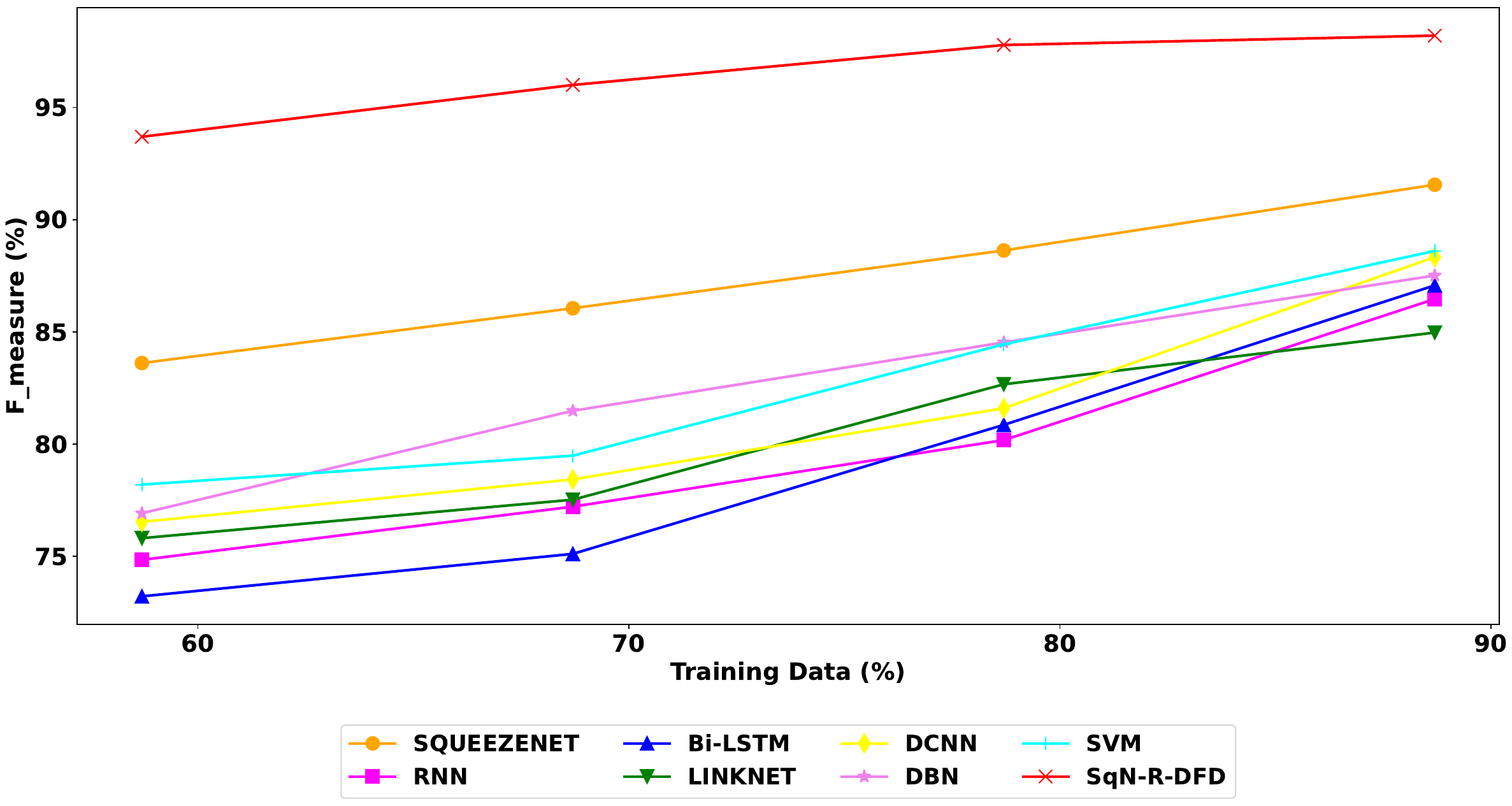}
        \caption{F-measure for Faceswapped Case using Dataset2 \label{figfmeasure_f}}
    \end{subfigure}
    \begin{subfigure}{0.45\textwidth}
        \includegraphics[width=\linewidth]{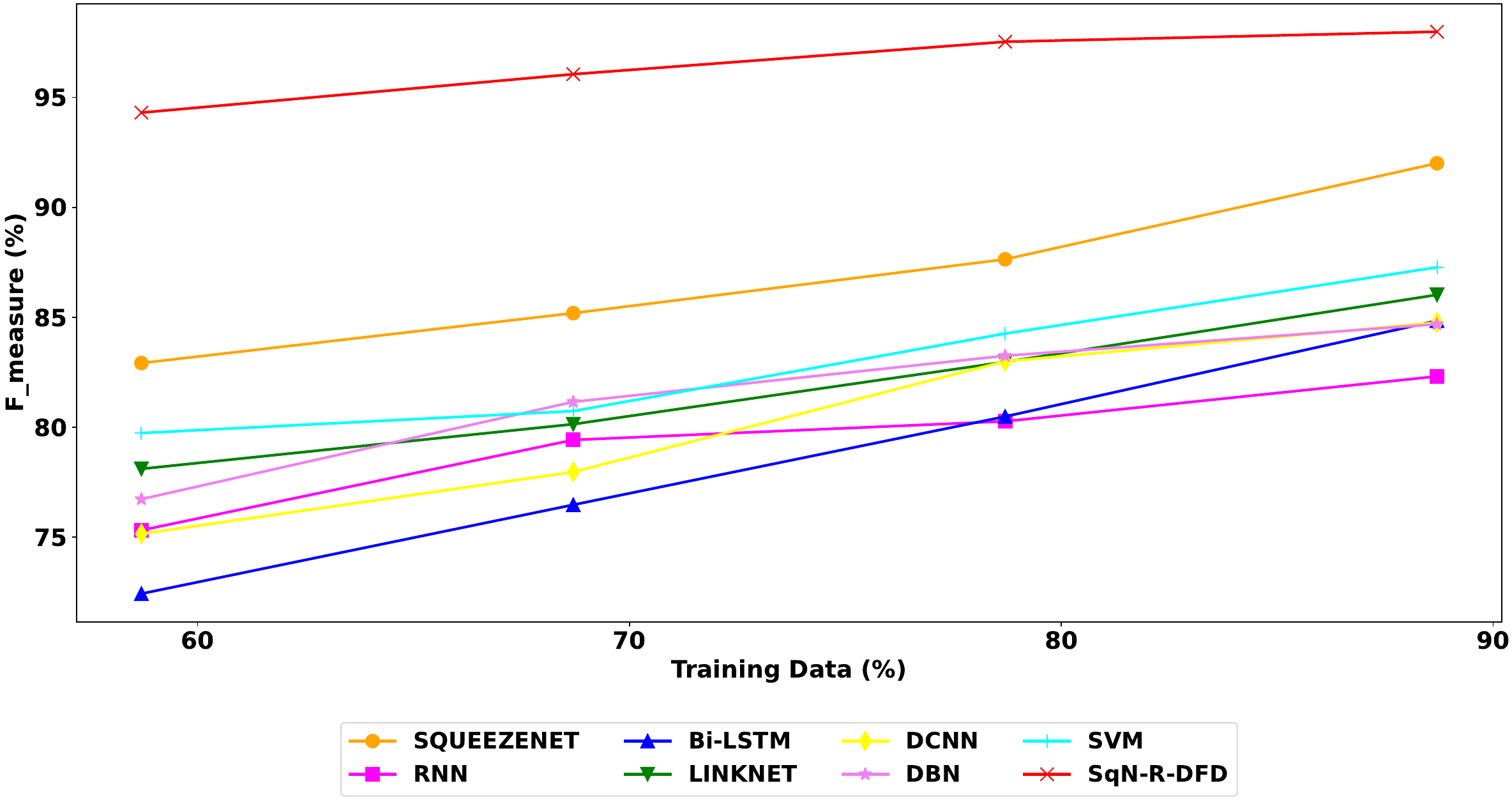}
        \caption{F-measure for Resized Case using Dataset2 \label{figfmeasure_g}}
    \end{subfigure}
      \caption{Comparison on SqN-R-DFD Versus Conventional Methods for  Dataset2 (a) F-Measure for Original Case (b)  F-Measure for Noisy Case (c)  F-Measure for Blurred Case (d) F-Measure for Compressed Case (e) F-Measure for Wav2lip Case (f) F-Measure for Faceswapped Case (g) F-Measure for Resized Case}
      \label{figfmeasure}
\end{figure}

\textbf{Accuracy:} Figure \ref{fig14a}, the SqN-R-DFD model consistently achieved the highest accuracy of $97.348$, whereas the SqueezeNet, RNN, Bi-LSTM, LinkNet, DCNN, DBN \cite{suganthi2022deep}, and SVM \cite{tu2024face} scored the least accuracy, ranging from $83.845$ to $89.786$ at $90\%$ training data. This indicates the effective performance of the SqN-R-DFD approach in deepfake detection. In Figure \ref{fig14c}, the SqN-R-DFD approach attained the accuracy score of $95.473$ in $80\%$ training data, which is extremely higher than the conventional methods. The F-measure obtained by the SqN-R-DFD model is $98.432$ in $90\%$ training data, while SqueezeNet, RNN, Bi-LSTM, LinkNet, DCNN, DBN \cite{suganthi2022deep}, and SVM \cite{tu2024face} generated the least F-measure values of $89.543$, $86.512$, $87.845$, $83.143$, $85.917$, $83.915$, and $84.943$, respectively, in Figure \ref{figfmeasure_f}.

\textbf{F-Measure:} Regarding the F-measure metric at Figure \ref{figfmeasure_b}, the conventional methods exhibited the least F-measure scores across all training data, while the SqN-R-DFD scheme recorded the F-measure of $93.643$ at $60\%$ training data, $94.845$ at $70\%$ training data, $96.842$ at $80\%$ training data, and $98.518$ at $90\%$ training data, respectively. In addition, the ROC achieved by the SqN-R-DFD strategy is higher than the conventional methods, signifying better classification performance. In the SqN-R-DFD approach, the proximity of training and testing loss signifies robust performance and good generalization ability.

\subsubsection{\textbf{ROC Curve Analysis}} \label{roc2}
\begin{figure}[!htbp]
    \centering
    \begin{subfigure}[t]{0.45\textwidth}
        \includegraphics[width=\linewidth]{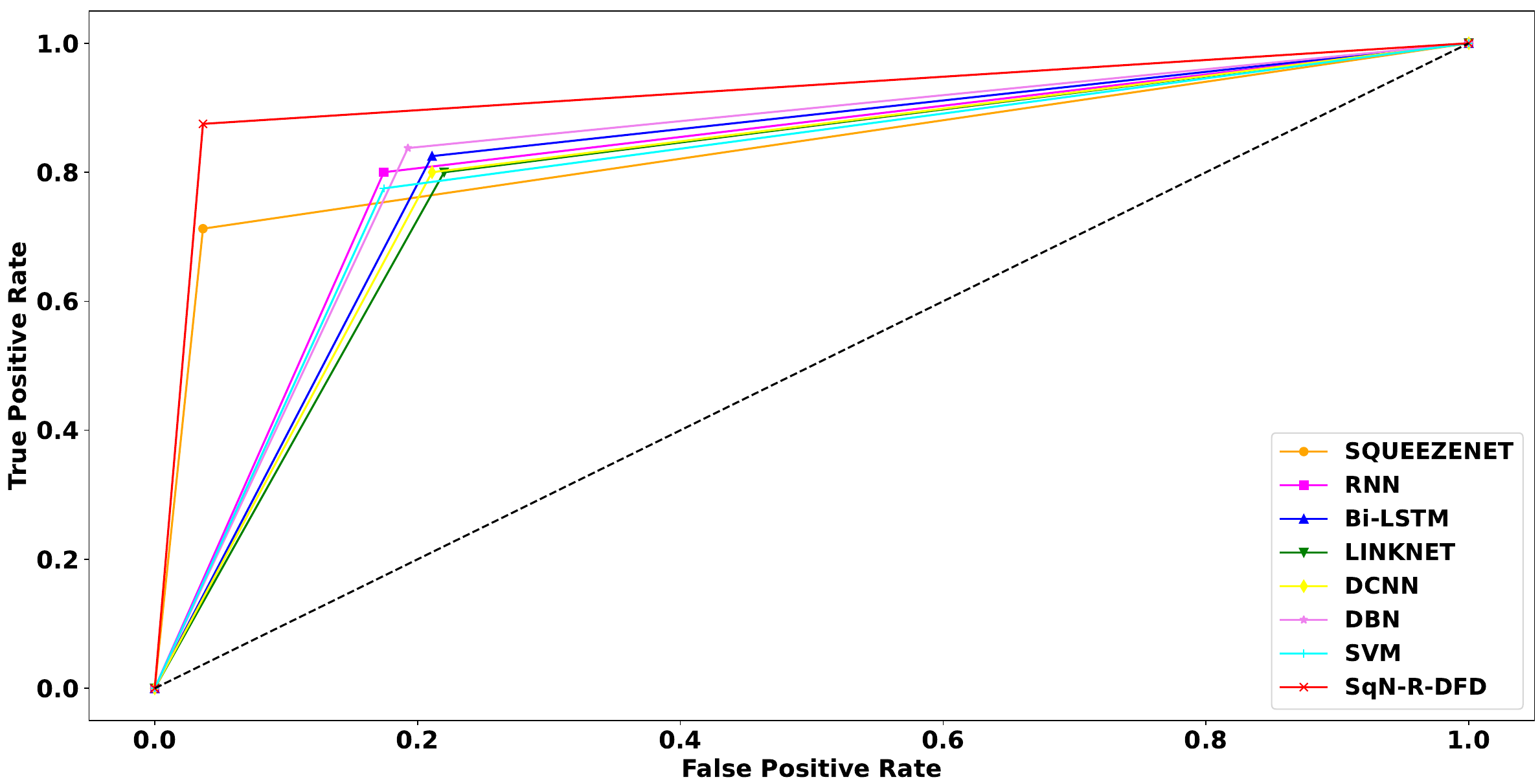}
        \caption{ROC Curve for Original Case using Dataset2\label{rocoriginaldataset2}}
    \end{subfigure}
    \hfill
    \begin{subfigure}[t]{0.45\textwidth}
        \includegraphics[width=\linewidth]{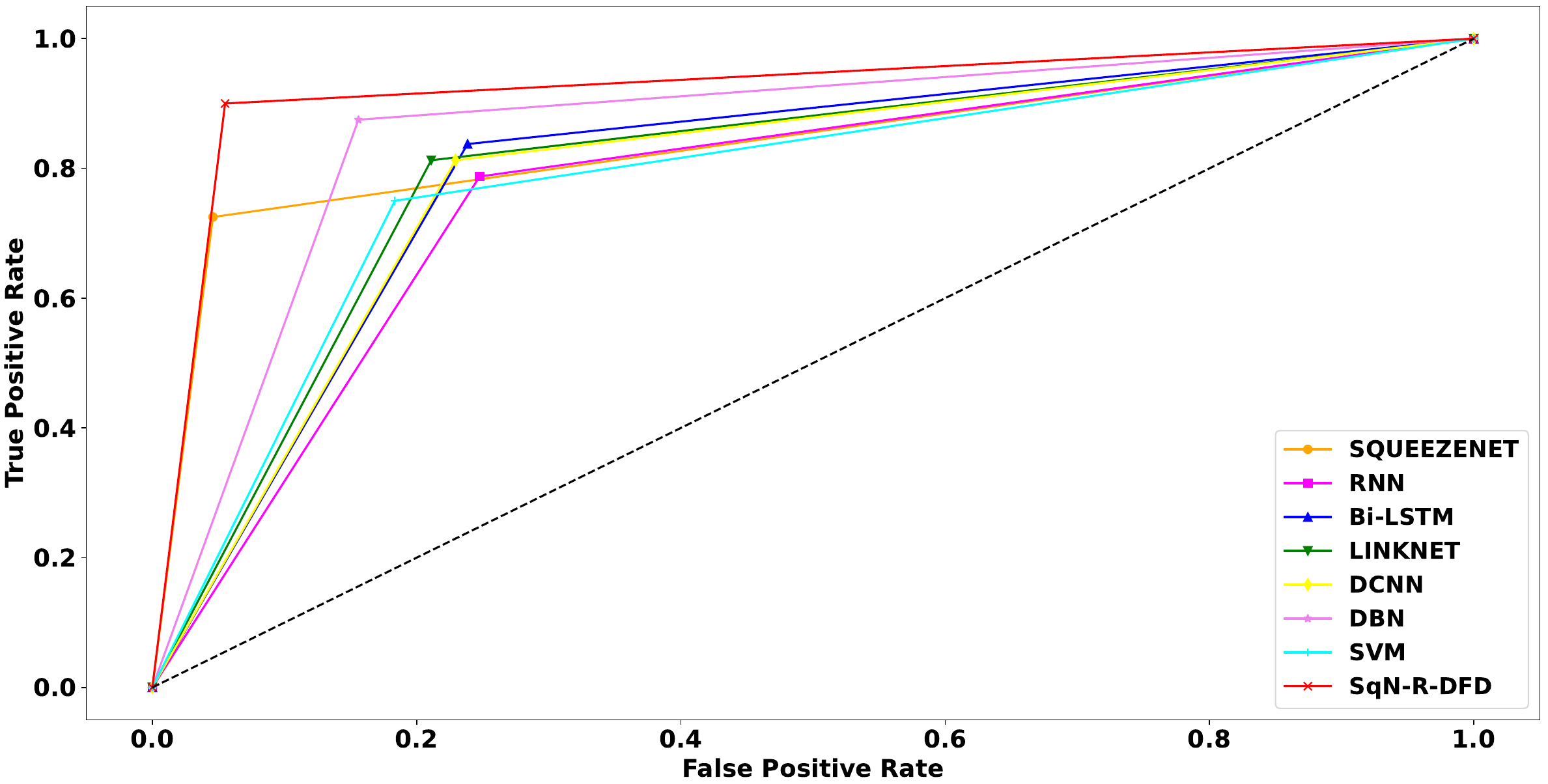} 
        \caption{ROC Curve for Noise Case using Dataset2\label{rocnoisydataset2}}
    \end{subfigure} 
    \begin{subfigure}{0.45\textwidth}
        \includegraphics[width=\linewidth]{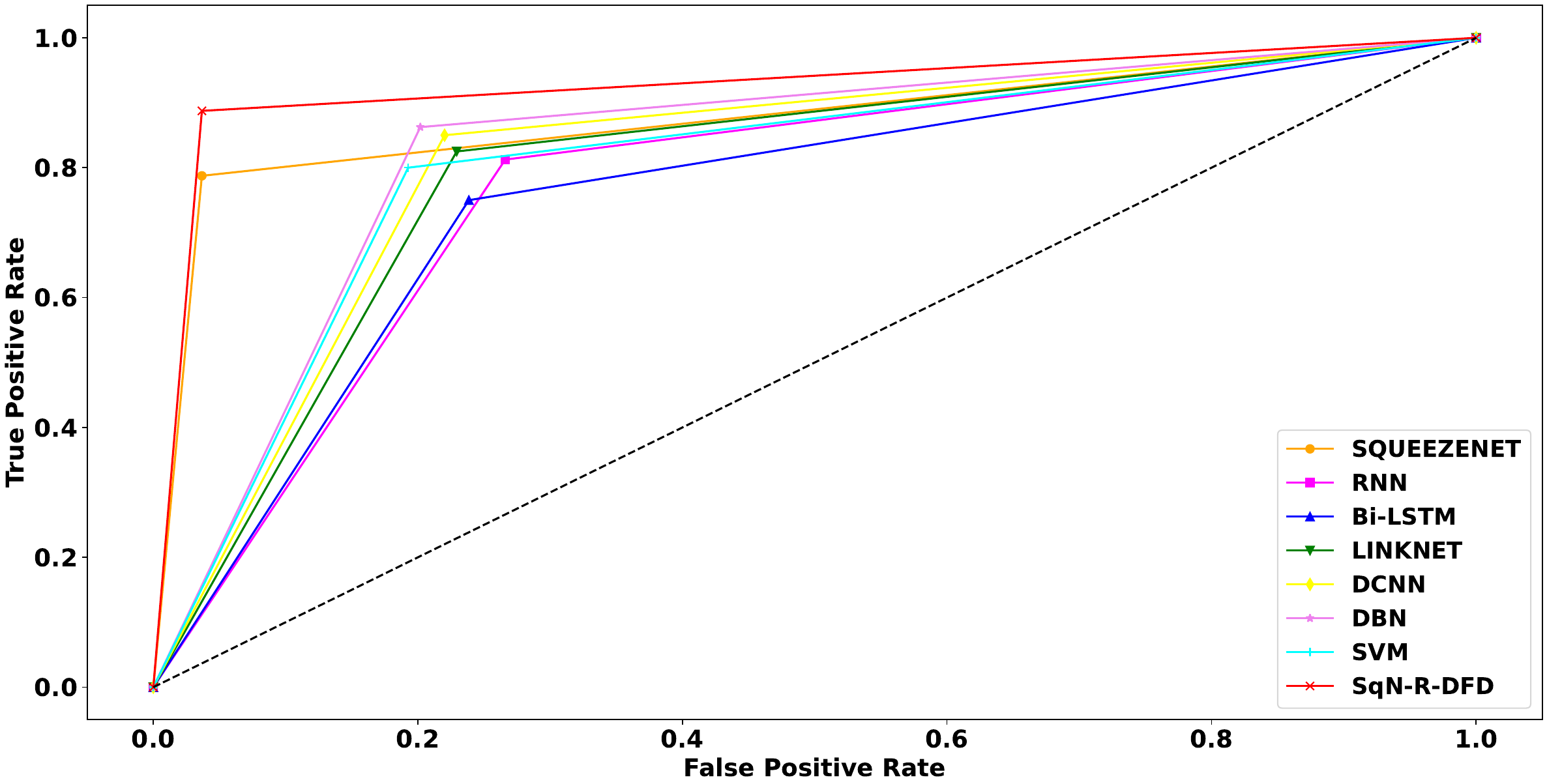} 
        \caption{ROC Curve for Blurred Case using Dataset2 \label{rocblurreddataset2}}
    \end{subfigure}
    \hfill
    \begin{subfigure}{0.45\textwidth}
        \includegraphics[width=\linewidth]{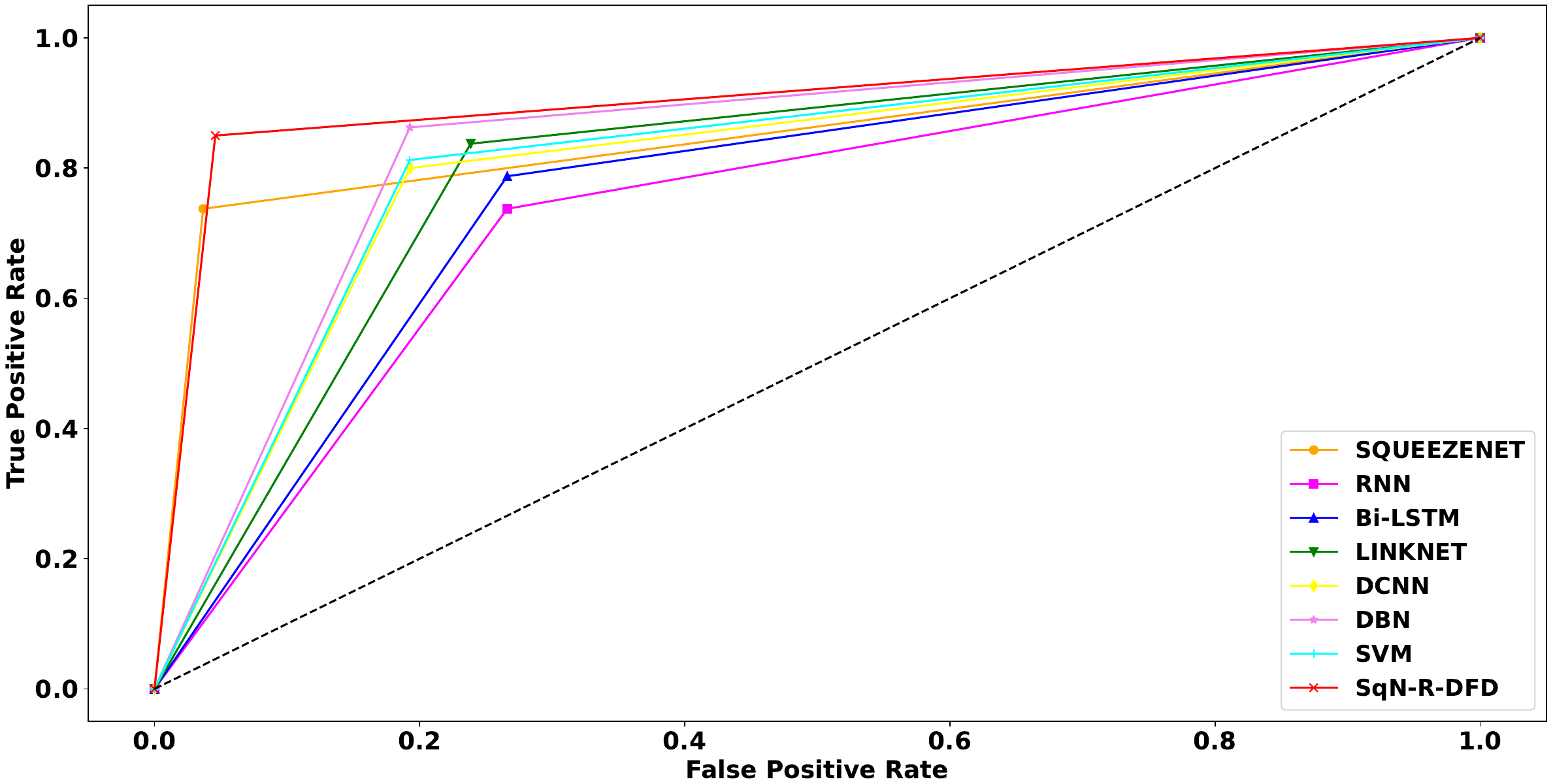} 
        \caption{ROC Curve for Compressed Case using Dataset2\label{roccompresseddataset2}}
    \end{subfigure}   
    \begin{subfigure}{0.45\textwidth}
        \includegraphics[width=\linewidth]{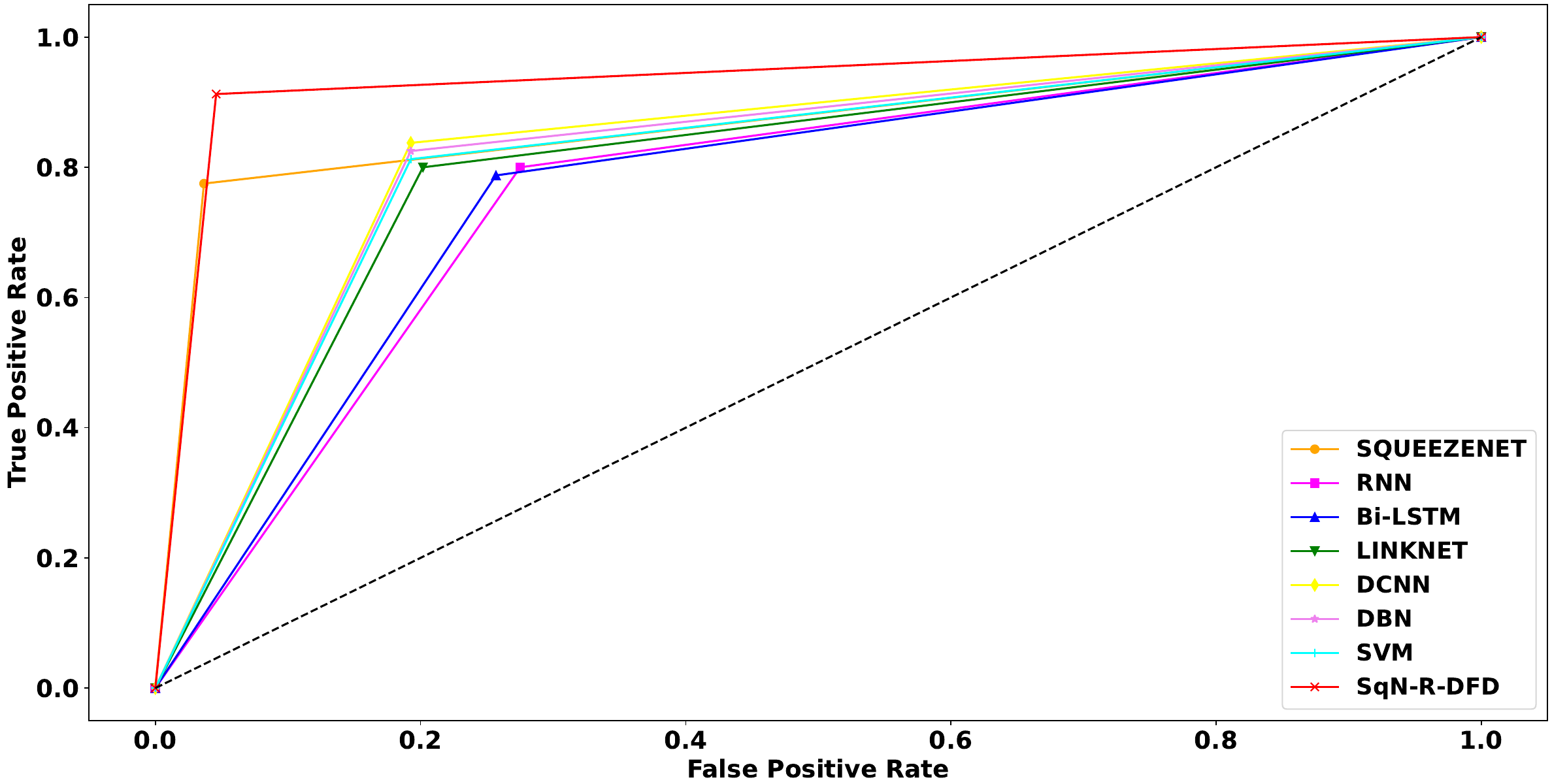}
        \caption{ROC Curve for Wave2lip Case using Dataset2 \label{rocwav2lipdataset2}} 
    \end{subfigure}
    \hfill
    \begin{subfigure}{0.45\textwidth}
        \includegraphics[width=\linewidth]{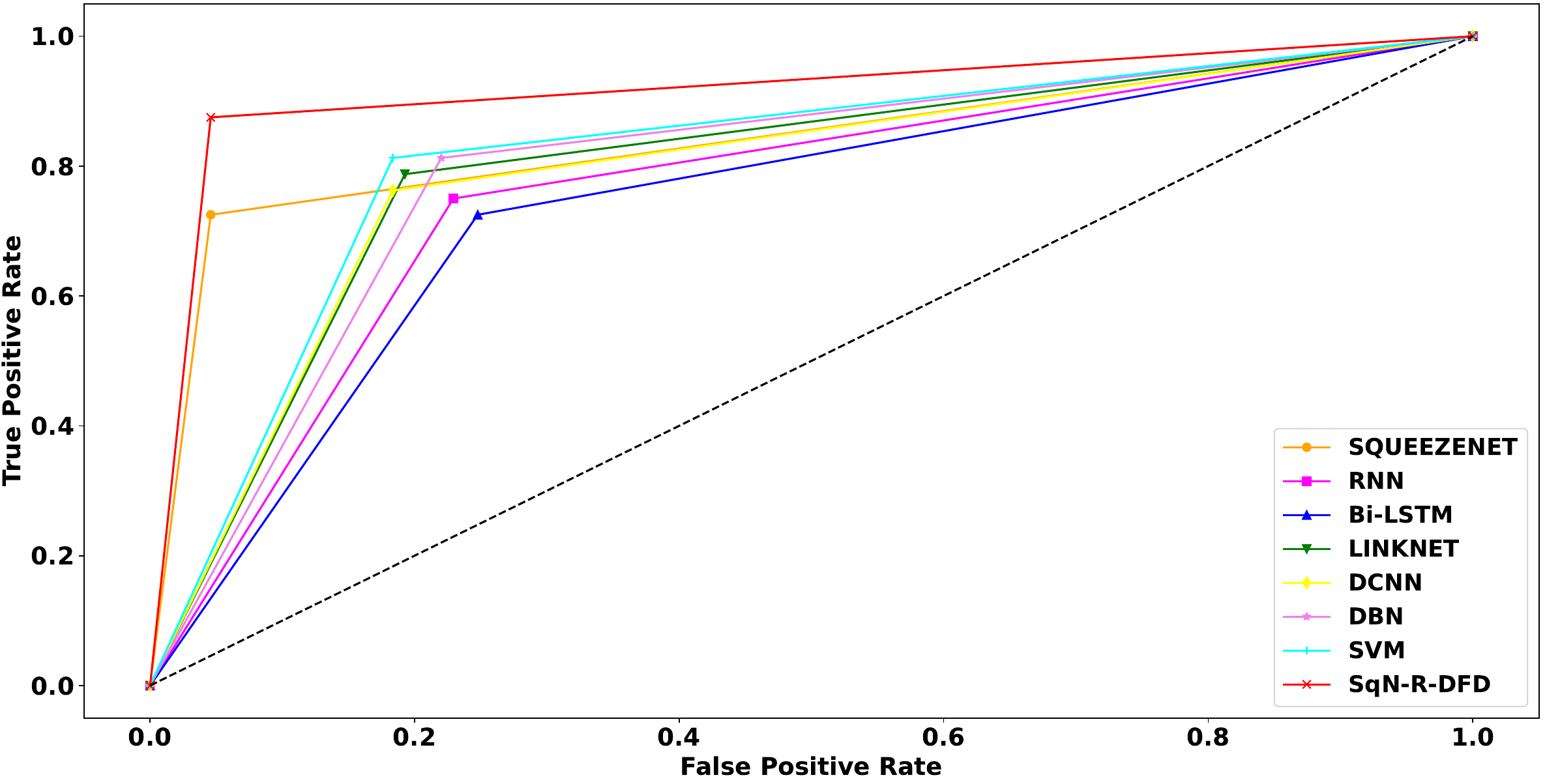}
        \caption{ROC Curve for Faceswapped Case using Dataset2\label{rocfaceswappeddataset2}}
    \end{subfigure}    
    \begin{subfigure}{0.45\textwidth}
        \includegraphics[width=\linewidth]{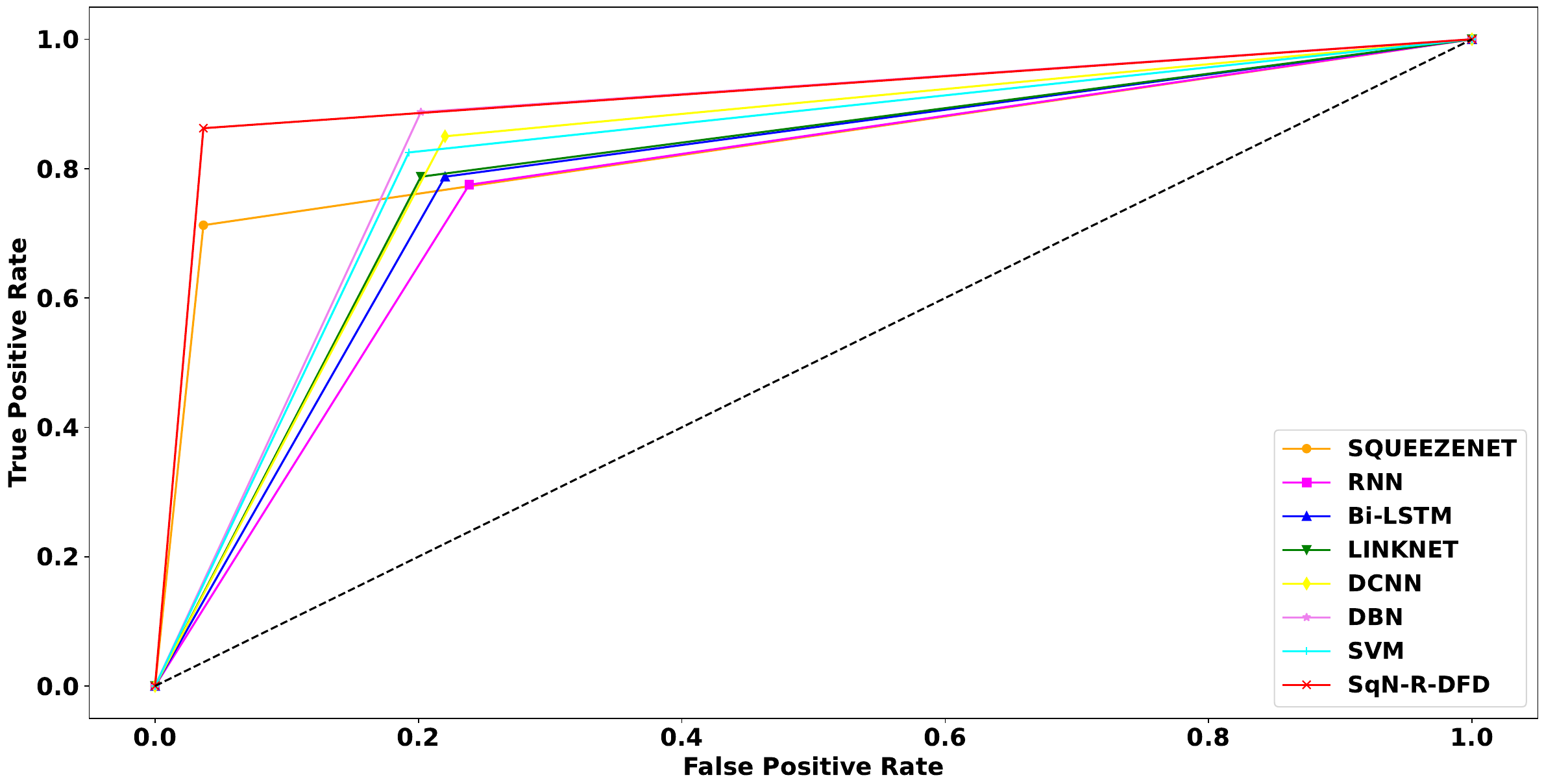}
        \caption{ROC Curve for Resized Case using Dataset2\label{rocresizeddataset2}}
    \end{subfigure}
    \caption{Analysis on ROC Curve using Dataset2 (a) ROC Curve for Original Case (b) ROC Curve for Noisy Case (c) ROC Curve for Blurred Case (d) ROC Curve for Compressed Case (e) ROC Curve for Wav2lip Case (f) ROC Curve for Faceswapped Case (g) ROC Curve for Resized Case}
    \label{figroc10}
    \end{figure}

Figure \ref{rocoriginaldataset2} depicts the ROC curve analysis on the SqN-R-DFD scheme and conventional methods for deepfake detection. The SqN-R-DFD approach recorded the highest AUC score of $0.917$, signifying that it has a superior capability to differentiate between positive and negative classes. Conversely, the conventional methods such as SqueezeNet, RNN, Bi-LSTM, LinkNet, DCNN, DBN \cite{suganthi2022deep}, and SVM \cite{tu2024face} demonstrated an AUC score of greater than $0.7$, indicating better classification but not surpassing the SqN-R-DFD scheme. The highest AUC value is achieved using the SqN-R-DFD model is $0.952$, indicating an excellent performance. In comparison, the SqueezeNet, RNN, Bi-LSTM, LinkNet, DCNN, DBN \cite{suganthi2022deep}, and SVM \cite{tu2024face} demonstrated an AUC value ranging from $0.785$ to $0.825$, signifying acceptable performance at Figure \ref{roccompresseddataset2}.

\subsubsection{\textbf{Training-testing loss Analysis }} \label{training2}
\begin{figure}[!htbp]
    \centering  
    \begin{subfigure}{0.45\textwidth}
        \includegraphics[width=\linewidth]{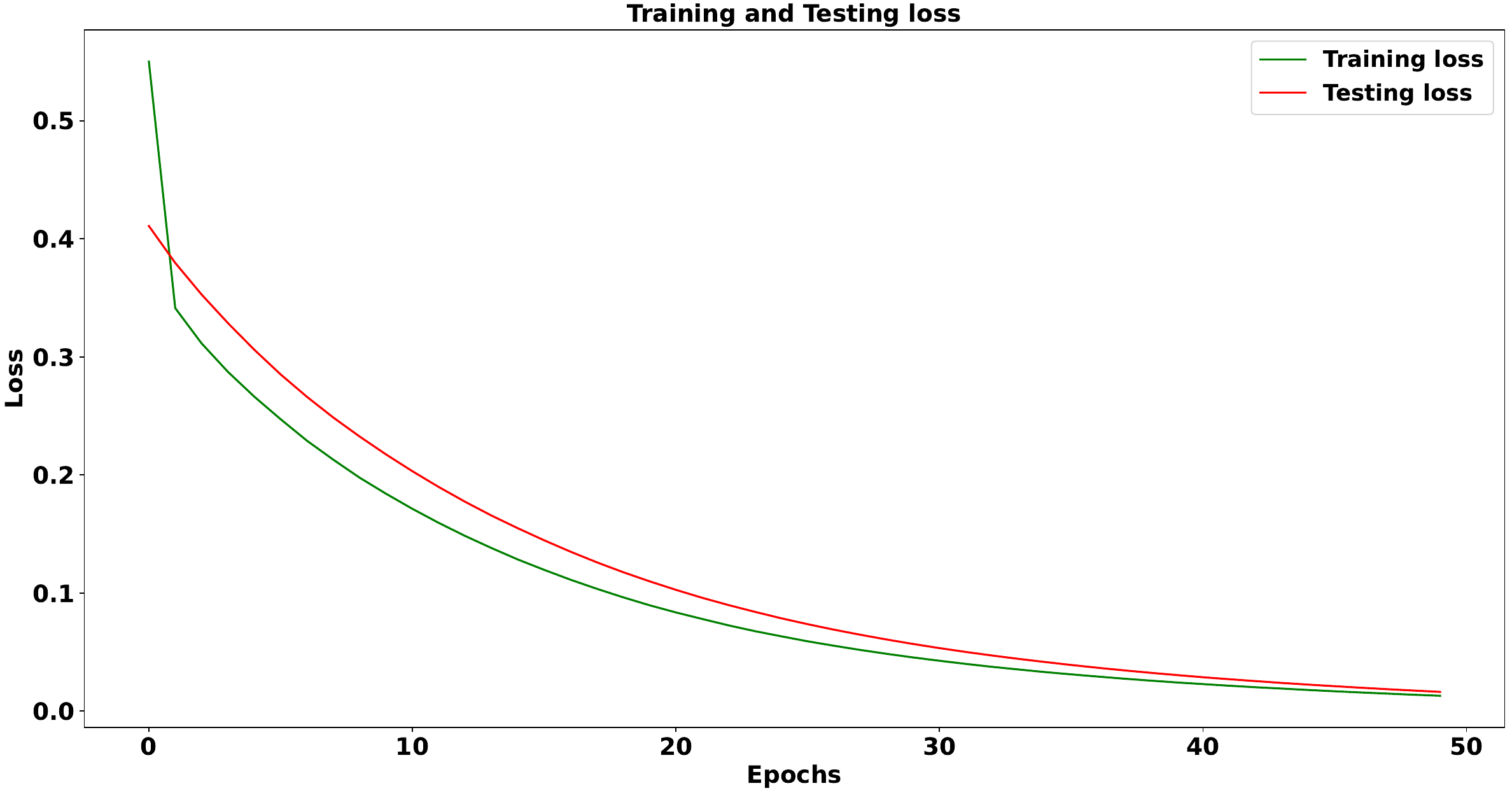} 
        \caption{Training-Testing Loss Curve for Original Case using Dataset2 \label{traininglossoriginaldataset2}}
    \end{subfigure}
    \hfill
    \begin{subfigure}{0.45\textwidth}
        \includegraphics[width=\linewidth]{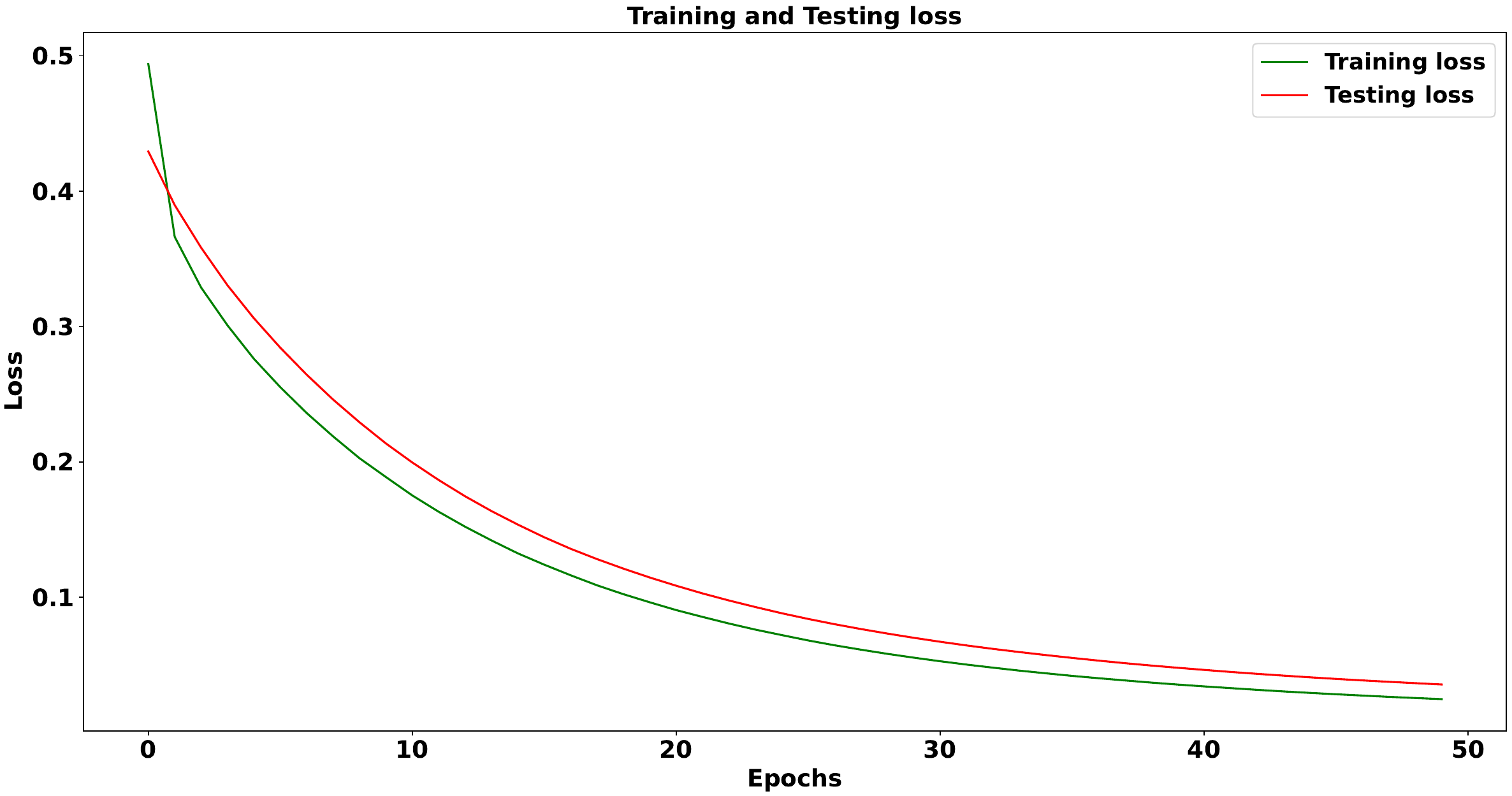}
        \caption{Training-Testing Loss Curve for Noise Case using Dataset2\label{trainingnoisydataset2}}
    \end{subfigure} 
    \begin{subfigure}{0.45\textwidth}
        \includegraphics[width=\linewidth]{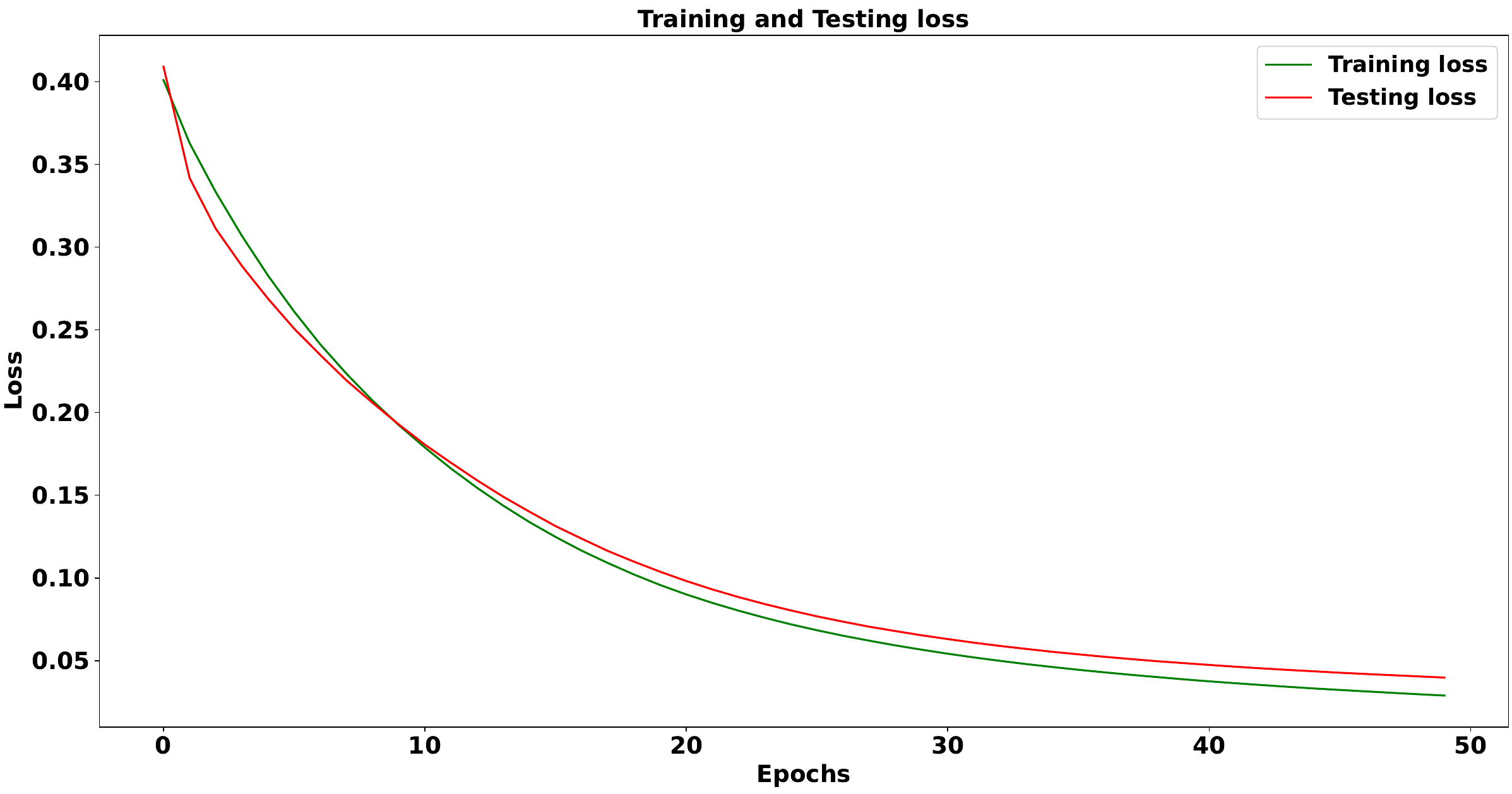} 
        \caption{Training-Testing Loss Curve for Blurred Case using Dataset2\label{traininglossblurreddataset2}}
    \end{subfigure}
    \hfill
    \begin{subfigure}{0.45\textwidth}
        \includegraphics[width=\linewidth]{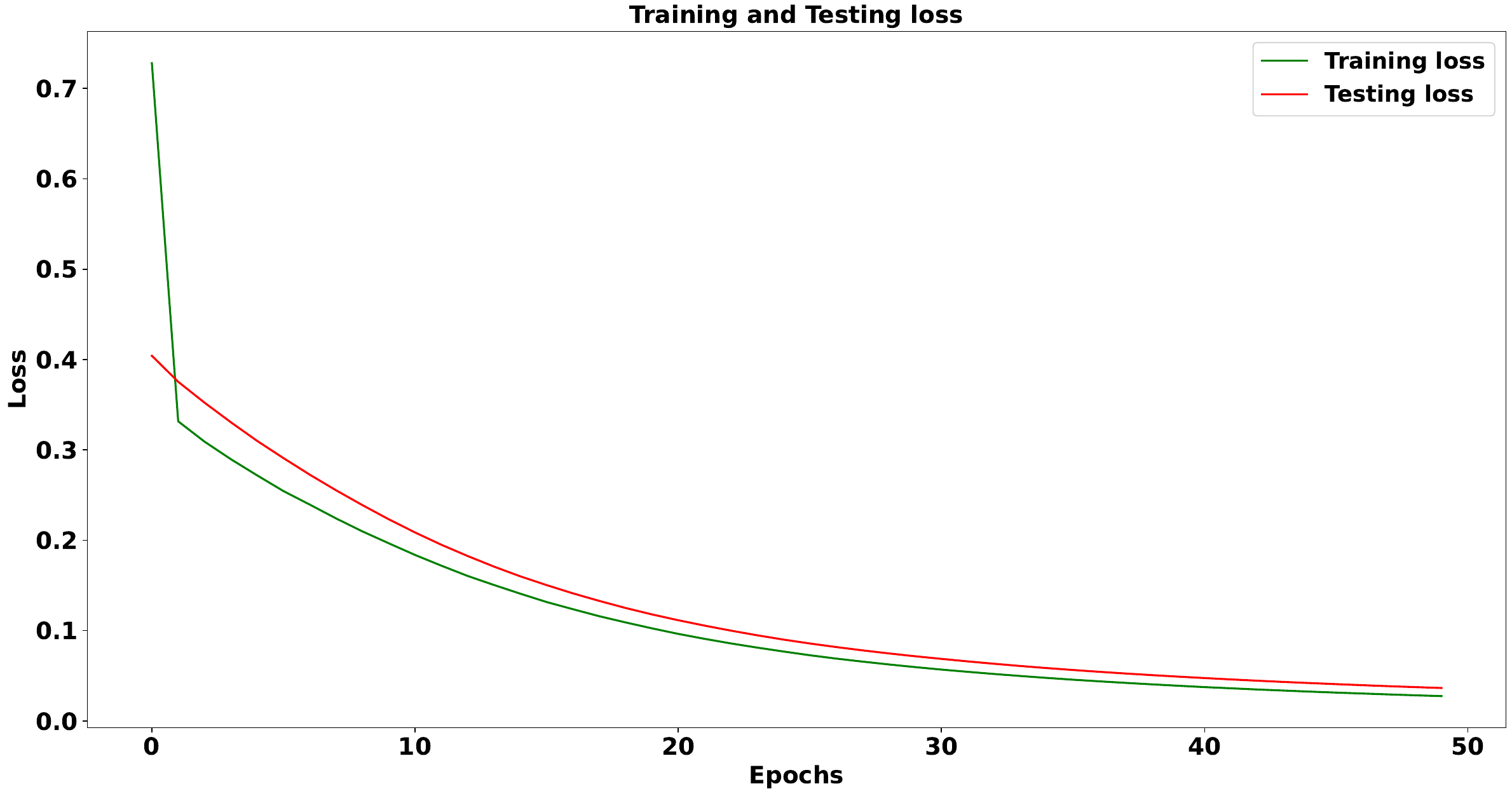}
        \caption{Training-Testing Loss Curve for Compressed Case using Dataset2\label{trainingcompresseddataset2}}
    \end{subfigure}
    \begin{subfigure}{0.45\textwidth}
        \includegraphics[width=\linewidth]{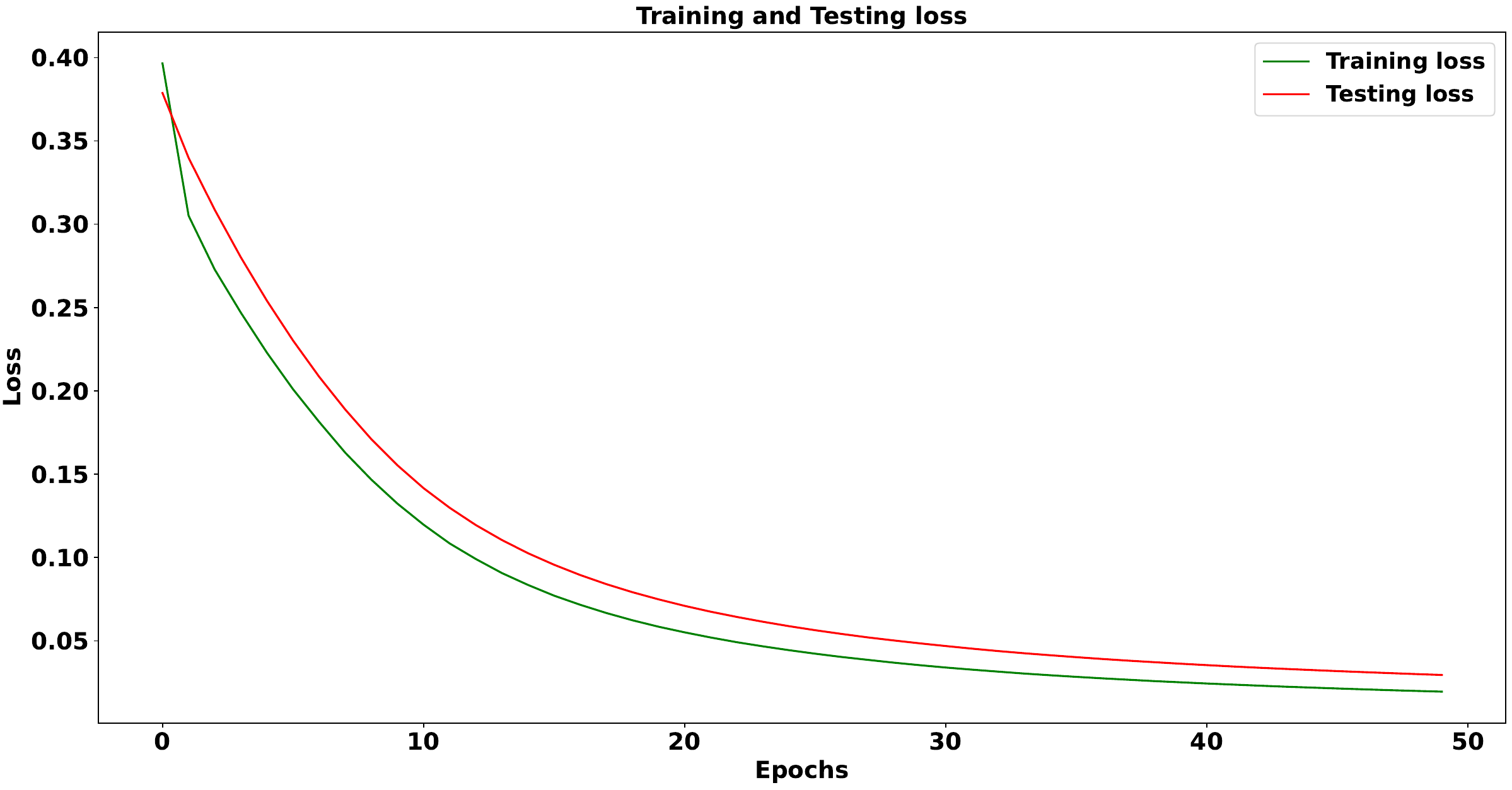} 
        \caption{Training-Testing Loss Curve for Wave2lip Case using Dataset2  \label{traininglosswav2lipdataset2}}
    \end{subfigure}
    \hfill
    \begin{subfigure}{0.45\textwidth}
        \includegraphics[width=\linewidth] {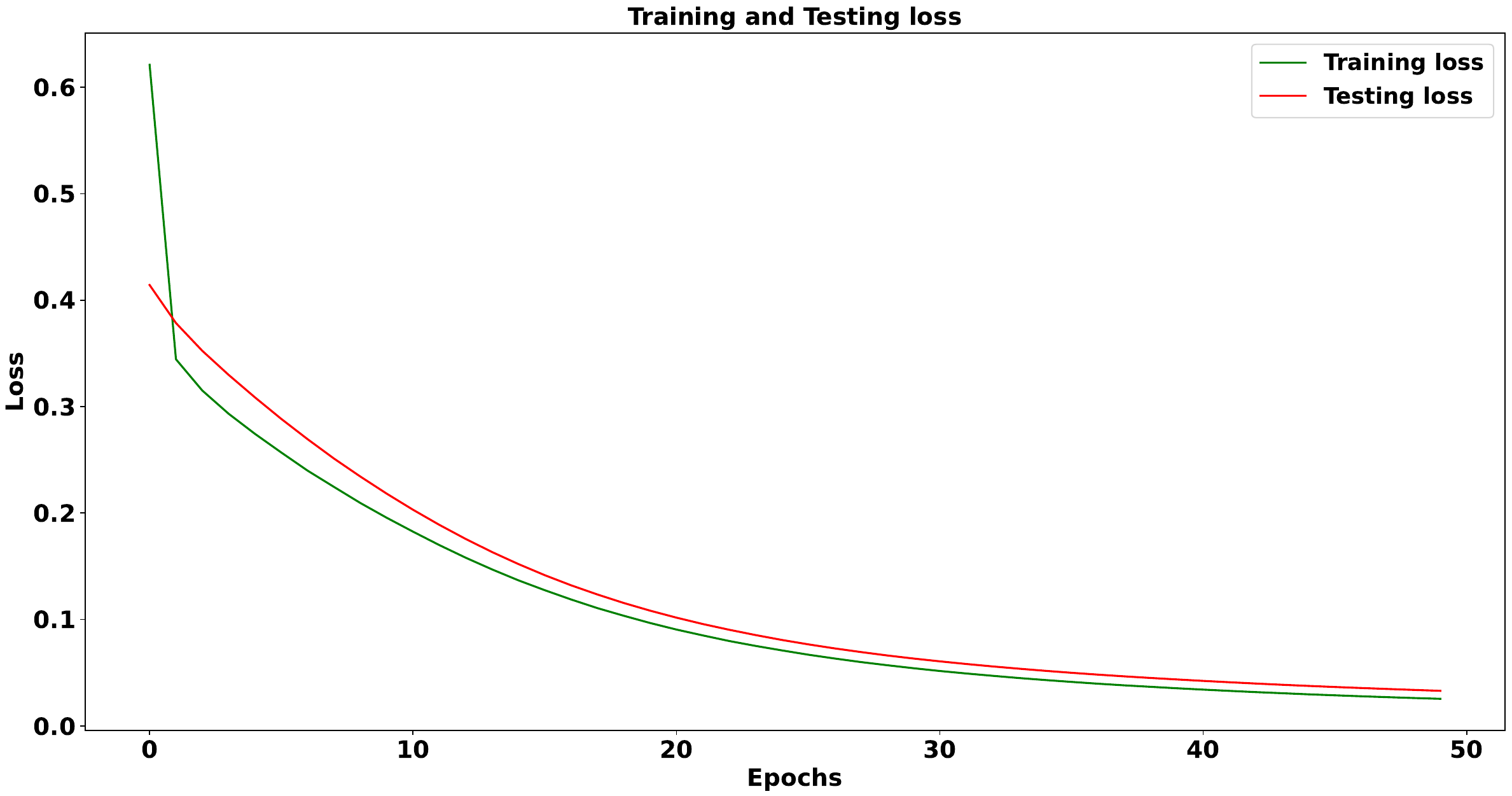} 
        \caption{Training-Testing Loss Curve for Faceswapped Case using Dataset2\label{traininglossfaceswappeddataset2}}
    \end{subfigure}    
    \begin{subfigure}{0.45\textwidth}
        \includegraphics[width=\linewidth]{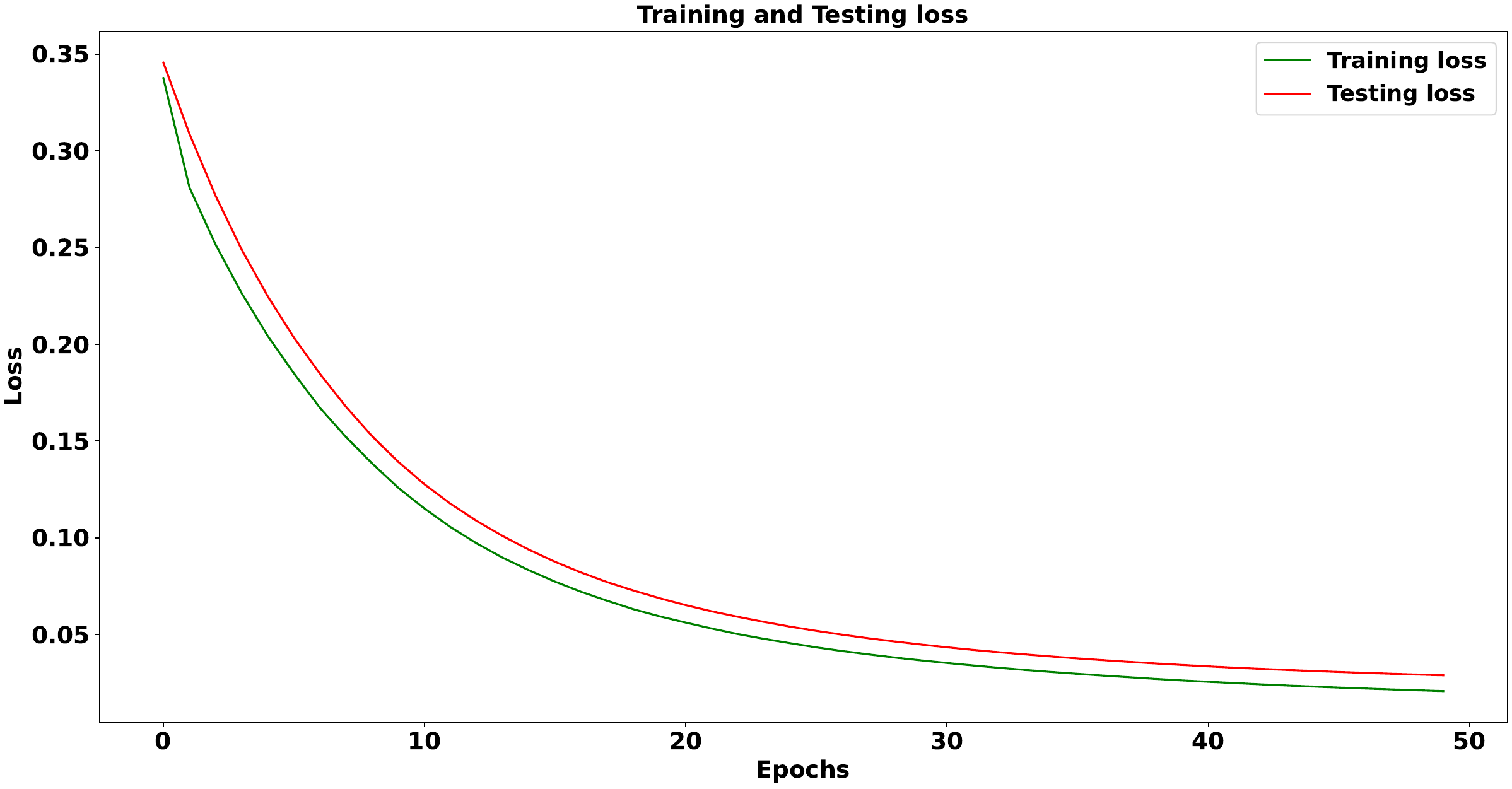} 
        \caption{Training-Testing Loss Curve for Resized Case using Dataset2 \label{traininglossresizeddataset2}}
    \end{subfigure}
    \caption{Analysis on Training-Testing Curve using Dataset2 (a) Training-Testing Loss for Original Case (b)  Training-Testing Loss for Noisy Case (c) Training-Testing Loss for Blurred Case (d) Training-Testing Loss for Compressed Case (e) Training-Testing Loss for Wav2lip Case (f) Training-Testing Loss for Faceswapped Case (g) F-Measure for Resized Case}
    \label{figtrainingloss10}
    \end{figure}

Figure \ref{traininglossoriginaldataset2} presents the training and testing loss analysis for the SqN-R-DFD scheme over $50$ epochs. As the epochs increase, the training-testing loss gradually declines. Both training and testing loss reach a minimal loss value, signifying that the SqN-R-DFD approach generalizes well to unseen data and reduces overfitting. The slight gap between the training and testing loss curves indicates a steady and consistent model performance.

\subsubsection{\textbf{Comparative Analysis on Feature Selection}}\label{featureselection2}
Table \ref{table6} explains the comparative evaluation of the SMA-SOA approach in comparison to conventional feature selection methods such as PCA, RFE-RF, SOA, KOA, and SMA feature selection methods for deepfake detection. The SMA-SOA strategy achieved the highest accuracy of $0.942$ and F-measure of $0.972$, while the PCA, RFE-RF, SOA, KOA, and SMA attained the least accuracy and F-measure values.
\begin{table}[H]
 \centering
    \begin{tabular}{|l|l|l|}
     \hline
      \textbf{Methods} & \begin{tabular}[c]{@{}c@{}} \textbf{Accuracy} \end{tabular}  & \begin{tabular}[c]{@{}c@{}}\textbf{F-measure}\\ \end{tabular} \\  
 \hline    
PCA & 0.919 & 0.902 \\ \hline
RFE-RF & 0.924 & 0.919 \\ \hline 
SOA  & 0.918 & 0.901 \\ \hline
KOA & 0.922 & 0.913 \\ \hline
SMA & 0.921 & 0.903 \\ \hline
SMA-SOA & 0.942 & 0.972\\ \hline
\end{tabular}%
\caption{Analysis on Feature Selection Methods for Dataset2}
    \label{table6}
\end{table}

\subsubsection{\textbf{Ablation Study}} \label{ablation2}
The ablation evaluation on the SqN-R-DFD strategy is compared with diverse variants such as SqN-R-DFD with existing PNCC, SqN-R-DFD with existing SLBT, and SqN-R-DFD with existing Feature Selection for deepfake detection, illustrated in Table \ref{table7}. The SqN-R-DFD with existing feature selection accomplished the specificity values exceeding $90\%$. Nonetheless, the SqN-R-DFD approach achieved the greatest specificity score of $0.986$, indicating enhanced performance in deepfake detection.
\begin{table}[H]
 \centering
    \begin{tabular}{|l|l|l|l|l|}
     \hline
 \textbf{Metrics} & \begin{tabular}[c]{@{}c@{}}\textbf{SqN-R-DFD with}\\ \textbf{Existing PNCC} \end{tabular} & \begin{tabular}[c]{@{}c@{}}\textbf{SqN-R-DFD with}\\ \textbf{Existing SLBT}\end{tabular}  & \begin{tabular}[c]{@{}c@{}}\textbf{SqN-R-DFD with}\\ \textbf{Existing Feature}\\ \textbf{Selection} \end{tabular}  & \begin{tabular}[c]{@{}c@{}}\textbf{SqN-R-DFD} \end{tabular}       \\ 
 \hline    
Accuracy & 0.926 & 0.915 & 0.921 & 0.942 \\ \hline
FPR & 0.046 & 0.056 & 0.037 & 0.014 \\ \hline 
F-Measure & 0.911  & 0.899 & 0.903 & 0.972 \\ \hline
Sensitivity & 0.889 & 0.877 & 0.864 & 0.950 \\ \hline
FNR & 0.111 & 0.123 & 0.136 & 0.050  \\ \hline
Specificity & 0.954 & 0.944 & 0.963 & 0.986 \\ \hline
NPV & 0.920 & 0.911 & 0.904 & 0.929  \\ \hline
Precision & 0.935 & 0.922 & 0.946 & 0.995  \\ \hline
MCC & 0.849 & 0.827 & 0.839 & 0.882 \\ \hline
    \end{tabular}%
    \caption{Ablation Analysis on SqN-R-DFD approach, SqN-R-DFD with Existing PNCC, SqN-R-DFD with Existing SLBT and SqN-R-DFD with Existing feature selection using Dataset2}
    \label{table7}
\end{table}
\subsubsection{\textbf{Comparison of SqN-R-DFD Model with Cross-Dataset Performance and Unbalanced Data}} \label{crossdataset2}
Table \ref{table8} shows the performance comparison of the SqN-R-DFD strategy with cross-dataset analysis and unbalanced data. The SqN-R-DFD model achieved the least FNR score of $0.050$, compared to cross-dataset ($0.126$) and unbalanced data ($0.075$). The outcomes signify that the SqN-R-DFD approach (balanced data) provides a significant enhancement in performance compared to cross-dataset and unbalanced data.

\begin{table}[H]
 \centering
    \begin{tabular}{|l|l|l|l|l|}
     \hline
 \textbf{Metrics} & \begin{tabular}[c]{@{}c@{}} \textbf{Cross Dataset Analysis}\\ \textbf{(Trained using Dataset2}\\ \textbf{and tested on Dataset1)} \end{tabular} & \begin{tabular}[c]{@{}c@{}}\textbf{Unbalanced Data} \end{tabular}  & \begin{tabular}[c]{@{}c@{}}\textbf{SqN-R-DFD}\\ \end{tabular} \\ 
 \hline    
Accuracy & 0.933 & 0.939 & 0.942 \\ \hline
Sensitivity & 0.874 & 0.925 & 0.950 \\ \hline 
Specificity & 0.976 & 0.974 & 0.986 \\ \hline
Precision & 0.976 & 0.971 & 0.995 \\ \hline
F-Measure & 0.922 & 0.948 & 0.972 \\ \hline
MCC & 0.872 & 0.876 & 0.882 \\ \hline
NPV & 0.888 & 0.922 & 0.929  \\ \hline
FPR & 0.024 & 0.026 & 0.014  \\ \hline
FNR & 0.126 & 0.075 & 0.050\\ \hline
    \end{tabular}%
    \caption{Performance Analysis for SqN-R-DFD Model compared to Cross-Dataset Analysis and Unbalanced Data using Dataset2}
    \label{table8}
\end{table}
\subsubsection{\textbf{Analysis on K-Fold Cross Validation}} \label{kfold2}
Table \ref{table9} explains the K-Fold cross-validation analysis on the SqN-R-DFD approach in comparison to conventional methods such as SqueezeNet, RNN, Bi-LSTM, LinkNet, DCNN, DBN \cite{suganthi2022deep}, and SVM \cite{tu2024face} for deepfake detection. The SqN-R-DFD scheme exhibits higher performance with minimal variations across K-Folds, signifying reduced overfitting. Conversely, the conventional approach demonstrates the greater fluctuations in accuracy.

\begin{table}[H]
 \centering
 \resizebox{\textwidth}{!}{
    \begin{tabular}{|l|l|l|l|l|l|l|l|l|}
     \hline
 \textbf{K-Folds} & \begin{tabular}[c]{@{}c@{}}\textbf{SqueezeNet}\\  \end{tabular} & \begin{tabular}[c]{@{}c@{}}\textbf{RNN} \end{tabular}  & \begin{tabular}[c]{@{}c@{}}\textbf{Bi-LSTM}\\ \end{tabular} & \begin{tabular}[c]{@{}c@{}}\textbf{LinkNet} \end{tabular}  & \begin{tabular}[c]{@{}c@{}}\textbf{DCNN} \end{tabular}  & \begin{tabular}[c]{@{}c@{}}\textbf{DBN \cite{suganthi2022deep}} \end{tabular}  & \begin{tabular}[c]{@{}c@{}}\textbf{SVM\cite{tu2024face}} \end{tabular}  & \begin{tabular}[c]{@{}c@{}}\textbf{Sqn-R-DFD} \end{tabular}\\ 
 \hline    
2 & 0.857 & 0.771 & 0.764 & 0.787 & 0.798 & 0.793 & 0.819 & 0.935 \\ \hline
3 & 0.881 & 0.800 & 0.790 & 0.793 & 0.811 & 0.811 & 0.822 & 0.942 \\ \hline
4 & 0.896 & 0.821 & 0.808 & 0.798 & 0.846 & 0.842 & 0.848 & 0.965 \\ \hline
5 & 0.898 & 0.823 & 0.811 & 0.802 & 0.853 & 0.852 & 0.897 & 0.967 \\ \hline
6 & 0.910 & 0.827 & 0.843 & 0.824 & 0.879 & 0.865 & 0.900 & 0.977 \\ \hline
\end{tabular}%
}
 \caption{K-Fold Cross Validation on SqN-R-DFD and Conventional Methods in terms of Accuracy using Dataset2}
\label{table9}
\vspace{-0.3cm}
\end{table}

\section{\textbf{Conclusion}} \label{conclusion}
Our research proposes a novel approach for the effective detection of deepfakes, employing a multimodal data fusion technique that integrates information from two modalities: face images and speech signals extracted from input videos. The initial step involved preprocessing the face images and speech signals to enhance their quality, employing the face cascade method for face images and the DSN approach for speech signals. Subsequently, feature extraction was performed on the pre-processed face images, encompassing both local and global features, as well as ISLBT, DSBME, and deep features. Similarly, features such as MPNCC, MFCC, and chroma features were extracted from preprocessed speech signals. The HFS technique was then introduced to optimally select features using the SASMA strategy. Finally, in the detection phase, models such as SqueezeNet and RNN were used to train the selected features individually, resulting in an efficient and accurate detection output. At $90\%$ training data, the highest F-measure value is achieved using the SqN-R-DFD approach is $97.453$, whereas the conventional methods attained the least F-measure values with SqueezeNet at $90.543$, RNN at $83.621$, Bi-LSTM at $82.156$, LinkNet at $84.532$, DCNN at $85.954$, DBN \cite{suganthi2022deep} at $88.521$, and SVM\cite{tu2024face} at $87.413$, respectively. In the near future, the model could be enhanced by testing on multilingual and diverse demographic datasets to ensure scalability and generalization across the globe.
\bibliographystyle{elsarticle-num}
\bibliography{mybib}

\end{document}